\documentclass[10pt,twocolumn,letterpaper]{article}

\usepackage[pagenumbers]{cvpr} 

\usepackage{tikz}
\usetikzlibrary{shapes.geometric, arrows.meta, positioning, calc, backgrounds, decorations.pathreplacing, fit}

\usepackage{amsthm}
\usepackage{amsmath}          
\usepackage{amssymb}          
\usepackage{amsfonts}         
\usepackage{bm}               
\usepackage{bbm}
\usepackage{mathtools}        
\usepackage{dsfont}
\usepackage{comment}

\usepackage[T1]{fontenc}      
\usepackage[utf8]{inputenc}   
\usepackage{microtype}        

\usepackage{graphicx}         
\usepackage{caption}          
\usepackage{subcaption}       
\usepackage{float}            

\usepackage{booktabs}         
\usepackage{tabularx}         
\usepackage{multirow}         
\usepackage{array}            
\usepackage{adjustbox}        
\usepackage{siunitx}          

\usepackage{algorithm}        
\usepackage{algpseudocode}    
 
\usepackage{xcolor}           
\usepackage{enumitem}         
\usepackage{lipsum}           
\usepackage{todonotes}        

\usepackage{pgfplots}
\pgfplotsset{compat=1.18}
\usepgfplotslibrary{groupplots}
\usepgfplotslibrary{statistics}
\usepackage{pgfplotstable}

\makeatletter
\newenvironment{customlegend}[1][]{%
    \begingroup
    \pgfplots@init@cleared@structures
    \pgfplotsset{#1}%
}{%
    \pgfplots@createlegend
    \endgroup
}%
\def\addlegendimage{\pgfplots@addlegendimage}

\usepackage{import}
\usepackage{textcomp}

\usepackage{pifont}
\newcommand{\cmark}{\ding{51}}
\newcommand{\xmark}{\ding{55}}

\newcommand{\ours}{\textsc{Manekko}\xspace}

\newcommand{\omours}{OM-\textsc{Manekko}\xspace}
\newcommand{\pcours}{PC-\textsc{Manekko}\xspace}
\usepackage{twemojis}
\newcommand{\ice}{\texttwemoji{ice}\xspace}
\newcommand{\fire}{\texttwemoji{fire}\xspace}
\newcommand{\dev}{$^\dagger$}
\usepackage{xspace}

\definecolor{oursrow}{rgb}{0.90,0.95,1.0}
\definecolor{paars}{RGB}{132, 86, 219}       
\definecolor{blauw}{RGB}{86, 154, 219}       
\definecolor{groen}{RGB}{86, 219, 170}       
\definecolor{lichtgroen}{RGB}{116, 219, 86}
\definecolor{gele}{RGB}{219, 208, 86}        
\definecolor{rode}{RGB}{219, 94, 86}         
\definecolor{grijs}{RGB}{128, 128, 128}      
\colorlet{paarsD}{paars!75!black}
\colorlet{blauwD}{blauw!70!black}
\colorlet{groenD}{groen!60!black}
\colorlet{lichtgroenD}{lichtgroen!60!black}
\colorlet{geleD}{gele!60!black}
\colorlet{rodeD}{rode!75!black}
\colorlet{grijsD}{grijs!70!black}
\tikzset{
    m om/.style={draw=paarsD, fill=paars},
    m pc/.style={draw=blauwD, fill=blauw},
    m nnf/.style={draw=groenD, fill=groen},
    m dino/.style={draw=geleD, fill=gele},
    m ssl/.style={draw=rodeD, fill=rode},
    m rand/.style={draw=grijsD, fill=grijs},
}
\definecolor{devisles}{HTML}{D55E00}
\definecolor{devacdc}{HTML}{0072B2}
\definecolor{devpros}{HTML}{009E73}
\pgfplotsset{
    dev isles/.style={devisles, thick, mark=*, mark size=1.6pt},
    dev acdc/.style={devacdc, thick, mark=square*, mark size=1.5pt},
    dev pros/.style={devpros, thick, mark=triangle*, mark size=2pt},
}
\tikzset{dummy/.style={draw=gray!60, fill=gray!12, inner sep=0pt, font=\tiny, text=gray!60!black, align=center}}
\newcommand{\pngslot}[3]{%
  \IfFileExists{figures/pca/#3.png}%
    {\includegraphics[width=#1,height=#2,keepaspectratio]{figures/pca/#3.png}}%
    {{\setlength{\fboxsep}{0pt}\fcolorbox{gray!60}{gray!12}{\parbox[c][#2][c]{#1}{\centering\tiny\color{gray!60!black}\texttt{\detokenize{#3}}}}}}}
\usepackage{colortbl}
\newcolumntype{g}{c}  

\definecolor{cvprblue}{rgb}{0.21,0.49,0.74}
\usepackage[pagebackref,breaklinks,colorlinks,allcolors=cvprblue]{hyperref}

\def\confName{CVPR}
\def\confYear{2027}

\title{
\ours: Training Small but Foundational 3D Medical Models 
by \\ Mimicking Natural Vision Model Features
}

\author{
Puru Vaish\textsuperscript{1}\thanks{Corresponding author, $^\dagger$Senior Supervision} \;
Michael Baumgartner\textsuperscript{2} \;
Tobias Heimann\textsuperscript{2} \;
Jelmer M. Wolterink\textsuperscript{1$\dagger$} \;
Christoph Brune\textsuperscript{1$\dagger$} \\
\textsuperscript{1}University of Twente, The Netherlands \quad
\textsuperscript{2}Siemens Healthineers, Erlangen, Germany \\
}

\begin{document}

\maketitle

\begin{abstract}
Self-supervised learning (SSL) for 3D medical imaging typically learns representations directly from unlabelled volumes.
We study whether features from pretrained 2D natural-image models provide effective targets to train 3D medical vision foundation models.
Our method, \ours, extracts features from axial, coronal, and sagittal slices, normalizes them using foreground statistics, and trains a 3D encoder to regress to the resulting patch and affinity statistics.
We pretrain separate models on brain MRI and multi-anatomy CT and evaluate over various anatomies, modalities and tasks.
Across 15 segmentation datasets, \ours achieves matches or exceeds in mean DSC among the evaluated encoders, both frozen and after full fine-tuning, with approximately $8\times$ fewer encoder parameters and $18\times$ fewer medical pretraining volumes compared to larger encoders.
Additional frozen classification, regression, and retrieval evaluations demonstrate the utility of \ours's global representations.
We find that across seven reference models to regress from, in-context scene understanding on natural images strongly correlates with frozen segmentation evaluation providing a recipe to choose a reference model pre-hoc.
These results support natural-image features as effective targets for training 3D medical foundation models\footnote{Code: \href{https://github.com/MIAGroupUT/Manekko}{github.com/MIAGroupUT/Manekko}.}.
\end{abstract}

\section{Introduction}\label{sec:intro}

\begin{figure}[t]
    \centering
    \begin{tikzpicture}[
    bubble/.style={only marks, mark=*, mark size=2.5pt,
        mark options={line width=0.6pt}},
    point label/.style={font=\tiny, inner sep=1pt}
]
    \begin{axis}[
        name=teaser,
        width=\linewidth,
        height={0.7\linewidth+0.25cm+4ex},
        xlabel={Full fine-tuning (mean DSC, \%)},
        ylabel={Frozen encoder (mean DSC, \%)},
        xmin=54, xmax=76, ymin=48, ymax=70,
        grid=major, grid style={gray!15},
        tick label style={font=\footnotesize},
        label style={font=\small},
        clip=true,
    ]
        \addplot[draw=none, fill=gray!12]
            coordinates {(48,48) (76,76) (48,76)} -- cycle;

        \node[font=\scriptsize, text=gray!70!black, anchor=north west]
            at (axis cs:54.3,69.7) {frozen $>$ fine-tuned};

        \addplot[gray!70, thick, domain=48:76] {x};
        \foreach \c in {110,120,130,140} {
            \addplot[gray!50, dashed, domain=48:76] {\c - x};
        }

        \draw[-{Stealth[length=2mm]}, gray!70!black, thick]
            (axis cs:69.5,51) -- (axis cs:71.0,54.75)
            node[right, font=\scriptsize] {better};

        \addplot[grijsD, densely dotted, thick, domain=48:76] {49.7};
        \addplot[grijsD, densely dashdotted, thick]
            coordinates {(64.8,48) (64.8,70)};

        \addplot[bubble, draw=rodeD, fill=rode]
            coordinates {(69.0,62.6) (67.8,59.9)};
        \addplot[bubble, draw=geleD, fill=gele]
            coordinates {(55.8,60.6)};
        \addplot[bubble, draw=groenD, fill=groen]
            coordinates {(72.6,66.5)};
        \addplot[bubble, draw=paarsD, fill=paarsD!75!white]
            coordinates {(73.1,68.0)};
        \addplot[bubble, draw=blauwD, fill=blauwD!75!white]
            coordinates {(71.9,66.3)};

        \addplot[bubble, draw=rodeD, fill=rode] coordinates {
            (65.6,57.4)
            (67.2,59.4)
            (65.8,57.8)
            (66.5,58.8)
            (67.6,59.5)
        };
        
        \node[point label, anchor=north east, text=rodeD]
            at (axis cs:65.4,57.2) {VoCo};
        \node[point label, anchor=south east, text=rodeD]
            at (axis cs:67.0,59.6) {MG};
        \node[point label, anchor=south east, text=rodeD]
            at (axis cs:65.6,57.6) {SwinUNETR};
        \node[point label, anchor=east, text=rodeD]
            at (axis cs:66.3,58.8) {SimCLR};
        \node[point label, anchor=south west, text=rodeD]
            at (axis cs:67.5,58.5) {VF};

        \node[point label, anchor=west, text=rodeD]
            at (axis cs:69.3,62.6) {MAE};
        \node[point label, anchor=west, text=rodeD]
            at (axis cs:68.1,59.9) {SimMIM};
        \node[point label, anchor=north, text=geleD]
            at (axis cs:55.8,60.1) {3DINO};
        \node[point label, anchor=north, text=groenD]
            at (axis cs:74.2,66.2) {nnFoundation};
        \node[point label, anchor=south, text=paarsD]
            at (axis cs:72.5,68.3) {\ours (ours)};
        \node[point label, anchor=east, text=blauwD]
            at (axis cs:71.3,66.3) {\ours (ours)};

        \node[point label, anchor=south west, text=grijsD]
            at (axis cs:64.9,48.0) {\scriptsize scratch};
        \node[point label, anchor=south west, text=grijsD]
            at (axis cs:54.2,49.7) {\scriptsize random encoder};
    \end{axis}
\end{tikzpicture}

\begin{tikzpicture}[
    bubble/.style={only marks, mark=*,
        mark options={line width=0.6pt}},
    bubble nnf/.style={draw=groenD, fill=groen},
    bubble dino/.style={draw=geleD, fill=gele},
    bubble ssl/.style={draw=rodeD, fill=rode},
    bubble pcmanekko/.style={draw=blauwD, fill=blauwD!75!white},
    bubble ommanekko/.style={draw=paarsD, fill=paarsD!75!white}
]
    \begin{customlegend}[
        legend columns=3,
        legend cell align=left,
        legend style={
            font=\scriptsize,
            draw=none,
            column sep=1.75mm,
            row sep=0.4mm,
            inner sep=0pt
        },
        legend entries={
            {\shortstack[l]{nnFoundation\\2.1M vols.; 551M}},
            {\shortstack[l]{3DINO\\100k vols.; 308M}},
            {\shortstack[l]{PrimusM baselines\\120k vols.; 156M}},
            {\shortstack[l]{\ours \\120k vols.; 68M}},
            {\shortstack[l]{\ours \\22k vols.; 68M}}
        }
    ]
        \addlegendimage{bubble, bubble nnf, mark size=2.000ex}
        \addlegendimage{bubble, bubble dino, mark size=1.2ex}
        \addlegendimage{bubble, bubble ssl, mark size=0.6ex}
        \addlegendimage{bubble, bubble ommanekko}
        \addlegendimage{bubble, bubble pcmanekko}
    \end{customlegend}
\end{tikzpicture}
    \caption{
        \ours performs strongly in both full-fine-tune and frozen performance despite using an encoder approximately $8\times$ smaller than the largest model shown and less broad pretraining datasets. 
        The figure shows mean DSC over 15 segmentation datasets.
        Frozen evaluation measures how useful the pretrained representation out-of-the-box, while full fine-tuning measures its task-specific adaptability.
        \vspace{-2ex}
    }
    \label{fig:teaser}
\end{figure}
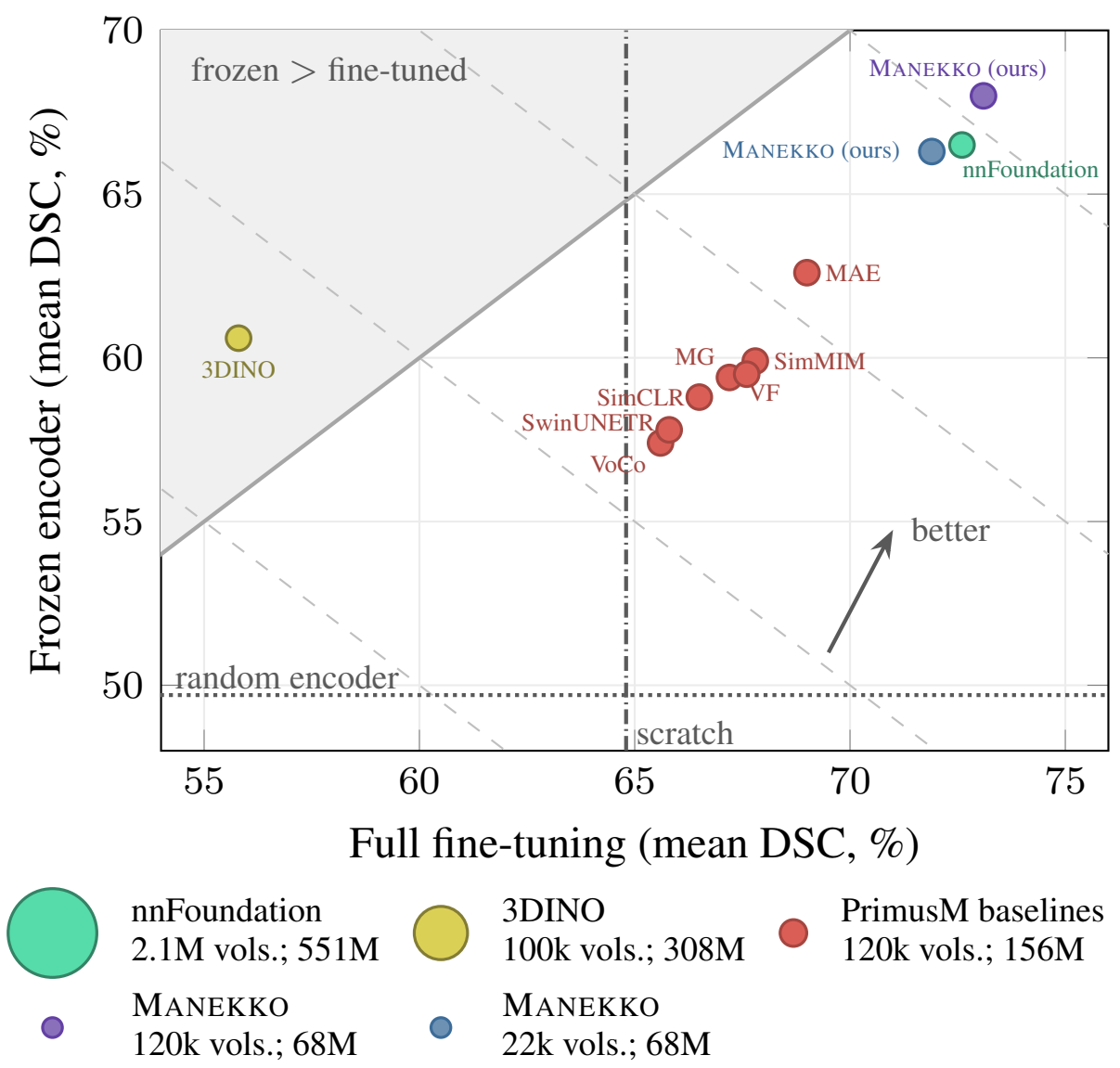

Self-supervised learning (SSL) in computer vision has turned large collections of unlabeled natural images into general-purpose visual representations.
From contrastive and Siamese objectives~\cite{chen2020simclr,chen2021simsiam}, through self-distillation~\cite{caron2021dino,zhou2022ibot,oquab2024dinov2} and latent prediction~\cite{assran2023ijepa}, to objectives that explicitly target dense scene understanding~\cite{ziegler2022leopart,pariza2025neco}, progress has increasingly produced patch features that capture both semantic content and spatial structure.
These representations can be useful even without adapting the encoder: linear probes and nearest-neighbour label transfer already recover semantic structure from frozen features~\cite{balazevic2023hummingbird,oquab2024dinov2}.

In volumetric medical image analysis, progress has followed a similar trajectory.
Increasingly large collections of unlabeled 3D medical scans have become available~\cite{wald2025openmind,munkFOMO,hamamci2024ctrate,flare2026}, and many SSL strategies have been adapted to volumetric data~\cite{zhou2019modelsgenesis,tang2022swinunetr,wu2024voco,xie2022simmim,he2022mae,wang2023misfm,wald2025revisitingmae,cvaVaish,nnfoundation} to then be used for segmentation, regression, classification and retrieval tasks. 
This is a logical approach: medical imaging volumes are abundant while labels are expensive, and volumetric models should benefit from learning directly from the domain in which they will be used.

Yet medical pretraining data differ in structure from the natural-image corpora~\cite{intrinsicManifold} on which modern vision foundation models are trained.
Within a fixed modality and anatomy, many volumes contain the same organs in similar locations, with similar acquisition physics and intensity patterns.
A dataset may therefore grow substantially in the number of scans without growing equally in the diversity~\cite{beyondscaling2025} of visual concepts presented to the learning objective, or, alternatively, the differentiating signal might be too small.
A natural-image foundation model has already organized a much broader visual world into semantic and spatially meaningful features.
Even when applied to an unfamiliar medical image, those features could provide a structured target that is richer than the raw signal from which a 3D SSL objective must learn from scratch.

Therefore, one possibility is simply to use the natural image model directly to obtain medical image embeddings.
Pretrained natural image models can be applied to medical volumes slice by slice~\cite{medsam,medinov3}, and recent work successfully adapts these models for medical image \emph{classification}~\cite{revisiting2dfm3dcls}.
However, independent slice encoding provides only planar context, requires repeated 2D inference throughout a volume, and does not yield a native 3D encoder.
These limitations matter most for segmentation, where representations must remain spatially precise and coherent in three dimensions~\cite{medAdapt3d,dinounet,dinoreg}.

This motivates a different use of natural-image foundation models: not as the final model to be applied to medical images, but as a source of supervision for a medical imaging model.
Given a fixed collection of unlabelled medical volumes, we therefore ask:
\emph{for training a medical image foundation model, is it effective to train a 3D encoder to reproduce the representation that a pretrained natural-image model produces on the medical volumes?}

We propose \ours\footnote{Japanese for ``copycat'', used for children imitating others playfully} to answer this question.
\ours applies a frozen reference natural image model to all axial, coronal, and sagittal slices of a volume and combines the resulting features into a 3D transformer.
Because image feature statistics shift when the natural image model is applied to medical scans, we normalize the target using statistics estimated over foreground regions identified by the reference's global-to-patch affinities.
The 3D transformer is then trained with two regression terms, matching patch features and global-to-patch affinities, without trainable projectors or a momentum teacher.
After pretraining, the reference model is discarded, and the newly trained 3D medical image transformer operates directly on 3D volumes.

We evaluate the learned representation from two complementary perspectives.
With the 3D encoder frozen, downstream performance reflects how much useful structure is already present in the pretrained features.
With full fine-tuning of the 3D encoder, we measure how well that initialization adapts to segmentation in particular.
We evaluate both regimes across 15 3D CT and MRI segmentation datasets, and additionally evaluate frozen global representations on classification, regression, and retrieval tasks.
Our contributions can be summarized as:
\begin{enumerate}[leftmargin=*,itemsep=1pt,topsep=2pt]
    \item We introduce a method for distilling dense natural-image representations into a native 3D medical encoder using tri-plane targets, foreground-aware feature normalization, and global-to-patch affinity regression.
    \item On the evaluated benchmarks, the resulting compact encoder outperforms SSL encoders pretrained on the same medical data on average and remains competitive with, or exceeds, substantially larger medical vision foundation models both frozen and after full fine-tuning (\cref{fig:teaser}).
    \item Across seven reference models, natural-image in-context scene understanding strongly correlates with frozen segmentation evaluation, providing a reference-selection criterion that requires no medical annotations.
\end{enumerate}

\section{Related Work}\label{sec:related}

\noindent\textbf{Self-supervised learning in natural images.}
Natural-image SSL has progressed from contrastive and Siamese learning~\cite{chen2020simclr,chen2021simsiam}, to self-distillation~\cite{caron2021dino,zhou2022ibot,oquab2024dinov2,simeoni2025dinov3}, and masked prediction of pixels~\cite{he2022mae,xie2022simmim} or latent features~\cite{assran2023ijepa}.
Beyond global image representations, recent methods increasingly target patch-level semantic structure.
DINOv2 and its variants with registers~\cite{darcet2024registers} support frozen dense prediction, while Leopart~\cite{ziegler2022leopart}, NeCo~\cite{pariza2025neco}, and CrIBo~\cite{lebailly2024cribo} explicitly improve dense representations.
Rather than adapting these objectives to 3D, we use the resulting frozen patch features as supervision and study which properties of a natural-image reference predict the quality of the distilled medical representation.

\noindent\textbf{Self-supervised learning for 3D medical images.}
Medical SSL defines supervision directly from unlabelled volumes through restoration~\cite{zhou2019modelsgenesis}, contrastive, geometric, and contextual objectives~\cite{tang2022swinunetr,wu2024voco,he2023gvsl,cvaVaish}, volume fusion~\cite{wang2023misfm}, masked modelling~\cite{wald2025revisitingmae,nnfoundation}, and self-distillation~\cite{3dino}.
Benchmarking efforts provide several such methods under shared architectures and evaluation protocols~\cite{wald2025openmind,wald2025revisitingmae}.
Our approach differs in its supervision source: the medical volumes provide the inputs, but a pretrained natural-image model produces the regression targets.
This allows us to compare distillation with SSL baselines trained on the same medical corpus while changing how the training signal is constructed.

\noindent\textbf{Medical pretraining data.}
Large public corpora now cover brain MRI~\cite{wald2025openmind,fomo300k}, chest CT with reports~\cite{hamamci2024ctrate}, and multi-anatomy CT~\cite{flare2026}, enabling increasingly large image-only and image--language models~\cite{blankemeier2024merlin}.
At the same time, prior work has examined when in-domain medical pretraining improves over natural-image initialization~\cite{wen2021rethinking} and whether informative subsets can reduce the amount of medical pretraining data required~\cite{beyondscaling2025}.
These observations motivate considering not only how much medical data is used, but also what supervision is extracted from it.
We therefore study distillation on the same pretraining volumes as medical SSL baselines where possible, and separately examine sensitivity to the amount of medical pretraining data.

\noindent\textbf{Natural image models for 3D medical volumes.}
Pretrained natural image models can be applied to medical volumes slice by slice, either frozen or after adaptation~\cite{medinov3,medsam,medAdapt3d,dinounet}.
This provides access to strong features but requires repeated slice encoding and does not itself produce a representation with native volumetric context.
Recent work adapts foundation models for 3D medical \emph{classification}~\cite{revisiting2dfm3dcls}.
We instead transfer dense representations into a 3D encoder, so that the reference model is used only during pretraining and the resulting model can subsequently be evaluated and fine-tuned directly for segmentation.

\noindent\textbf{Knowledge and feature distillation.}
Knowledge distillation transfers teacher predictions~\cite{hinton2015distilling}, intermediate features~\cite{romero2015fitnets,zagoruyko2017attention}, or relations between representations~\cite{park2019relational,tung2019similarity}.
It has been used to train transformers~\cite{touvron2021deit}, preserve pretrained features during adaptation~\cite{wei2022fd}, and combine several pretrained models into a single student~\cite{ranzinger2024radio,sariyildiz2024unic,shang2024theia}.
Across 2D and 3D domains, image features have also supervised point-cloud representations through pixel-to-point correspondences~\cite{zhangLearning3DRepresentations2023,sautier2022slidr,dinoRoom} and radiance fields through multi-view aggregation~\cite{kobayashi2022dff}.
Our setting differs in that no calibrated multi-view correspondence is available: each 3D location is observed through orthogonal medical slices whose reference features can differ across orientations.
We therefore construct a common volumetric target from three planes, normalize its feature statistics across domains, and regress it densely with a 3D student.

\noindent\textbf{In-context dense prediction.}\label{sec:incontext}
UniverSeg~\cite{butoi2023universeg}, Tyche~\cite{rakic2024tyche}, Neuralizer~\cite{czolbe2023neuralizer}, and SegMIC~\cite{segmic} perform medical segmentation from labelled support examples without task-specific retraining.
Hummingbird~\cite{balazevic2023hummingbird,pariza2024hbeval} instead evaluates the organization of frozen patch features through similarity-weighted nearest-neighbour label transfer from an annotated memory bank, without training a decoder.
We use Hummingbird mIoU on natural images to measure dense representation quality and test whether it correlates with the downstream performance of a 3D student, providing a reference-selection criterion that does not require medical labels.

\section{Method}
\label{sec:method}

\ours trains a 3D encoder on unlabelled medical volumes by regressing features of a frozen natural-image model.
This objective motivates choosing reference features whose semantic organization already supports dense prediction.
We hypothesize that stronger Hummingbird performance on natural images identifies better targets for 3D medical representation learning, and test this in \cref{sec:reference}.

\noindent\textbf{Tri-plane reference features.}
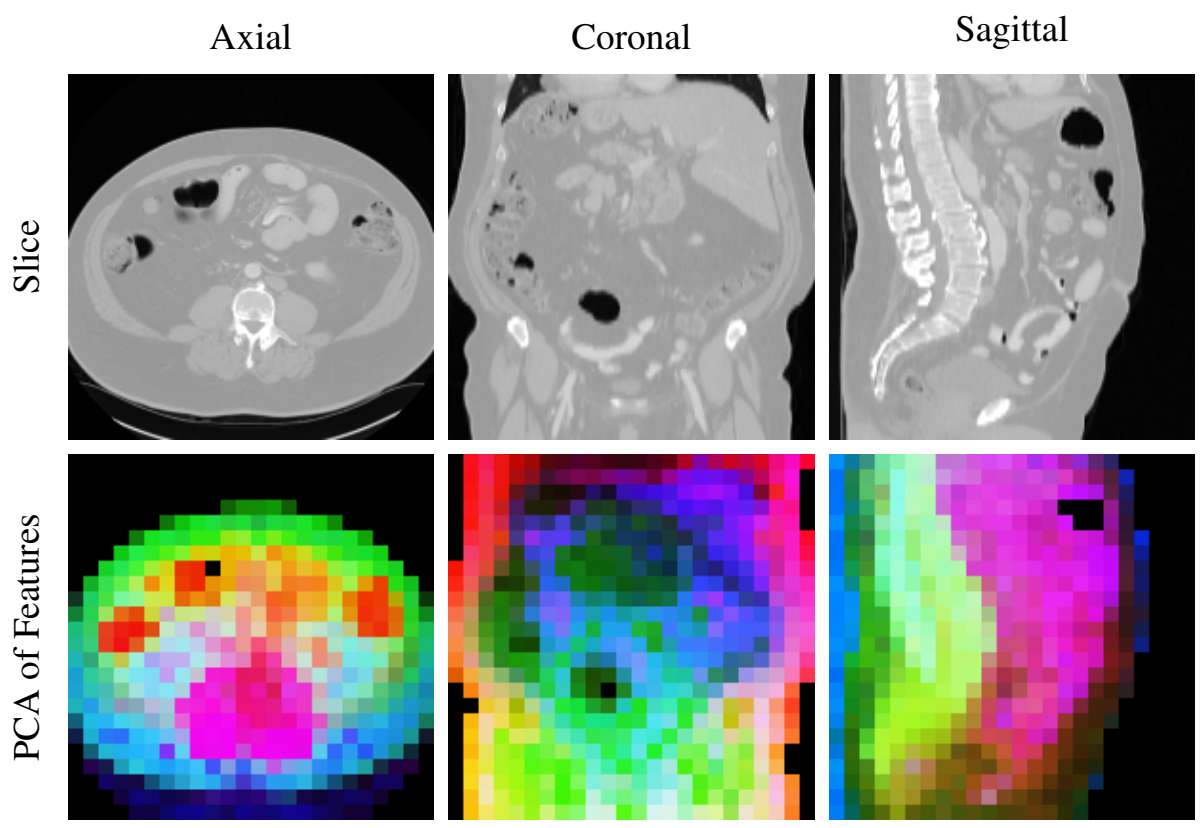
\begin{figure}[t]
    \centering
    \begin{tikzpicture}[lab/.style={font=\scriptsize}]
    \node[lab, anchor=south] at (0.000\linewidth, 0.150\linewidth) {Axial};
    \node[lab, anchor=south] at (0.302\linewidth, 0.150\linewidth) {Coronal};
    \node[lab, anchor=south] at (0.604\linewidth, 0.150\linewidth) {Sagittal};
    \node[font=\scriptsize, rotate=90, anchor=south] at (-0.153\linewidth, 0.000\linewidth) {Slice};
    \node[inner sep=0pt] at (0.000\linewidth, 0.000\linewidth) {\pngslot{0.290\linewidth}{0.290\linewidth}{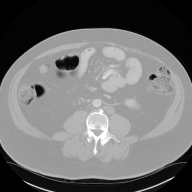}};
    \node[inner sep=0pt] at (0.302\linewidth, 0.000\linewidth) {\pngslot{0.290\linewidth}{0.290\linewidth}{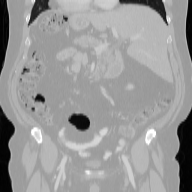}};
    \node[inner sep=0pt] at (0.604\linewidth, 0.000\linewidth) {\pngslot{0.290\linewidth}{0.290\linewidth}{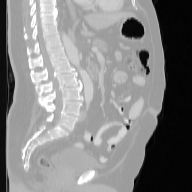}};
    \node[font=\scriptsize, rotate=90, anchor=south] at (-0.153\linewidth, -0.302\linewidth) {PCA of Features};
    \node[inner sep=0pt] at (0.000\linewidth, -0.302\linewidth) {\pngslot{0.290\linewidth}{0.290\linewidth}{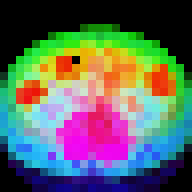}};
    \node[inner sep=0pt] at (0.302\linewidth, -0.302\linewidth) {\pngslot{0.290\linewidth}{0.290\linewidth}{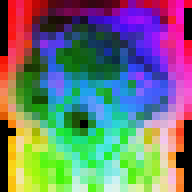}};
    \node[inner sep=0pt] at (0.604\linewidth, -0.302\linewidth) {\pngslot{0.290\linewidth}{0.290\linewidth}{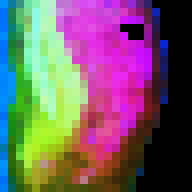}};
\end{tikzpicture}
    \caption{
        Patch features of the frozen natural-image model NeCo (DINOv2-R) on axial, coronal, and sagittal slices.
        RGB values show projections onto the first three principal components, fitted on one orientation and applied unchanged to all three.
        The features delineate anatomical structures but vary across orientations, motivating their combination into a single volumetric reference.
        \vspace{-2ex}
    }
    \label{fig:prelim}
\end{figure}
\begin{figure*}[t]
    \centering
    \def\svgwidth{\textwidth}
    \import{figures}{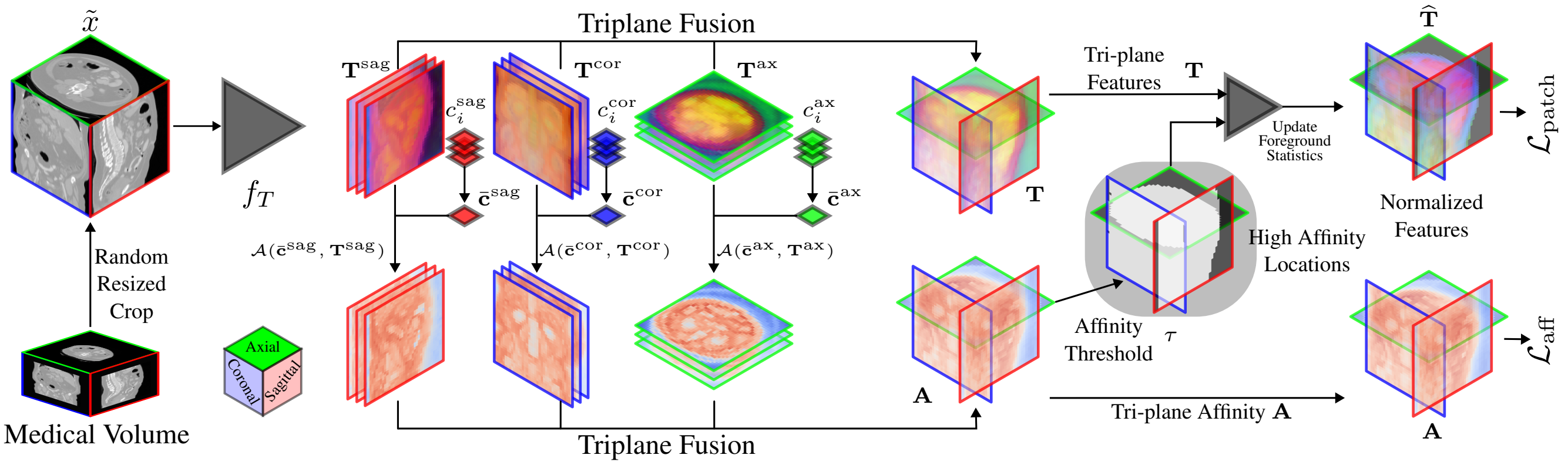}
    \caption{
    We extract the training signal from a frozen pretrained 2D model, $f_{T}$, which is applied slice-by-slice to medical volume along each axes $p\in \{\mathrm{ax}, \mathrm{cor}, \mathrm{sag}\}$, giving patch features, $\mathrm{T}^p$ and global representation per slice, $c_i^{p}$.
    $\mathcal{A}(\bar{\mathbf{c}}^p,\mathbf{T}^p)$ computes the affinity using per axes medoid, $\bar{\mathbf{c}}^p$ and the stacked axes features $\mathbf{T}^p$.
    Patch features are fused in a tri-plane fashion, $\mathbf{T}$, and similarly tri-plane fusion of per-axes affinities gives global-to-local affinity target $\mathbf{A}$.
    The foreground of $\mathbf{A}$ defines where feature statistics should be estimated to normalize $\mathbf{T}$ into $\widehat{\mathbf{T}}$.
    A 3D student is trained to regress $\widehat{\mathbf{T}}$ with its patch tokens and $\mathbf{A}$ with the affinity between its class and patch tokens.
    }
    \label{fig:method}
\end{figure*}
Given a volume $\mathbf{x}\in\mathbb{R}^{C_{\mathrm{in}}\times H\times W\times D}$, the student encoder $f_S$ produces a global token $\mathbf{s}_{\mathrm{cls}}\in\mathbb{R}^{C}$ and patch tokens $\mathbf{S}\in\mathbb{R}^{C\times H_p\times W_p\times D_p}$ on a 3D patch lattice $\Omega$ with $|\Omega|=H_pW_pD_p$.
A 2D encoder observes different planar contexts across orientations, yielding different features for the same anatomical structures (\cref{fig:prelim}).
We therefore apply the frozen reference $f_T$ independently to axial, coronal, and sagittal slices, then combine their features to supervise $\mathbf{S}$ and its affinities with $\mathbf{s}_{\mathrm{cls}}$.
Since $f_T$ expects natural images, each volume is first mapped to the unit range,
\begin{equation}
    \widetilde{\mathbf{x}}
    =
    \frac{\mathbf{x}-\min_{\mathbf{v}}\mathbf{x}(\mathbf{v})}
         {\max_{\mathbf{v}}\mathbf{x}(\mathbf{v})-\min_{\mathbf{v}}\mathbf{x}(\mathbf{v})},
    \label{eq:intensity_normalization}
\end{equation}
normalized with the ImageNet mean and standard deviation, and replicated to three channels.
This is applied only after the percentile normalization and only to the input of the reference model.
Let $\mathcal{P}=\{\mathrm{ax},\mathrm{cor},\mathrm{sag}\}$ denote the three planes and $\mathcal{S}_{p}(\widetilde{\mathbf{x}})=\{\mathbf{x}^{p}_{i}\}_{i=1}^{N_p}$ the slices along plane $p$.
For every slice, $f_T$ produces a class token and patch features,
\begin{equation}
    \left(
        \mathbf{c}^{p}_{i},
        \mathbf{Z}^{p}_{i}
    \right)
    =
    f_T\!\left(\mathbf{x}^{p}_{i}\right),
    \qquad
    p\in\mathcal{P}.
    \label{eq:teacher_slice}
\end{equation}
The slice-wise patch representations are stacked at their original positions along the slicing axis and lifted onto the common 3D patch lattice,
\begin{equation}
    \mathbf{T}^{p}
    =
    \mathcal{R}_{p}
    \left(
        \operatorname{Stack}_{i}
        \left[\mathbf{Z}^{p}_{i}\right]
    \right)
    \in
    \mathbb{R}^{C\times H_p\times W_p\times D_p},
    \label{eq:triplane_lift}
\end{equation}
where $\mathcal{R}_{p}$ average-pools along the slicing axis of plane $p$, which has one entry per voxel, to the resolution of the patch lattice.
The dense reference representation averages the three views,
\begin{equation}
    \mathbf{T}
    =
    \frac{1}{3}
    \sum_{p\in\mathcal{P}}
    \mathbf{T}^{p}.
    \label{eq:triplane_fusion}
\end{equation}
Compared to a single canonical plane, this exposes the student to structures that are informative in any of the three orientations and averages out plane-specific features of the 2D reference model.

\noindent\textbf{Student--reference patch resolution.}
The reference and student generally use different patch sizes, so evaluating the reference model at the native slice resolution yields an in-plane grid that differs from the student lattice.
We therefore choose the resolution at which the reference model is evaluated based on the student lattice rather than from the volume.
If the reference model uses patch size $q_T$ and the student lattice has in-plane size $(G_1,G_2)$, we resize each slice to
\begin{equation}
    (R_1,R_2)
    =
    \left(q_T G_1,\; q_T G_2\right),
    \label{eq:active_resolution}
\end{equation}
so that the reference model directly produces a $G_1\times G_2$ grid.
For example, a $192^3$ crop with student patch size $8^3$ gives $G=24$, so a reference model with $q_T=14$ is evaluated at $336\times336$.

\noindent\textbf{Global-to-patch affinity.}
Dense patch features support segmentation, while a global token provides a volume representation for classification, regression, and retrieval.
A volume yields multiple reference class tokens, and global and patch tokens need not share the same representation geometry~\cite{compositionalityPrimitives}.
We therefore supervise the student global token through its affinities with patch tokens.
For a class representation $\mathbf{c}$ and dense feature map $\mathbf{Z}$, define the cosine-affinity field
\begin{equation}
    \mathcal{A}(\mathbf{c},\mathbf{Z})(\mathbf{v})
    =
    \frac{
        \mathbf{c}^{\top}\mathbf{Z}(\mathbf{v})
    }{
        \|\mathbf{c}\|_2
        \|\mathbf{Z}(\mathbf{v})\|_2
    }.
    \label{eq:cosine_affinity}
\end{equation}
For each plane, we compute the Euclidean medoid,
\begin{equation}
    \bar{\mathbf{c}}^{p}
    =
    \underset{\mathbf{c}^{p}_{i}}{\arg\min}
    \sum_{j=1}^{N_p}
    \left\|
        \mathbf{c}^{p}_{i}
        -
        \mathbf{c}^{p}_{j}
    \right\|_2,
    \label{eq:cls_medoid}
\end{equation}
so that each plane is represented by an observed slice token.
The reference affinity target and the student affinity are
\begin{equation}
    \mathbf{A}
    =
    \frac{1}{3}
    \sum_{p\in\mathcal{P}}
    \mathcal{A}
    \left(
        \bar{\mathbf{c}}^{p},
        \mathbf{T}^{p}
    \right),
    \qquad
    \mathbf{A}_{S}
    =
    \mathcal{A}
    \left(
        \mathbf{s}_{\mathrm{cls}},
        \mathbf{S}
    \right).
    \label{eq:affinity}
\end{equation}
This constrains the global token through its spatial relationships while leaving the student free to choose a global representation appropriate for the 3D domain.

\noindent\textbf{Foreground feature normalization.}
Reference features on medical slices can have shifted channel means and variances, causing regression to emphasize high-magnitude channels.
We therefore standardize $\mathbf{T}$ per channel.
To reduce the influence of background regions such as air, we estimate the statistics on foreground estimated from the reference affinity map.
Using the affinity map $\mathbf{A}$, we define the foreground as the locations above the midpoint of its range,
\begin{equation}
    \tau
    =
    \tfrac{1}{2}
    \Big(
        \min \mathbf{A}
        +
        \max \mathbf{A}
    \Big),
    \;
    \mathbf{M}
    =
    \mathbbm{1}\!\left[
        \mathbf{A}\geq\tau
    \right],
    \label{eq:active_mask}
\end{equation}
computed per volume.
The channel mean $\bm{\mu}$ and variance $\bm{\sigma}^2$ of $\mathbf{T}$ over $\{\mathbf{v}:\mathbf{M}(\mathbf{v})=1\}$ are computed for each batch and tracked with an exponential moving average whose momentum is annealed to zero early in training, after which the statistics are fixed.
The target is
\begin{equation}
    \widehat{\mathbf{T}}
    =
    \frac{\mathbf{T}-\bm{\mu}}{\sqrt{\bm{\sigma}^2+\epsilon}}.
    \label{eq:domain_normalization}
\end{equation}
Freezing the statistics keeps the target stationary for most of training.
Note that the normalization is applied to the whole volume and the mask only determines where its statistics are estimated.
We ablate both the foreground mask and the normalization as a whole in \cref{sec:ablation}.

\noindent\textbf{Overall objective.}
The student regresses the unweighted, normalized patch and affinity target, given by:
\begin{align}
    \mathcal{L}_{\mathrm{patch}}
    &=
    \frac{1}{|\Omega|}
    \sum_{\mathbf{v}\in\Omega}
    \left\|
        \mathbf{S}(\mathbf{v})
        -
        \widehat{\mathbf{T}}(\mathbf{v})
    \right\|_2^2,
    \label{eq:patch_loss}
    \\
    \mathcal{L}_{\mathrm{aff}}
    &=
    \frac{1}{|\Omega|}
    \sum_{\mathbf{v}\in\Omega}
    \left(
        \mathbf{A}_{S}(\mathbf{v})
        -
        \mathbf{A}(\mathbf{v})
    \right)^2,
    \label{eq:affinity_loss}
\end{align}
\begin{equation}
    \mathcal{L}
    =
    \mathcal{L}_{\mathrm{patch}}
    +
    \mathcal{L}_{\mathrm{aff}}.
    \label{eq:final_objective}
\end{equation}
The reference model is frozen throughout.
The width of the student matches that of the reference model, so the student tokens regress the target directly and no projection head is needed.
Training uses only spatial augmentations shared by the student and reference, without any intensity-augmented views, global-local crops, masking, or a momentum encoder.
Pseudocode is given in \cref{supp:impl}. Summary of extracting the training signal is given in \cref{fig:method}.

\section{Experimental Setup and Results}\label{sec:experiments}

We validate \ours on both dense and global downstream tasks. 
Our primary evaluation is semantic segmentation across 15 datasets, where we test both the representation available \emph{out of the box} with a frozen encoder and its performance after full task-specific fine-tuning.
We also evaluate frozen global representations on classification, regression, and retrieval tasks spanning MRI and CT to place our method's utility more broadly.

\subsection{Experimental Setup}\label{sec:setup}

\noindent\textbf{Pretraining data.}
We pretrain separately on OpenMind~\cite{wald2025openmind} (OM), comprising 113{,}880 brain MRI volumes from 34{,}132 subjects, and PanCancerCT~\cite{flare2026} (PC), comprising 21{,}802 CT volumes across anatomies.
By default, we refer to the model trained on OpenMind as \ours.
When we need to distinguish the two models, we denote them with \omours and \pcours.
Dataset composition and preprocessing are detailed in Supp.~\ref{supp:data}.
\begin{table*}[t]
    \centering
    \caption{
    OpenMind pretraining, mean foreground DSC on 9 MRI datasets, with the \ice frozen encoder and after \fire full fine-tuning for 150 epochs. \ours is best on average under both evaluation regimes. \textbf{Bold} indicates best metric per dataset. \dev: developmental dataset.
    }
    \label{tab:main}
    \resizebox{\textwidth}{!}{%
        \begin{tabular}{l ccccccccc c ccccccccc c}
    \toprule
    & \multicolumn{5}{c}{Brain MRI \ice} & \multicolumn{4}{c}{Body MRI \ice} & & \multicolumn{5}{c}{Brain MRI \fire} & \multicolumn{4}{c}{Body MRI \fire} & \\
    \cmidrule(lr){2-6}\cmidrule(lr){7-10}\cmidrule(lr){12-16}\cmidrule(lr){17-20}
    Method & BrTu & Hipp & YBM & ISLES\dev & GLI & Heart & Pros & ACDC\dev & P158\dev & All & BrTu & Hipp & YBM & ISLES\dev & GLI & Heart & Pros & ACDC\dev & P158\dev & All \\
    \midrule
    Random init.                                 & 67.9 & 85.7 & 41.7 & 65.6 & 50.8 & 71.8 & 58.5 & 68.8 & 64.2 & {63.9} & 71.8 & 87.4 & 49.5 & 71.3 & 60.4 & 90.9 & 66.1 & 84.6 & 74.5 & {72.9} \\
    Models Genesis  & 68.5 & 86.8 & 45.8 & 68.0 & 50.9 & 89.3 & 60.6 & 81.6 & 69.1 & {69.0} & 72.3 & 87.5 & 48.7 & 71.0 & 64.7 & 91.7 & 69.9 & 87.9 & 76.9 & {74.5} \\
    Swin UNETR          & 66.8 & 86.5 & 41.6 & 66.2 & 46.9 & 85.6 & 58.3 & 76.3 & 67.7 & {66.2} & 71.2 & 87.1 & 47.4 & 69.3 & 60.1 & 90.0 & 67.2 & 84.2 & 73.7 & {72.3} \\
    SimCLR                 & 67.7 & 86.3 & 45.1 & 67.4 & 49.8 & 88.8 & 61.1 & 77.5 & 67.9 & {68.0} & 71.7 & 87.0 & 49.8 & 69.6 & 62.0 & 91.3 & 67.8 & 86.5 & 73.9 & {73.3} \\
    VoCo                       & 66.7 & 85.9 & 40.9 & 65.5 & 46.2 & 86.4 & 56.6 & 75.4 & 67.3 & {65.7} & 71.4 & 86.9 & 47.6 & 68.6 & 60.0 & 90.9 & 64.8 & 84.6 & 71.4 & {71.8} \\
    VF                      & 68.4 & 86.4 & 47.8 & 68.0 & 51.9 & 88.9 & 61.9 & 80.5 & 69.0 & {69.2} & 72.5 & 87.3 & 53.3 & 71.4 & 63.9 & 91.5 & 70.3 & 88.1 & 78.4 & {75.2} \\
    SimMIM                  & 68.1 & 87.0 & 47.1 & 68.3 & 49.8 & 90.6 & 64.2 & 82.1 & 70.9 & {69.8} & \textbf{72.6} & 88.0 & 51.0 & 71.1 & 63.1 & \textbf{92.7} & 71.9 & 90.1 & \textbf{79.8} & {75.6} \\
    MAE   & 70.2 & 87.1 & \textbf{52.0} & 70.0 & 54.9 & 90.4 & 67.9 & 83.9 & 71.7 & {72.0} & 72.4 & 87.8 & 51.3 & 73.0 & 64.7 & 92.5 & 71.7 & 89.3 & 77.9 & {75.6} \\
    \midrule
    \rowcolor{oursrow}
    \ours                                        & \textbf{71.1} & \textbf{87.6} & 50.8 & \textbf{72.8} & \textbf{57.3} & \textbf{91.2} & \textbf{68.3} & \textbf{86.1} & \textbf{72.7} & \textbf{73.1} & \textbf{72.6} & \textbf{88.2} & \textbf{55.8} & \textbf{77.5} & \textbf{67.5} & 92.5 & \textbf{76.4} & \textbf{89.8} & 79.5 & \textbf{77.8} \\
    \bottomrule
\end{tabular}

    }
\end{table*}
\begin{table*}[t]
    \centering
    \caption{Comparison across all 15 datasets, \ice frozen encoder and \fire fully fine-tuned. \textbf{Bold} metrics reflect best per dataset per evaluation regime. \ours remains competitive with larger pretrained models with broader pretraining datasets. \dev: developmental dataset.}
    \label{tab:large}
    \resizebox{\textwidth}{!}{%
        \begin{tabular}{l c cccccc ccccc ccccccc c}
    \toprule
    & & \multicolumn{6}{c}{{Brain MRI}} & \multicolumn{5}{c}{{Body MRI}} & \multicolumn{7}{c}{{CT}} & \\
    \cmidrule(lr){3-8}\cmidrule(lr){9-13}\cmidrule(lr){14-20}
    Method & Regime & BrTu & Hipp & YBM & ISLES\dev & GLI & Avg. & Heart & Pros & ACDC\dev & P158\dev & Avg. & Liver & Lung & Panc & HVes & Spl & Colon & Avg. & All \\
    \midrule
    Random init.                    & \ice  & 67.9 & 85.7 & 41.7 & 65.6 & 50.8 & {62.4} & 71.8 & 58.5 & 68.8 & 64.2 & {65.8} & 46.4 & 18.1 & 12.7 & 37.6 & 48.9 & 7.2 & {28.5} & {49.7} \\
                                    & \fire & 71.8 & 87.4 & 49.5 & 71.3 & 60.4 & {68.1} & 90.9 & 66.1 & 84.6 & 74.5 & {79.0} & 64.9 & 36.0 & 44.3 & 60.2 & 90.6 & 19.0 & {52.5} & {64.8} \\
    Slice-Ref.                      & \ice   & 68.8     & 86.2     & 45.5 & 68.4 & 52.1 & 64.2     & 89.5      & 67.0     & 79.9 & 70.8 & 76.8    & 58.5      & 45.2     & 32.5         & 50.0     & 71.0       & 16.3      & 45.6  & 60.1   \\
                                    & \fire  & 66.0     & 83.8     & 38.8 & 66.5 & 47.7 & 60.5     & 90.3      & 67.9     & 85.9 & 76.0 & 80.0    & 56.9      & 30.9     & 21.3         & 46.2     & 83.3       & 6.5      & 40.9  & 57.9   \\
    TriPlane Ref.                   & \ice   & 69.5     & 86.5     & 49.2 & 69.8 & 53.2 & 65.6     & 90.3      & \textbf{71.8}     & 84.1 & \textbf{75.0} & \textbf{80.3}    & 61.1      & 44.0     & 37.8         & 54.8     & 69.6       & 24.2      & 48.6  & 62.7   \\
                                    & \fire  & 67.6     & 84.2     & 41.4 & 70.3 & 51.6 & 63.0     & 90.9      & 72.6     & 89.1 & 80.5 & 83.3    & 61.5      & 38.8     & 30.5         & 52.7     & 82.0       & 10.2      & 45.9  & 61.6  \\
    \midrule
    \rowcolor{oursrow}
    \pcours                         & \ice  & 69.0 & 86.7 & 48.3 & 70.0 & 52.0 & {65.2} & 88.0 & 65.1 & 83.4 & 71.8 & {77.1} & \textbf{69.1} & 56.6 & \textbf{48.5} & \textbf{61.5} & 91.6 & 33.5 & {60.1} & {66.3} \\
    \rowcolor{oursrow}
    (22k vols.)                     & \fire & 72.5 & 87.9 & 50.9 & 75.8 & 64.1 & {70.3} & 86.3 & 72.9 & 89.2 & 79.2 & {81.9} & \textbf{77.0} & 59.0 & 59.5 & 65.9 & \textbf{96.2} & 41.7 & \textbf{66.6} & {71.9} \\
    \rowcolor{oursrow}
    \omours                         & \ice  & \textbf{71.1} & \textbf{87.6} & 50.8 & \textbf{72.8} & \textbf{57.3} & {67.9} & 91.2 & 68.3 & \textbf{86.1} & 72.7 & {79.6} & 67.3 & 58.3 & 47.5 & 60.2 & \textbf{92.2} & \textbf{35.8} & \textbf{60.2} & \textbf{68.0} \\
    \rowcolor{oursrow}
    (120k vols.)                    & \fire & 72.6 & \textbf{88.2} & \textbf{55.8} & \textbf{77.5} & \textbf{67.5} & \textbf{72.3} & 92.5 & \textbf{76.4} & 89.8 & 79.5 & {84.5} & 76.3 & 57.0 & 61.2 & \textbf{66.0} & 94.7 & \textbf{42.2} & {66.2} & \textbf{73.1} \\
    MAE                             & \ice  & 70.2 & 87.1 & 52.0 & 70.0 & 54.9 & {66.8} & 90.4 & 67.9 & 83.9 & 71.7 & {78.5} & 56.5 & 46.9 & 41.0 & 51.2 & 74.1 & 21.1 & {48.5} & {62.6} \\
    (120k vols.)                    & \fire & 72.4 & 87.8 & 51.3 & 73.0 & 64.7 & {69.8} & 92.5 & 71.7 & 89.3 & 77.9 & {82.9} & 69.7 & 56.1 & 48.9 & 61.3 & 90.9 & 26.8 & {58.9} & {69.0} \\
    SimMIM                          & \ice  & 68.1 & 87.0 & 47.1 & 68.3 & 49.8 & {64.1} & 90.6 & 64.2 & 82.1 & 70.9 & 77.0 & 54.7 & 48.0 & 37.7 & 50.1 & 60.3 & 19.6 & {45.1} & {59.9} \\
    (120k vols.)                    & \fire & 72.6 & 88.0 & 51.0 & 71.1 & 63.1 & {69.2} & \textbf{92.7} & 71.9 & 90.1 & 79.8 & {83.6} & 65.7 & 53.3 & 51.3 & 62.3 & 82.8 & 22.0 & {56.2} & {67.8} \\
    3DINO                           & \ice  & 68.9 & 86.2 & 41.9 & 68.2 & 50.4 & {63.1} & 89.6 & 63.3 & 81.7 & 70.0 & {76.2} & 57.3 & 53.1 & 33.1 & 52.4 & 81.6 & 10.9 & {48.1} & {60.6} \\
    (100k vols.)                    & \fire & 69.7 & 86.5 & 41.7 & 67.3 & 54.5 & {63.9} & 90.0 & 65.3 & 83.9 & 67.2 & {76.6} & 55.3 & 13.7 & 17.0 & 46.8 & 69.7 & 8.5 & {35.2} & {55.8} \\
    nnFoundation                    & \ice  & \textbf{71.1} & \textbf{87.6} & \textbf{54.9} & 72.3 & 56.9 & \textbf{68.5} & \textbf{91.5} & {70.8} & 85.5 & {73.4} & \textbf{80.3} & 61.7 & \textbf{62.6} & 47.5 & 54.7 & 80.2 & 27.3 & {55.7} & {66.5} \\
    (2.1M vols.)                    & \fire & \textbf{72.8} & 88.1 & 53.9 & 73.6 & 67.1 & {71.1} & 92.5 & 76.2 & \textbf{90.4} & \textbf{81.1} & \textbf{85.1} & 71.6 & \textbf{62.4} & \textbf{62.1} & 64.1 & 93.3 & 39.1 & {65.4} & {72.6} \\
    \bottomrule
\end{tabular}

    }
\end{table*}

\noindent\textbf{Segmentation datasets.}
We evaluate on 15 datasets: all ten tasks of the Medical Segmentation Decathlon (MSD)~\cite{antonelli2022msd}, three further brain MRI datasets, namely Yale Brain Metastases (YBM)~\cite{ybm}, ISLES 2022 (ISLES)~\cite{hernandez2022isles}, and BraTS glioblastoma (GLI)~\cite{brats}, and two body MRI datasets: cardiac cine MRI from ACDC~\cite{bernard2018acdc} and prostate MRI from Prostate158 (P158)~\cite{adams2022p158}.
Following previous works~\cite{wald2025openmind,wald2025revisitingmae,nnfoundation}, we split each dataset into equal training and validation halves and report the Dice similarity coefficient (DSC) averaged over foreground classes on the validation half, with the same split for all methods (See Supp.~\labelcref{supp:data,supp:eval}).

\noindent\textbf{Global tasks.}
Classification comprises ABIDE~\cite{abide}, COVID-19 and severity prediction on STOIC~\cite{revelStudyThoracicCT2021}, MGMT status prediction on BraTS21~\cite{mgmt}, and breast lesion classification on ODELIA~\cite{odelia}.
For regression, we predict brain age on IXI~\cite{IXIDataset}.
For retrieval, we use cancer-status and lesion group retrieval on the liver, lung, pancreas, and colon subsets of 3D-MIR~\cite{mir3d}.
All encoders remain frozen, and each volume is processed at the encoder's native input resolution for these evaluations.
We report AUROC for binary classification, macro AUROC for ODELIA, RMSE in years for regression, and mean average precision (mAP) for retrieval.
All tasks and additional metrics are provided in \cref{supp:cls_reg_ret}.

\noindent\textbf{Developmental datasets.} ACDC, P158, and ISLES are our developmental datasets, marked with \dev. All ablations, hyper-parameter studies, and design decisions use only these three, under frozen evaluation, and the remaining datasets are held out from these decisions.

\noindent\textbf{Encoders and baselines.}
\ours uses Primus-S~\cite{wald2025primus} with NeCo~\cite{pariza2025neco} on DINOv2-R~\cite{darcet2024registers} (ViT-S/14) as the default reference.
The OpenMind baselines use Primus-M (156M) pretrained with MAE~\cite{wald2025revisitingmae}, SimMIM~\cite{xie2022simmim}, SimCLR~\cite{chen2020simclr}, VoCo~\cite{wu2024voco}, Models Genesis~\cite{zhou2019modelsgenesis}, Volume Fusion~\cite{wang2023misfm}, and Swin UNETR objectives~\cite{tang2022swinunetr} and evaluated on nine MRI datasets.
We additionally evaluate MAE, SimMIM, nnFoundation (ViT)~\cite{nnfoundation}, a 551M-parameter Primus encoder pretrained using MAE, 3DINO~\cite{3dino}, a 308M-parameter ViT-L pretrained with a DINOv2 objective, and the reference model under a slice wise segmentation~\cite{medAdapt3d} (Slice-Ref.) and using our tri-plane feature extraction (TriPlane-Ref.) to gauge the necessity of distilling beyond reducing computational burden and ease of use, on all 15 datasets.
A randomly initialized Primus-M is evaluated frozen (mean of three runs) and trained from scratch.
Architecture details and compute burden are given in Supp.~\labelcref{supp:eval,supp:burden}.

\noindent\textbf{Segmentation protocol.}
Every encoder uses a UNETR decoder~\cite{hatamizadeh2022unetr} within nnU-Net~\cite{isensee2021nnunet}, sharing input spacing and resolution, intensity normalization, augmentations, and training schedules across methods.
We train only the decoder in \ice frozen-encoder evaluation and all parameters in \fire full fine-tuning, for 150 epochs in either regime.
For inputs with multiple modalities, we replicate the pretrained patch-embedding weights across channels and divide them by the number of channels~\cite{wald2025openmind,wald2025revisitingmae,nnfoundation}.

\subsection{Results}\label{sec:result}

\noindent\textbf{Same-data medical SSL comparison.}
\Cref{tab:main} compares \ours with seven SSL methods pretrained on the same OpenMind volumes across nine MRI segmentation datasets.
\ours achieves the highest mean DSC with the encoder frozen (73.1 vs.\ 72.0 for MAE) and after full fine-tuning (77.8 vs.\ 75.6 for MAE and SimMIM).
The frozen improvement over MAE is consistent across the two MRI groups, from 66.8 to 67.9 DSC on brain MRI and from 78.5 to 79.6 on body MRI.
These results show that the advantage of \ours is already present in the pretrained representation and remains after task-specific adaptation.

\noindent\textbf{Comparison with broader pretrained encoders.}\label{sec:frozen}
\Cref{tab:large} extends the segmentation comparison to all 15 datasets and includes nnFoundation and 3DINO, whose encoders are substantially larger and are pretrained on broader medical corpora.
\omours obtains the highest overall mean DSC in both evaluation regimes: 68.0 compared with 66.5 for nnFoundation when frozen, and 73.1 compared with 72.6 after full fine-tuning.
Significance and bootstrapping results are shown in Supp.~\ref{supp:ranking_reloaded}.
The fine-tuning gap of \omours and \pcours is 5.1 and 5.6 points, respectively, compared with 6.4 for MAE and 6.1 for nnFoundation.
Thus, the compact \ours encoders provide strong features before adaptation while retaining competitive performance after optimizing the entire model.

\noindent\textbf{\ours vs Tri-plane vs Slice-based.} Applying the reference model per axial slice is on average worse than using tri-plane features (+2.6) and distilling into a 3D encoder is better than using tri-plane features directly (+5.2), both on OpenMind frozen evaluation. Therefore, \ours inherits better volumetric features that go beyond fused planar contexts, rather than just being easier to use and having a smaller compute burden (Supp.~\ref{supp:burden}).

\noindent\textbf{Out-of-domain transfer.}
The largest advantage over strong OpenMind SSL baselines appears when transferring to CT.
\omours, reaches 60.2 DSC across the six CT datasets, similar to the CT-pretrained \pcours at 60.1 and above nnFoundation at 55.7, while improving over the OpenMind MAE baseline by 11.7 points.
Its frozen CT performance lies within 6.0 points of its fully fine-tuned result, compared with gaps of 9.7--11.1 for nnFoundation, MAE, and SimMIM.
Transfer also occurs in the opposite direction as \pcours, pretrained on CT, reaches 70.5 DSC across the nine MRI datasets with its encoder frozen, exceeding $6 / 7$ SSL methods pretrained on OpenMind.

\noindent\textbf{Global tasks.}
\Cref{tab:benchmark_comparison} evaluates the frozen global representations beyond segmentation on classification, brain-age regression, and cancer-status retrieval.
Across the ten metrics, \ours is always competitive.
Notably, these results are obtained without adapting the encoder, indicating that the distilled representation supports both dense prediction and volume-level downstream tasks.
Complete results, including additional metrics and datasets, are reported in Supp.~\ref{supp:cls_reg_ret}.
\begin{table*}[t]
    \centering
    \caption{
        \textbf{Frozen global-task evaluation.}
        All encoders are frozen and run at native pretrained resolution.
        Higher is better except for RMSE.
        Best results are shown in \textbf{bold} and second-best results are \underline{underlined}.
        \ours is competitive with other methods.
    }
    \label{tab:benchmark_comparison}
    \resizebox{\linewidth}{!}{
        \begin{tabular}{lcccccccccc}
    \toprule
    & \multicolumn{5}{c}{Classification}
    & \multicolumn{1}{c}{Regression}
    & \multicolumn{4}{c}{Retrieval} \\
    \cmidrule(lr){2-6}
    \cmidrule(lr){7-7}
    \cmidrule(lr){8-11}
    Method
    & ABIDE
    & COVID
    & COVID Severity
    & BraTS21 MGMT
    & ODELIA
    & Brain Age (IXI)
    & Liver Cancer
    & Lung Cancer
    & Pancreas Cancer
    & Colon Cancer \\
    & AUROC $\uparrow$
    & AUROC $\uparrow$
    & AUROC $\uparrow$
    & AUROC $\uparrow$
    & Macro AUROC $\uparrow$
    & RMSE (years) $\downarrow$
    & mAP $\uparrow$
    & mAP $\uparrow$
    & mAP $\uparrow$
    & mAP $\uparrow$ \\
    \midrule

    \rowcolor{oursrow}
    \pcours
    & \textbf{0.674}
    & \underline{0.758}
    & \textbf{0.761}
    & 0.571
    & 0.745
    & 7.348
    & \underline{0.732}
    & \textbf{0.886}
    & \textbf{0.950}
    & \textbf{0.729} \\

    \rowcolor{oursrow}
    \omours
    & \underline{0.655}
    & \textbf{0.762}
    & \underline{0.740}
    & \textbf{0.588}
    & 0.744
    & \textbf{6.994}
    & \textbf{0.742}
    & 0.834
    & \underline{0.941}
    & \underline{0.721} \\

    MAE
    & 0.653
    & 0.717
    & 0.727
    & 0.532
    & \textbf{0.761}
    & 8.988
    & 0.711
    & 0.761
    & 0.932
    & 0.674 \\

    SimMIM
    & 0.641
    & 0.701
    & 0.715
    & \underline{0.586}
    & 0.744
    & 10.709
    & 0.676
    & 0.720
    & 0.922
    & 0.657 \\

    SimCLR
    & 0.635
    & 0.735
    & 0.715
    & 0.559
    & \underline{0.750}
    & 9.222
    & 0.660
    & 0.707
    & 0.902
    & 0.655 \\

    VoCo
    & 0.599
    & 0.714
    & 0.703
    & 0.482
    & 0.693
    & 11.716
    & 0.675
    & 0.696
    & 0.894
    & 0.612 \\

    3DINO
    & 0.606
    & 0.652
    & 0.649
    & 0.516
    & 0.698
    & 12.839
    & 0.703
    & \underline{0.854}
    & 0.930
    & 0.687 \\

    nnFoundation
    & 0.617
    & 0.712
    & 0.700
    & 0.550
    & 0.738
    & \underline{7.344}
    & 0.690
    & 0.819
    & \underline{0.941}
    & 0.685 \\
    \bottomrule
\end{tabular}
    }
\end{table*}

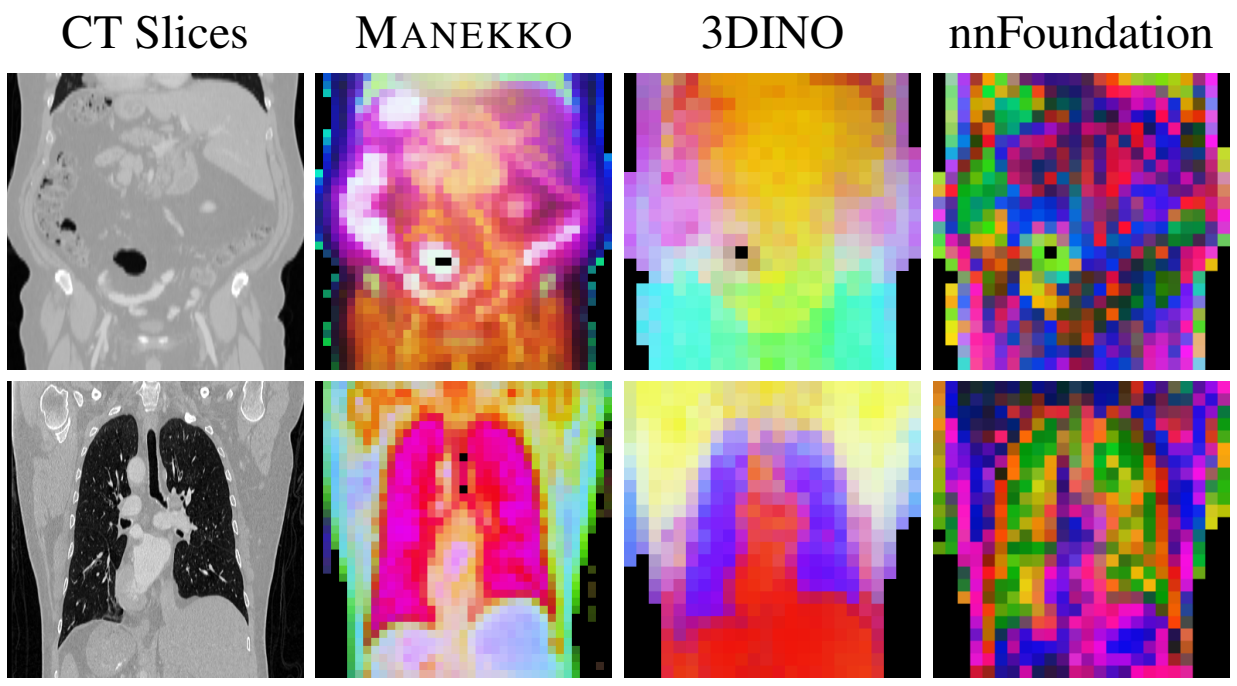
\begin{figure}[t]
    \centering
    \begin{tikzpicture}[lab/.style={font=\small}]
    \node[lab, anchor=south] at (0.000\linewidth, 0.122\linewidth) {CT Slices};
    \node[lab, anchor=south] at (0.245\linewidth, 0.122\linewidth) {\ours};
    \node[lab, anchor=south] at (0.490\linewidth, 0.122\linewidth) {3DINO};
    \node[lab, anchor=south] at (0.735\linewidth, 0.122\linewidth) {nnFoundation};
    \node[inner sep=0pt] at (0.000\linewidth, 0.000\linewidth) {\pngslot{0.235\linewidth}{0.235\linewidth}{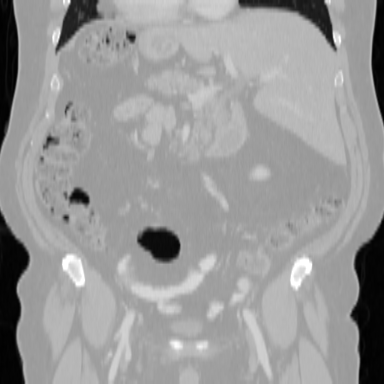}};
    \node[inner sep=0pt] at (0.245\linewidth, 0.000\linewidth) {\pngslot{0.235\linewidth}{0.235\linewidth}{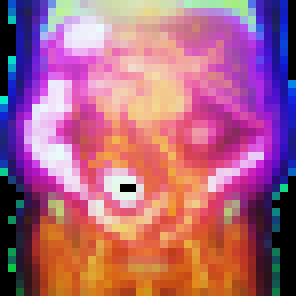}};
    \node[inner sep=0pt] at (0.490\linewidth, 0.000\linewidth) {\pngslot{0.235\linewidth}{0.235\linewidth}{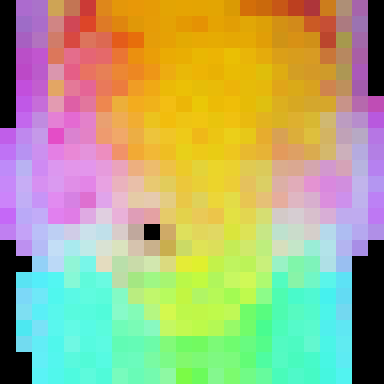}};
    \node[inner sep=0pt] at (0.735\linewidth, 0.000\linewidth) {\pngslot{0.235\linewidth}{0.235\linewidth}{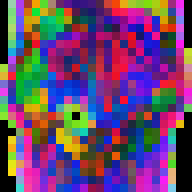}};
    \node[inner sep=0pt] at (0.000\linewidth, -0.245\linewidth) {\pngslot{0.235\linewidth}{0.235\linewidth}{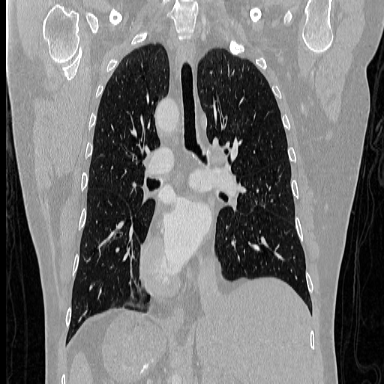}};
    \node[inner sep=0pt] at (0.245\linewidth, -0.245\linewidth) {\pngslot{0.235\linewidth}{0.235\linewidth}{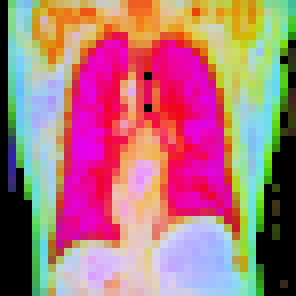}};
    \node[inner sep=0pt] at (0.490\linewidth, -0.245\linewidth) {\pngslot{0.235\linewidth}{0.235\linewidth}{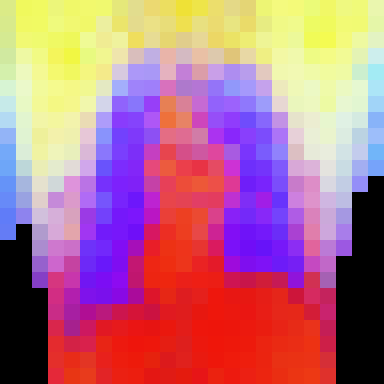}};
    \node[inner sep=0pt] at (0.735\linewidth, -0.245\linewidth) {\pngslot{0.235\linewidth}{0.235\linewidth}{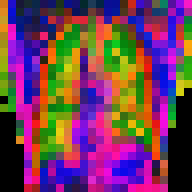}};
\end{tikzpicture}
    \caption{
        CT slices (first column) and the first three principal components of frozen patch features (remaining columns), shown as RGB for two volumes (rows).
        PCA is fitted separately per volume using foreground patches, with the background shown in black.
        Qualitatively, \ours projections exhibit fine spatial variation and coherent anatomical regions in these examples.
    }
    \vspace{-2ex}
    \label{fig:pca}
\end{figure}

\noindent\textbf{Patch representations.}
\Cref{fig:pca} visualizes frozen patch features on CT volumes for \ours, nnFoundation, and 3DINO.
The projections of \ours retain fine spatial variation and visually separate several anatomical regions, whereas others appear more homogeneous or less distinct.
PCA provides only a qualitative view of the representation; however, these examples are consistent with the stronger frozen dense-prediction performance of \ours.

\subsection{Ablations and Hyperparameters}\label{sec:ablation}
All experiments in this section pretrain \ours on OpenMind and report frozen-encoder DSC on the developmental datasets unless stated otherwise. See Supp.~\ref{supp:sens} for loss weighting and downstream training duration experiments.

\noindent\textbf{Choosing the reference model.}\label{sec:reference}
We pretrain \ours using seven reference models of matching dimensions: DINO~\cite{caron2021dino}, DINOv2~\cite{oquab2024dinov2}, DINOv2 with registers (DINOv2-R)~\cite{darcet2024registers}, NeCo applied to DINOv2-R and DINOv2-XR~\cite{pariza2025neco}, Leopart~\cite{ziegler2022leopart}, and CrIBo~\cite{lebailly2024cribo}.
Across these references, Hummingbird mIoU on Pascal VOC (\cref{sec:incontext}) is strongly associated with the student's mean frozen DSC on the three developmental datasets.
Thus, in this set of models, references with stronger natural-image in-context segmentation also produce stronger frozen medical students.
Because Hummingbird uses only natural-image annotations, we can evaluate it before medical pre-training and use it as a practical reference-selection criterion without medical labels.
Correlations with ADE20K and ImageNet accuracy are reported in Supp.~\ref{supp:reference}.
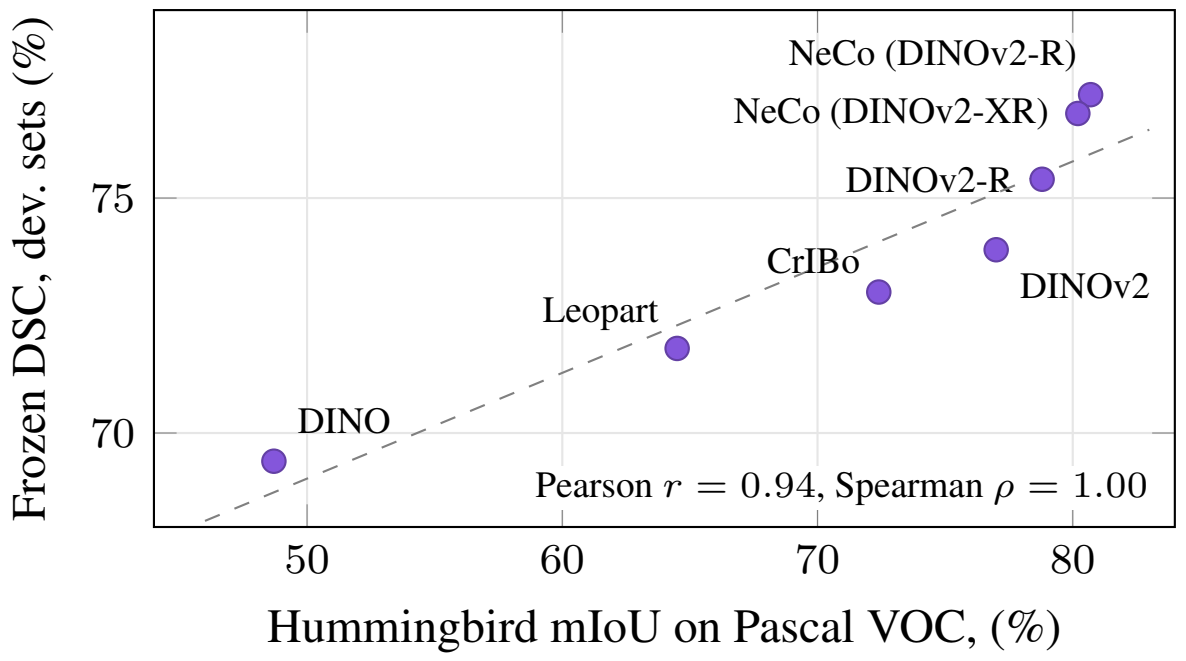
\begin{figure}[t]
    \centering
    \begin{tikzpicture}
    \begin{axis}[
        width=\linewidth, height=0.6\linewidth,
        xlabel={Hummingbird mIoU on Pascal VOC, (\%)},
        ylabel={Frozen DSC, dev.\ sets (\%)},
        xmin=44, xmax=84, ymin=68, ymax=79,
        grid=major, grid style={gray!20},
        tick label style={font=\footnotesize},
        label style={font=\small},
        nodes near coords, point meta=explicit symbolic,
        every node near coord/.append style={font=\scriptsize, anchor=south east, text=black},
    ]
        \addplot[only marks, mark=*, mark size=2.2pt, mark options={m om}, color=paarsD] coordinates {
            (64.5,71.8)  [Leopart]
            (72.4,73.0)  [CrIBo]
        };
        \addplot[only marks, mark=*, mark size=2.2pt, mark options={m om}, color=paarsD] coordinates {(77.0,73.9) (78.8,75.4) (80.7,77.2) (80.2,76.8)};
        \node[font=\scriptsize, anchor=north west] at (axis cs:77.2,73.8) {DINOv2};
        \node[font=\scriptsize, anchor=east] at (axis cs:78.4,75.4) {DINOv2-R};
        \node[font=\scriptsize, anchor=south east] at (axis cs:80.9,77.3) {NeCo (DINOv2-R)};
        \node[font=\scriptsize, anchor=east] at (axis cs:79.8,76.8) {NeCo (DINOv2-XR)};
        \addplot[only marks, mark=*, mark size=2.2pt, mark options={m om}, color=paarsD] coordinates {(48.7,69.4)};
        \node[font=\scriptsize, anchor=south west] at (axis cs:48.9,69.6) {DINO};
        \addplot[gray, dashed, domain=46:83] {0.225*x + 57.78};
        \node[anchor=south east, font=\scriptsize, fill=white, inner sep=1pt] at (rel axis cs:0.98,0.03) {Pearson $r=0.94$, Spearman $\rho=1.00$};
    \end{axis}
\end{tikzpicture}
    \caption{
        Natural-image in-context segmentation and student performance across seven reference models.
        Each point pairs the reference's Hummingbird mIoU on Pascal VOC (full memory bank) with the corresponding \ours student's mean frozen DSC on the three developmental datasets.
    }
    \vspace{-2ex}
    \label{fig:reference}
\end{figure}

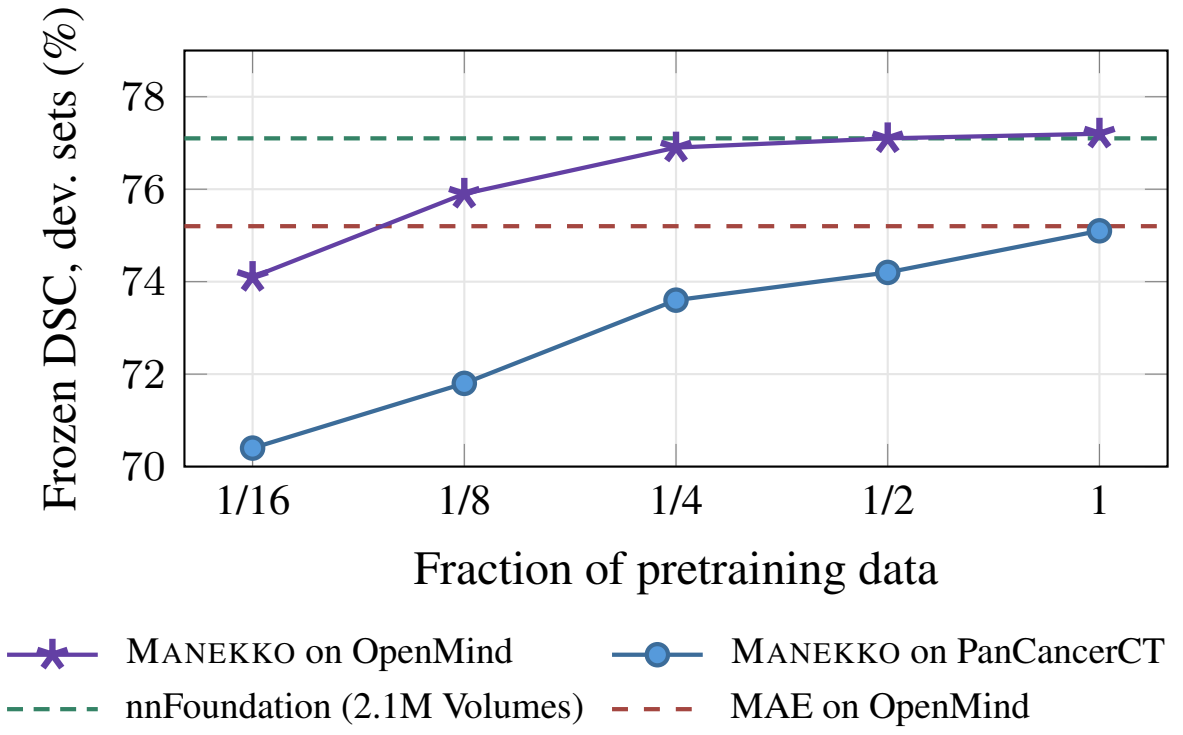
\begin{figure}[t]
    \centering
    \begin{tikzpicture}
    \begin{axis}[
        name=scal,
        width=0.97\linewidth, height=0.52\linewidth,
        xmode=log, log basis x=2,
        xtick={0.0625,0.125,0.25,0.5,1}, xticklabels={1/16,1/8,1/4,1/2,1},
        xmin=0.05, xmax=1.25,
        ymin=70, ymax=79,
        xlabel={Fraction of pretraining data},
        ylabel={Frozen DSC, dev.\ sets (\%)},
        grid=major, grid style={gray!20},
        tick label style={font=\footnotesize},
        label style={font=\small},
    ]
        \addplot[groenD, densely dashed, thick, domain=0.05:1.25] {77.1};
        \addplot[rodeD, dashed, thick, domain=0.05:1.25] {75.2};
        \addplot[paarsD, thick, mark=star, mark size=3pt, mark options={very thick}] coordinates {(0.0625,74.1) (0.125,75.9) (0.25,76.9) (0.5,77.1) (1,77.2)};
        \addplot[blauwD, thick, mark=*, mark size=2pt, mark options={m pc}] coordinates {(0.0625,70.4) (0.125,71.8) (0.25,73.6) (0.5,74.2) (1,75.1)};
    \end{axis}
\end{tikzpicture}

\begin{tikzpicture}
    \begin{customlegend}[legend columns=2, legend style={font=\scriptsize, draw=none, column sep=1mm, at={(scal.north)}, anchor=south}, legend cell align=left,
        legend entries={{\ours on OpenMind}, {\ours on PanCancerCT}, {nnFoundation (2.1M Volumes)}, {MAE on OpenMind}}]
        \addlegendimage{paarsD, thick, mark=star, mark size=3pt, mark options={very thick}}
        \addlegendimage{blauwD, thick, mark=*, mark size=2pt, mark options={m pc}}
        \addlegendimage{groenD, densely dashed, thick}
        \addlegendimage{rodeD, dashed, thick}
    \end{customlegend}
\end{tikzpicture}
    \caption{
        Frozen DSC trend on the three dev. datasets after pretraining \ours on random subsets of OpenMind and PanCancerCT\@.
    }
    \label{fig:scaling}
\end{figure}

\noindent\textbf{Data scaling.}
We pretrain on random subsets of each corpus and evaluate frozen segmentation on the developmental datasets (\cref{fig:scaling}).
Using one quarter of OpenMind yields 76.9 DSC, within 0.3 points of the full-data result, and one eighth reaches 75.9, above MAE pre-trained on all of OpenMind (75.2).
With the full corpus, \omours matches nnFoundation (77.2 vs.\ 77.1) using about $18\times$ fewer medical volumes, while \pcours, pre-trained on 22k CT volumes, matches OpenMind MAE (75.1 vs.\ 75.2).
OpenMind performance changes little beyond one quarter of the corpus, whereas PanCancerCT performance increases throughout the tested range, from 70.4 to 75.1 with all volumes.
The results demonstrate medical data efficiency and the benefit of data variation, but show that large-scale datasets like OpenMind still outperform smaller datasets despite being single-anatomy.

\noindent\textbf{Component Ablations.}
\Cref{tab:ablation} ablates the components of \ours.
Tri-plane lifting and feature normalization bring the largest gains.
Regressing the features of all three planes instead of the axial plane only improves the average from 70.7 to 73.0 DSC.
Normalizing the target with statistics of the whole volume adds 2.4 DSC, and estimating them on the foreground adds a further 1.4 DSC, consistent with background dominating the statistics of a medical crop.
The affinity loss further boosts the DSC by $+0.4$ without normalization and further $+0.3$ with it, and it gives the student a meaningful global token.
Without it, the CLS token of the student is not tied to its patch features and shares affinity mostly with the background, and with it vice versa (\cref{fig:affinity_fg}).
\begin{table}[t]
    \centering
    \caption{
    Component ablations.
    The first row regresses features only on the axial axes.
    ``Norm.'' without ``Fg.\ mask'' estimates the statistics over the whole volume.
    }
    \label{tab:ablation}
    \small
\setlength{\tabcolsep}{2.75pt}
\begin{tabular}{c c c c cccc}
    \toprule
    Tri. & Norm. & Fg.\ mask & Affinity & ISLES & ACDC & P158 & Avg. \\
    \midrule
    \xmark & \xmark & \xmark & \xmark & 65.0 & 79.6 & 67.6 & 70.7 \\ 
    \midrule
    \cmark & \xmark & \xmark & \xmark & 69.8 & 81.2 & 68.0 & 73.0 \\
    \cmark & \xmark & \xmark & \cmark & 70.1 & 81.8 & 68.4 & 73.4 \\
    \cmark & \cmark & \xmark & \cmark & 71.4 & 84.2 & 71.8 & 75.8 \\
    \cmark & \cmark & \cmark & \xmark & 72.4 & 86.0 & 72.3 & 76.9 \\
    \rowcolor{oursrow}
    \cmark & \cmark & \cmark & \cmark & 72.8 & 86.1 & 72.7 & \textbf{77.2} \\
    \bottomrule
\end{tabular}

\end{table}

\noindent\textbf{Decoder.}
\Cref{tab:sensitivity} compares the default UNETR decoder choice to Patch Decode~\cite{wald2025primus}, which decodes the last layer patch features.
UNETR is better by 21 DSC points, which we attribute to its use of features from several encoder depths therefore being able to utilize the multi-scale details that is lost by the last layer alone. 
UNETR decoder was used as default for all methods.

\noindent\textbf{Crop size and scale.}
The reference resolution scales with the student lattice (Eq.~\ref{eq:active_resolution}), a larger crop also means the reference model is evaluated at a higher resolution, where its features are more detailed~\cite{oquab2024dinov2}. However, a larger pretraining crop helps up to $192^3$, with benefit only to ISLES which is isotropic at $296^3$.
For random-resized cropping, a scale of $0.4^3$ works best, and less scale variation lowers performance.

\noindent\textbf{Batch size.}
Larger batches help (\cref{tab:sensitivity}): frozen DSC improves by 1.1 points from 8 to 64 (76.1 vs.\ 77.2), with smaller gains beyond 32.
This can be explained by normalization statistics, which are estimated from each batche during training and then fixed. Small batches give noisy estimates of the foreground statistics, which then define the target for the remainder of training. 
However, our method can make use of gradient accumulation steps as the loss does not rely on intra-batch contrast or entropy like VoCo, SimCLR and DINO like objectives, therefore circumventing multi-GPU parallelisation requirement if it is unavailable.

\begin{figure}[t]
    \centering
    \begin{tikzpicture}[lab/.style={font=\scriptsize}]
    \node[lab, anchor=south] at (0.000\linewidth, 0.160\linewidth) {Volume};
    \node[lab, anchor=south] at (0.330\linewidth, 0.160\linewidth) {w/o $\mathcal{L}_{\mathrm{aff}}$};
    \node[lab, anchor=south] at (0.660\linewidth, 0.160\linewidth) {with $\mathcal{L}_{\mathrm{aff}}$};
    \node[inner sep=0pt] at (0.000\linewidth, 0.000\linewidth) {\pngslot{0.315\linewidth}{0.315\linewidth}{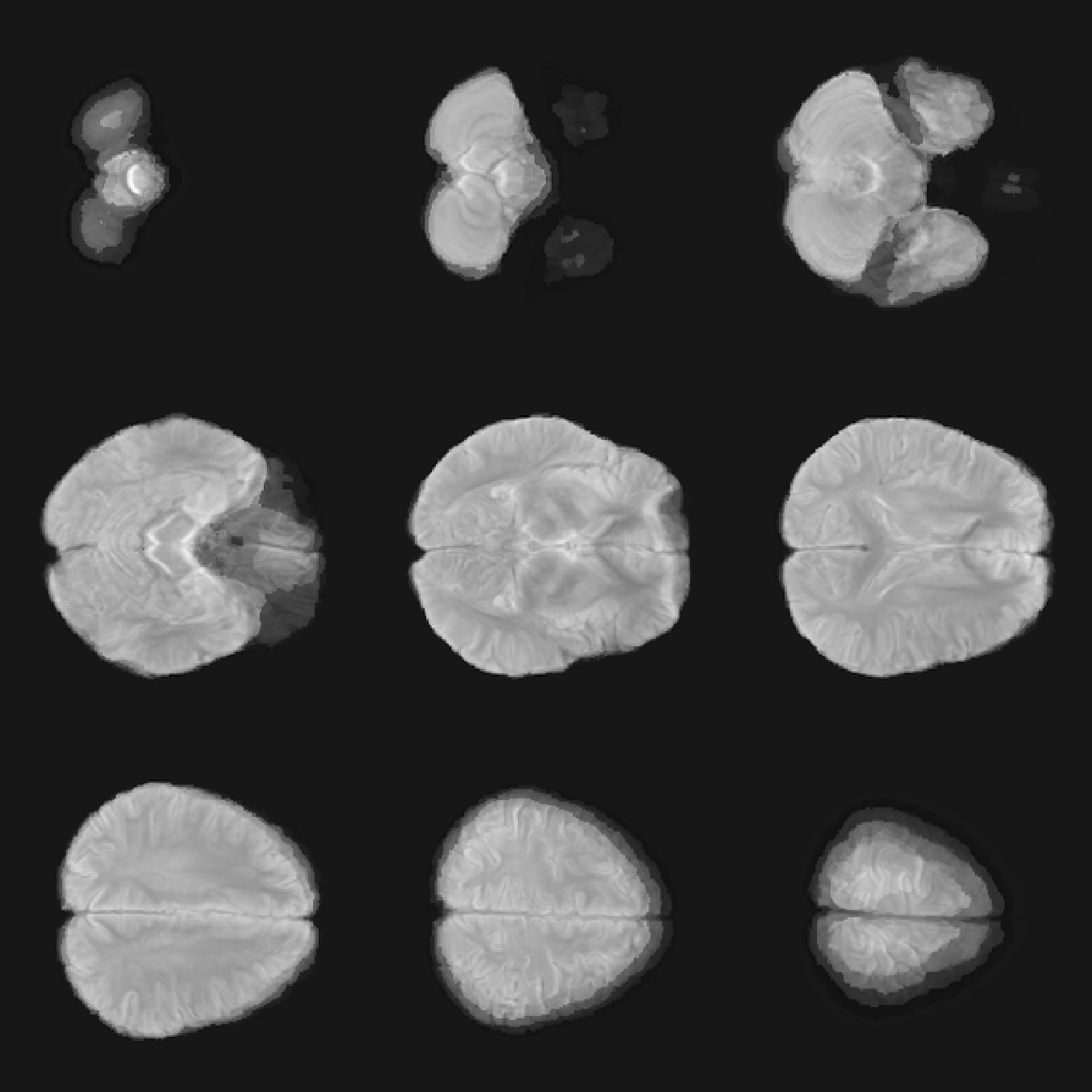}};
    \node[inner sep=0pt] at (0.330\linewidth, 0.000\linewidth) {\pngslot{0.315\linewidth}{0.315\linewidth}{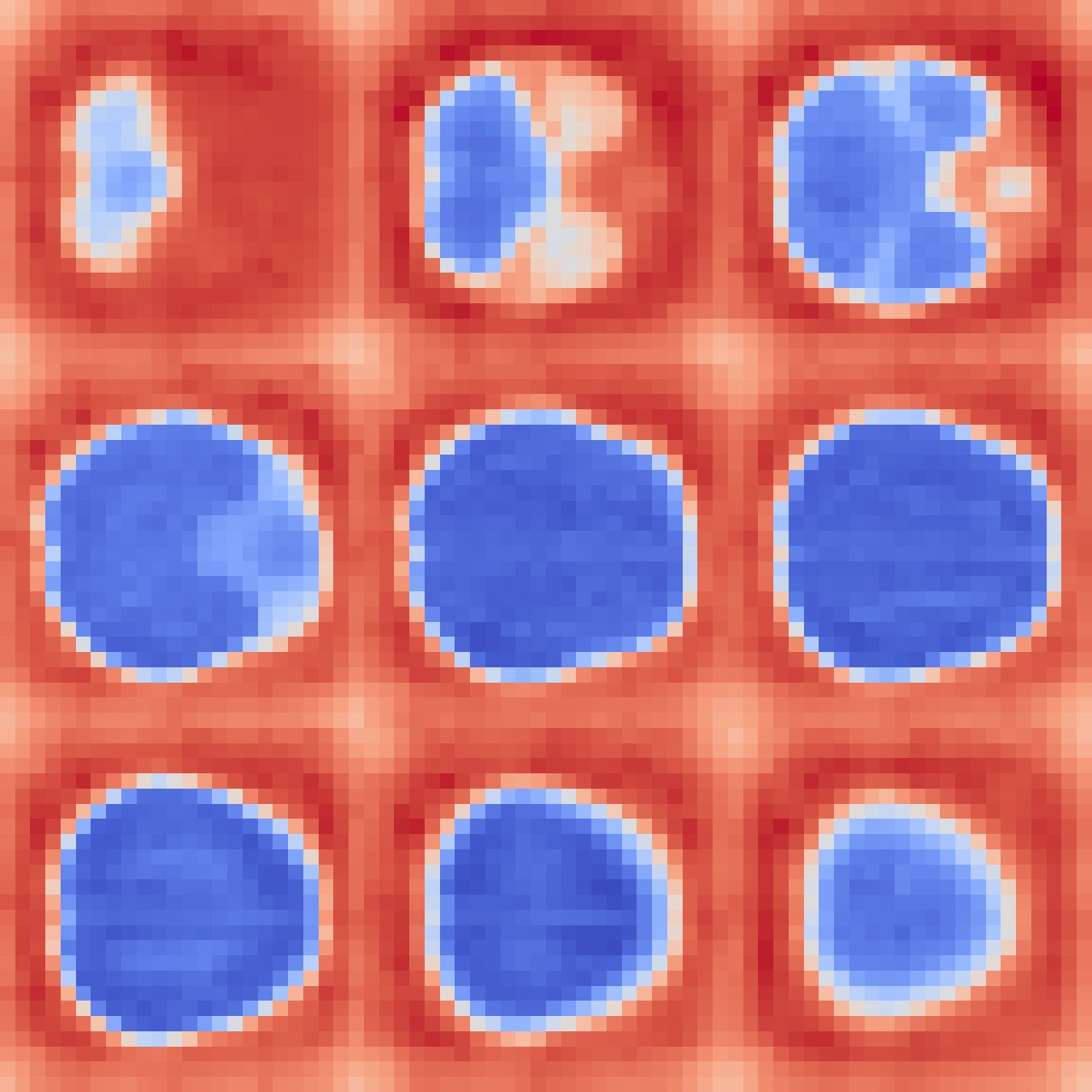}};
    \node[inner sep=0pt] at (0.660\linewidth, 0.000\linewidth) {\pngslot{0.315\linewidth}{0.315\linewidth}{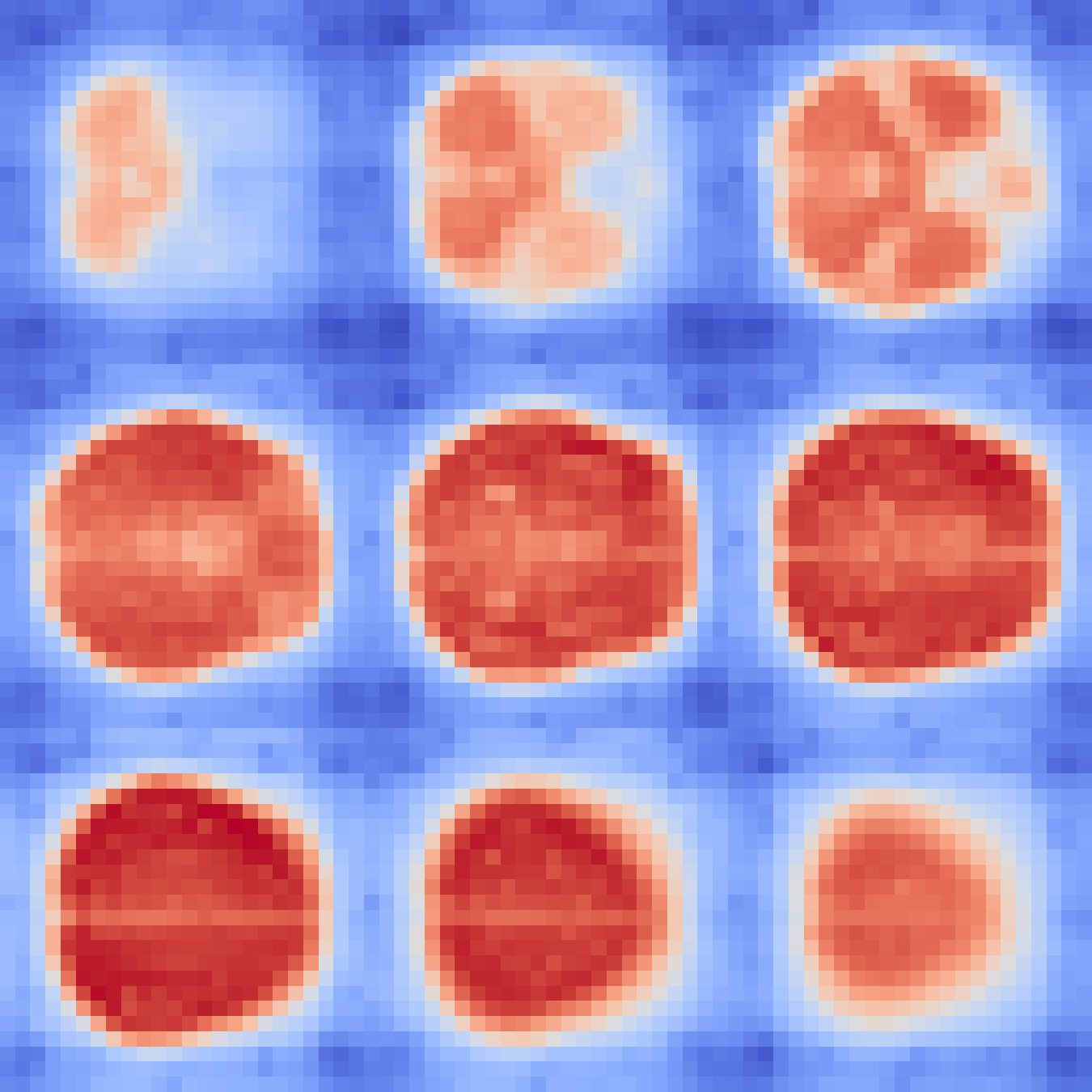}};
\end{tikzpicture}
    \caption{Affinity denoted as heatmap from blue (low) to high (red) between the CLS and the patch tokens, without and with $\mathcal{L}_{\mathrm{aff}}$.
    Without the affinity loss, the CLS token attends mostly to the background, with it, the CLS token attends to the foreground anatomy (brain). Therefore, using $\mathcal{L}_{\mathrm{aff}}$ corrects undesirable behaviour.
    }
    \label{fig:affinity_fg}
\end{figure}
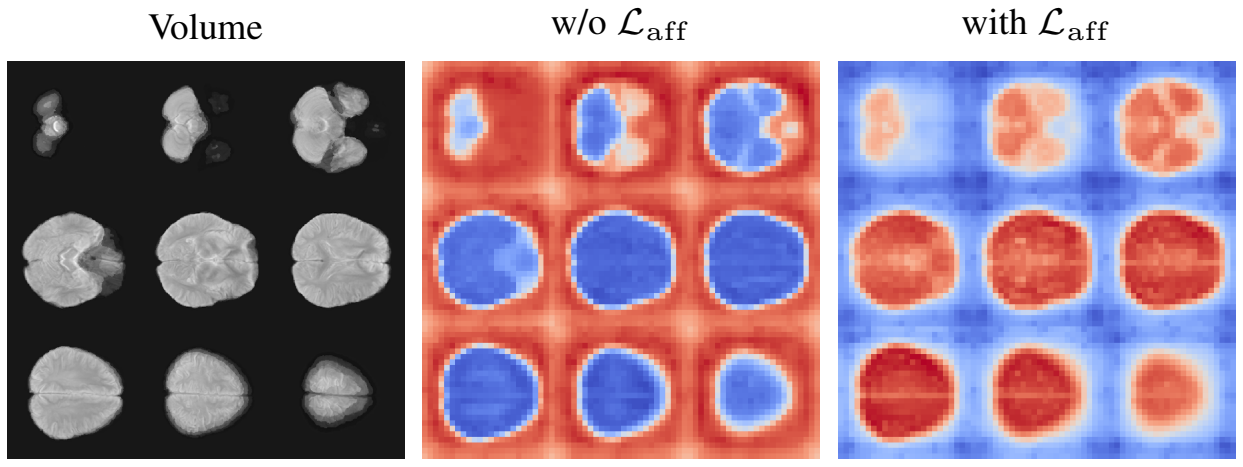
\begin{table}[t]
    \centering
    \caption{Decoder and pre-training hyper-parameters sweep, with frozen DSC of \ours on the developmental datasets.}
    \label{tab:sensitivity}
    \resizebox{\linewidth}{!}{
    \small
\begin{tabular}{l l cccc}
    \toprule
    Setting & Value & ISLES & ACDC & P158 & Avg. \\
    \midrule
    \multirow{2}{*}{Decoder} & Patch Decode (256K)      & 33.7 & 70.4 & 63.4 & 55.8 \\
                             & UNETR (6.5M)             & 72.8 & 86.1 & 72.7 & \textbf{77.2} \\
    \midrule
    \multirow{4}{*}{Crop size} & $96^3$  & 71.3 & 85.2 & 71.9 & 76.1 \\
                               & $128^3$ & 72.6 & 85.8 & 72.4 & 76.9 \\
                               & $192^3$ & 72.8 & 86.1 & 72.7 & \textbf{77.2} \\
                               & $296^3$ & 73.1 & 86.0 & 72.3 & 77.1 \\
    \midrule
    \multirow{3}{*}{Crop scale}& $0.4^3$ & 72.8 & 86.1 & 72.7 & \textbf{77.2} \\
                               & $0.6^3$ & 72.4 & 85.8 & 72.3 & 76.8 \\
                               & $0.8^3$ & 69.2 & 84.4 & 71.8 & 75.1 \\
    \midrule
    \multirow{4}{*}{Batch size} & 8  & 71.8 & 85.2 & 71.2 & 76.1 \\
                                & 16 & 72.0 & 85.7 & 71.9 & 76.5 \\
                                & 32 & 72.6 & 86.0 & 72.4 & 77.0 \\
                                & 64 & 72.8 & 86.1 & 72.7 & \textbf{77.2} \\
    \bottomrule
\end{tabular}

    }
\end{table}

\section{Discussion and Conclusion}\label{sec:discussion}\label{sec:conc}
We investigated whether natural-vision model representations provide an effective training signal for medical SSL.
\ours outperforms medical SSL baselines trained on the same data on average and remains competitive with substantially larger encoders pre-trained on broader medical corpora, both with frozen features and after full fine-tuning.
We find it improves beyond directly using the reference model on single slice or after tri-plane construction, resulting in an easy to use, easy to fine-tune and more efficient to run 3D encoder.
Its frozen representations transfer well to classification, regression, and retrieval, while natural-image in-context segmentation provides a strong criterion for selecting the reference without medical labels.
\noindent Our main limitation are reliance on the reference model, training on publicly accessible medical datasets limiting our ability to scale, requirement of patch feature grid resolution absent in CNNs and using larger encoders as reference models was intractable for us.
We see incorporating domain priors, like pathology, as an important step for future work.
Our results suggest that natural-image model representations can provide effective supervision for training small but foundational 3D medical models.

\clearpage
{
    \small
    \section*{Acknowledgements}
    \noindent This publication is part of the project ROBUST: Trustworthy AI-based Systems for Sustainable Growth with project number KICH3.LTP.20.006, which is (partly) financed by the Dutch Research Council (NWO), Siemens Healthineers, and the Dutch Ministry of Economic Affairs and Climate Policy (EZK) under the program LTP KIC 2020--2023.
    
    \section*{LLM Usage Statement}
    During the preparation of this work, the authors used a large language model, ChatGPT, to suggest improvements to phrasing, and Claude to generate code for making axonometric perspective transformations for raster images in method schema, boilerplate for \texttt{pgfplots} syntax and only debugging program code at various stage of developments. All AI-assisted text and code were subsequently reviewed, tested, and edited by the authors, who take full responsibility for the final content.
    
    \bibliographystyle{ieeenat_fullname}
    \bibliography{main}

\begin{thebibliography}{75}
\providecommand{\natexlab}[1]{#1}
\providecommand{\url}[1]{\texttt{#1}}
\expandafter\ifx\csname urlstyle\endcsname\relax
  \providecommand{\doi}[1]{doi: #1}\else
  \providecommand{\doi}{doi: \begingroup \urlstyle{rm}\Url}\fi

\bibitem[Adams et~al.(2022)Adams, Makowski, Engel, et~al.]{adams2022p158}
Lisa~C. Adams, Marcus~R. Makowski, G{\"u}nther Engel, et~al.
\newblock Prostate158 - {{An}} expert-annotated {{3T MRI}} dataset and algorithm for prostate cancer detection.
\newblock \emph{Computers in Biology and Medicine}, 148:\penalty0 105817, 2022.

\bibitem[Ahn et~al.(2026)Ahn, Lee, and Park]{medAdapt3d}
Suhyun Ahn, Donggyu Lee, and Jinah Park.
\newblock Adaptability of {{Vision Foundation Models}} for {{3D Medical Image Segmentation}}.
\newblock \emph{IEEE Open J. Signal Process.}, 7:\penalty0 82--90, 2026.

\bibitem[Antonelli et~al.(2022)Antonelli, Reinke, Bakas, et~al.]{antonelli2022msd}
Michela Antonelli, Annika Reinke, Spyridon Bakas, et~al.
\newblock The {{Medical Segmentation Decathlon}}.
\newblock \emph{Nature Communications}, 13\penalty0 (1):\penalty0 4128, 2022.

\bibitem[Assran et~al.(2023)Assran, Duval, Misra, et~al.]{assran2023ijepa}
Mahmoud Assran, Quentin Duval, Ishan Misra, et~al.
\newblock Self-{{Supervised Learning}} from {{Images}} with a {{Joint-Embedding Predictive Architecture}}.
\newblock In \emph{IEEE Conf Comput Vis Pattern Recog}, pages 15619--15629, 2023.

\bibitem[Balazevic et~al.(2023)Balazevic, Steiner, Parthasarathy, et~al.]{balazevic2023hummingbird}
Ivana Balazevic, David Steiner, Nikhil Parthasarathy, et~al.
\newblock Towards in-context scene understanding.
\newblock In \emph{Adv Neural Inform Process Syst}, 2023.

\bibitem[Ben~Abacha et~al.(2026)Ben~Abacha, {Santamar{\'i}a-Pang}, Lee, et~al.]{mir3d}
Asma Ben~Abacha, Alberto {Santamar{\'i}a-Pang}, Ho~Hin Lee, et~al.
\newblock Benchmarks and methods for {{3D}} medical image retrieval.
\newblock \emph{Scientific Reports}, 16\penalty0 (1):\penalty0 21016, 2026.

\bibitem[Bernard et~al.(2018)Bernard, Lalande, Zotti, et~al.]{bernard2018acdc}
Olivier Bernard, Alain Lalande, Clement Zotti, et~al.
\newblock Deep {{Learning Techniques}} for {{Automatic MRI Cardiac Multi-Structures Segmentation}} and {{Diagnosis}}: {{Is}} the {{Problem Solved}}?
\newblock \emph{IEEE Trans Med Imaging}, 37\penalty0 (11):\penalty0 2514--2525, 2018.

\bibitem[{Biomedical Image Analysis Group, Imperial College London}(2007)]{IXIDataset}
{Biomedical Image Analysis Group, Imperial College London}.
\newblock {IXI} - {Information eXtraction from Images}, 2007.
\newblock EPSRC grant GR/S21533/02.

\bibitem[Blankemeier et~al.(2026)Blankemeier, Cohen, Kumar, et~al.]{blankemeier2024merlin}
Louis Blankemeier, Joseph~Paul Cohen, Ashwin Kumar, et~al.
\newblock Merlin: A computed tomography vision--language foundation model and dataset.
\newblock \emph{Nature}, 652\penalty0 (8112):\penalty0 1318--1328, 2026.

\bibitem[Butoi et~al.(2023)Butoi, Gonzalez~Ortiz, Ma, et~al.]{butoi2023universeg}
Victor~Ion Butoi, Jose~Javier Gonzalez~Ortiz, Tianyu Ma, et~al.
\newblock {{UniverSeg}}: {{Universal Medical Image Segmentation}}.
\newblock In \emph{Int Conf Comput Vis}, pages 21381--21394, 2023.

\bibitem[Caron et~al.(2021)Caron, Touvron, Misra, et~al.]{caron2021dino}
Mathilde Caron, Hugo Touvron, Ishan Misra, et~al.
\newblock Emerging {{Properties}} in {{Self-Supervised Vision Transformers}}.
\newblock In \emph{Int Conf Comput Vis}, pages 9630--9640, 2021.

\bibitem[Cerri et~al.(2025)Cerri, Munk, Llambias, et~al.]{fomo300k}
Stefano Cerri, Asbj{\o}rn Munk, Sebastian~N{\o}rgaard Llambias, et~al.
\newblock A large-scale heterogeneous {{3D}} magnetic resonance brain imaging dataset for self-supervised learning.
\newblock 2025.
\newblock arxiv:2506.14432.

\bibitem[Chen et~al.(2020)Chen, Kornblith, Norouzi, et~al.]{chen2020simclr}
Ting Chen, Simon Kornblith, Mohammad Norouzi, et~al.
\newblock A simple framework for contrastive learning of visual representations.
\newblock In \emph{Int Conf Mach Learn}, pages 1597--1607. JMLR.org, 2020.

\bibitem[Chen and He(2021)]{chen2021simsiam}
Xinlei Chen and Kaiming He.
\newblock Exploring {{Simple Siamese Representation Learning}}.
\newblock In \emph{IEEE Conf Comput Vis Pattern Recog}, pages 15745--15753, 2021.

\bibitem[Czolbe and Dalca(2023)]{czolbe2023neuralizer}
Steffen Czolbe and Adrian~V. Dalca.
\newblock Neuralizer: {{General Neuroimage Analysis}} without {{Re-Training}}.
\newblock In \emph{IEEE Conf Comput Vis Pattern Recog}, pages 6217--6230, 2023.

\bibitem[Darcet et~al.(2024)Darcet, Oquab, Mairal, et~al.]{darcet2024registers}
Timoth{\'e}e Darcet, Maxime Oquab, Julien Mairal, et~al.
\newblock Vision transformers need registers.
\newblock In \emph{Int Conf on Learn Represent}, 2024.

\bibitem[{de Verdier} et~al.(2024){de Verdier}, Saluja, Gagnon, et~al.]{brats}
Maria~Correia {de Verdier}, Rachit Saluja, Louis Gagnon, et~al.
\newblock The 2024 {{Brain Tumor Segmentation}} ({{BraTS}}) {{Challenge}}: {{Glioma Segmentation}} on {{Post}}-treatment {{MRI}}.
\newblock 2024.
\newblock arxiv:2405.18368.

\bibitem[Hamamci et~al.(2026)Hamamci, Er, Wang, et~al.]{hamamci2024ctrate}
Ibrahim~Ethem Hamamci, Sezgin Er, Chenyu Wang, et~al.
\newblock Generalist foundation models from a multimodal dataset for {{3D}} computed tomography.
\newblock \emph{Nature Biomedical Engineering}, 10\penalty0 (8):\penalty0 1610--1628, 2026.

\bibitem[Harsy et~al.(2026)Harsy, Wald, Gotkowski, et~al.]{nnfoundation}
Constantin~Ulrich Harsy, Tassilo Wald, Karol Gotkowski, et~al.
\newblock {{nnFoundation}}: {{3D Foundation Models}} for {{Radiology}}, 2026.
\newblock arxiv:2609.26924.

\bibitem[Hatamizadeh et~al.(2022)Hatamizadeh, Tang, Nath, et~al.]{hatamizadeh2022unetr}
Ali Hatamizadeh, Yucheng Tang, Vishwesh Nath, et~al.
\newblock {{UNETR}}: {{Transformers}} for {{3D Medical Image Segmentation}}.
\newblock In \emph{Proc IEEE/CVF Winter Conf Appl Comput Vis}, pages 1748--1758, 2022.

\bibitem[He et~al.(2022)He, Chen, Xie, et~al.]{he2022mae}
Kaiming He, Xinlei Chen, Saining Xie, et~al.
\newblock Masked autoencoders are scalable vision learners.
\newblock In \emph{IEEE Conf Comput Vis Pattern Recog}, pages 16000--16009, 2022.

\bibitem[He et~al.(2023)He, Yang, Ge, et~al.]{he2023gvsl}
Yuting He, Guanyu Yang, Rongjun Ge, et~al.
\newblock Geometric {{Visual Similarity Learning}} in {{3D Medical Image Self-Supervised Pre}}-training.
\newblock In \emph{IEEE Conf Comput Vis Pattern Recog}, pages 9538--9547, 2023.

\bibitem[Hernandez~Petzsche et~al.(2022)Hernandez~Petzsche, {de la Rosa}, Wiest, et~al.]{hernandez2022isles}
Moritz~Roman Hernandez~Petzsche, Ezequiel {de la Rosa}, Roland Wiest, et~al.
\newblock {{ISLES}} 2022: {{A}} multi-center magnetic resonance imaging stroke lesion segmentation dataset.
\newblock 2022.
\newblock zenodo:7960856.

\bibitem[Hinton et~al.(2015)Hinton, Vinyals, and Dean]{hinton2015distilling}
Geoffrey Hinton, Oriol Vinyals, and Jeff Dean.
\newblock Distilling the {{Knowledge}} in a {{Neural Network}}.
\newblock 2015.
\newblock arxiv:1503.02531.

\bibitem[Isensee et~al.(2021)Isensee, Jaeger, Kohl, et~al.]{isensee2021nnunet}
Fabian Isensee, Paul~F. Jaeger, Simon A.~A. Kohl, et~al.
\newblock {{nnU-Net}}: A self-configuring method for deep learning-based biomedical image segmentation.
\newblock \emph{Nature Methods}, 18\penalty0 (2):\penalty0 203--211, 2021.

\bibitem[Knaebel et~al.(2026)Knaebel, Yilmaz, De~Geus, et~al.]{dinoRoom}
Karim Knaebel, Kadir Yilmaz, Daan De~Geus, et~al.
\newblock {{DINO}} in the {{Room}}: {{Leveraging 2D Foundation Models}} for {{3D Segmentation}}.
\newblock In \emph{Proc. Int. Conf. 3D Vis. 3DV}, pages 1884--1894, 2026.

\bibitem[Kobayashi et~al.(2022)Kobayashi, Matsumoto, and Sitzmann]{kobayashi2022dff}
Sosuke Kobayashi, Eiichi Matsumoto, and Vincent Sitzmann.
\newblock Decomposing ne{RF} for editing via feature field distillation.
\newblock In \emph{Adv Neural Inform Process Syst}, 2022.

\bibitem[Konz et~al.(2022)Konz, Gu, Dong, et~al.]{intrinsicManifold}
Nicholas Konz, Hanxue Gu, Haoyu Dong, et~al.
\newblock The {{Intrinsic Manifolds}} of~{{Radiological Images}} and~{{Their Role}} in~{{Deep Learning}}.
\newblock In \emph{MICCAI}, pages 684--694, Cham, 2022. Springer Nature Switzerland.

\bibitem[Lebailly et~al.(2024)Lebailly, Stegm{\"u}ller, Bozorgtabar, et~al.]{lebailly2024cribo}
Tim Lebailly, Thomas Stegm{\"u}ller, Behzad Bozorgtabar, et~al.
\newblock Cr{IB}o: Self-supervised learning via cross-image object-level bootstrapping.
\newblock In \emph{Int Conf on Learn Represent}, 2024.

\bibitem[Li et~al.(2026)Li, Gao, Yuan, et~al.]{dinounet}
Haoyue Li, Yifan Gao, Feng Yuan, et~al.
\newblock {Dino U-Net: Exploiting High-Fidelity Dense Features from Foundation Models for Medical Image Segmentation}.
\newblock In \emph{MICCAI}. Springer Nature Switzerland, 2026.

\bibitem[Li et~al.(2025)Li, Wu, Lai, Hu, and Yang]{medinov3}
Yuheng Li, Yizhou Wu, Yuxiang Lai, Mingzhe Hu, and Xiaofeng Yang.
\newblock {{MedDINOv3}}: {{How}} to adapt vision foundation models for medical image segmentation?, 2025.
\newblock arxiv:2509.02379.

\bibitem[Liu et~al.(2025)Liu, Georgescu, Zhang, et~al.]{revisiting2dfm3dcls}
Han Liu, Bogdan Georgescu, Yanbo Zhang, et~al.
\newblock Revisiting {{2D Foundation Models}} for {{Scalable 3D Medical Image Classification}}.
\newblock 2025.
\newblock arxiv:2512.12887.

\bibitem[Ma et~al.(2024)Ma, He, Li, et~al.]{medsam}
Jun Ma, Yuting He, Feifei Li, et~al.
\newblock Segment anything in medical images.
\newblock \emph{Nature Communications}, 15\penalty0 (1):\penalty0 654, 2024.

\bibitem[Ma et~al.(2026)Ma, Kim, and Wang]{flare2026}
Jun Ma, Sumin Kim, and Bo Wang.
\newblock Fast, {{Low}}-resource, {{Accurate}}, {{Robust}}, and {{Effectual Medical Image Segmentation}} and {{Classification}} in {{3D CT}} and {{MRI Scans}}.
\newblock 2026.
\newblock zenodo:19847954.

\bibitem[Martino and Mostofsky(2023)]{abide}
Adriana~Di Martino and Stewart Mostofsky.
\newblock Autism brain imaging data exchange (abide), 2023.
\newblock IEEE DataPort:10.21227/y3v9-b041.

\bibitem[{M{\"u}ller-Franzes} et~al.(2025){M{\"u}ller-Franzes}, S{\'a}nchez, Payne, et~al.]{odelia}
Gustav {M{\"u}ller-Franzes}, Lorena~Escudero S{\'a}nchez, Nicholas Payne, et~al.
\newblock A {{European Multi-Center Breast Cancer MRI Dataset}}, 2025.
\newblock arxiv:2506.00474.

\bibitem[Munk et~al.(2026)Munk, Cerri, Nersesjan, et~al.]{munkFOMO}
Asbj{\o}rn Munk, Stefano Cerri, Vardan Nersesjan, et~al.
\newblock Towards {{Brain MRI Foundation Models}} for the {{Clinic}}: {{Findings}} from the {{FOMO25 Challenge}}, 2026.
\newblock arxiv:2604.11679.

\bibitem[Oquab et~al.(2024)Oquab, Darcet, Moutakanni, et~al.]{oquab2024dinov2}
Maxime Oquab, Timoth{\'e}e Darcet, Th{\'e}o Moutakanni, et~al.
\newblock {DINO}v2: Learning robust visual features without supervision.
\newblock \emph{Trans Mach Learn Res}, 2024.

\bibitem[Pariza et~al.(2024)Pariza, Salehi, and Asano]{pariza2024hbeval}
Valentinos Pariza, Mohammadreza Salehi, and Yuki Asano.
\newblock {Hummingbird Evaluation for vision encoders}, 2024.
\newblock Available: https://github.com/vpariza/open-hummingbird-eval.

\bibitem[Pariza et~al.(2025)Pariza, Salehi, Burghouts, et~al.]{pariza2025neco}
Valentinos Pariza, Mohammadreza Salehi, Gertjan~J. Burghouts, et~al.
\newblock Near, far: Patch-ordering enhances vision foundation models' scene understanding.
\newblock In \emph{Int Conf on Learn Represent}, 2025.

\bibitem[Park et~al.(2019)Park, Kim, Lu, et~al.]{park2019relational}
Wonpyo Park, Dongju Kim, Yan Lu, et~al.
\newblock Relational {{Knowledge Distillation}}.
\newblock In \emph{IEEE Conf Comput Vis Pattern Recog}, pages 3962--3971, 2019.

\bibitem[Qureshi et~al.(2023)Qureshi, Hussain, Ibrar, et~al.]{mgmt}
Shahzad~Ahmad Qureshi, Lal Hussain, Usama Ibrar, et~al.
\newblock Radiogenomic classification for {{MGMT}} promoter methylation status using multi-omics fused feature space for least invasive diagnosis through {{mpMRI}} scans.
\newblock \emph{Scientific Reports}, 13\penalty0 (1):\penalty0 3291, 2023.

\bibitem[Rakic et~al.(2024)Rakic, Wong, Gonzalez~Ortiz, et~al.]{rakic2024tyche}
Marianne Rakic, Hallee~E. Wong, Jose~Javier Gonzalez~Ortiz, et~al.
\newblock Tyche: {{Stochastic}} in-{{Context Learning}} for {{Medical Image Segmentation}}.
\newblock In \emph{IEEE Conf Comput Vis Pattern Recog}, pages 11159--11173, 2024.

\bibitem[Ramakrishnan et~al.(2024)Ramakrishnan, Jekel, Chadha, et~al.]{ybm}
Divya Ramakrishnan, Leon Jekel, Saahil Chadha, et~al.
\newblock A large open access dataset of brain metastasis {{3D}} segmentations on {{MRI}} with clinical and imaging information.
\newblock \emph{Scientific Data}, 11\penalty0 (1):\penalty0 254, 2024.

\bibitem[Ranzinger et~al.(2024)Ranzinger, Heinrich, Kautz, et~al.]{ranzinger2024radio}
Mike Ranzinger, Greg Heinrich, Jan Kautz, et~al.
\newblock {{AM-RADIO}}: {{Agglomerative Vision Foundation Model Reduce All Domains Into One}}.
\newblock In \emph{IEEE Conf Comput Vis Pattern Recog}, pages 12490--12500, 2024.

\bibitem[Revel et~al.(2021)Revel, Boussouar, {de Margerie-Mellon}, et~al.]{revelStudyThoracicCT2021}
Marie-Pierre Revel, Samia Boussouar, Constance {de Margerie-Mellon}, et~al.
\newblock Study of {{Thoracic CT}} in {{COVID-19}}: {{The STOIC Project}}.
\newblock \emph{Radiology}, 301\penalty0 (1):\penalty0 E361--E370, 2021.

\bibitem[Romero et~al.(2014)Romero, Ballas, Kahou, et~al.]{romero2015fitnets}
Adriana Romero, Nicolas Ballas, Samira~Ebrahimi Kahou, et~al.
\newblock {{FitNets}}: {{Hints}} for {{Thin Deep Nets}}.
\newblock 2014.
\newblock arxiv:1412.6550.

\bibitem[Roy and Vetterli(2007)]{roy2007effectiverank}
Olivier Roy and Martin Vetterli.
\newblock The effective rank: {{A}} measure of effective dimensionality.
\newblock In \emph{2007 15th {{European Signal Processing Conference}}}, pages 606--610, 2007.

\bibitem[Sar{\i}y{\i}ld{\i}z et~al.(2025)Sar{\i}y{\i}ld{\i}z, Weinzaepfel, Lucas, et~al.]{sariyildiz2024unic}
Mert~B{\"u}lent Sar{\i}y{\i}ld{\i}z, Philippe Weinzaepfel, Thomas Lucas, et~al.
\newblock {{UNIC}}: {{Universal Classification Models}} via~{{Multi}}-teacher {{Distillation}}.
\newblock In \emph{Eur Conf Comput Vis}, pages 353--371, Cham, 2025. Springer Nature Switzerland.

\bibitem[Sautier et~al.(2022)Sautier, Puy, Gidaris, et~al.]{sautier2022slidr}
Corentin Sautier, Gilles Puy, Spyros Gidaris, et~al.
\newblock Image-to-{{Lidar Self-Supervised Distillation}} for {{Autonomous Driving Data}}.
\newblock In \emph{IEEE Conf Comput Vis Pattern Recog}, pages 9881--9891, 2022.

\bibitem[Shang et~al.(2025)Shang, Schmeckpeper, May, et~al.]{shang2024theia}
Jinghuan Shang, Karl Schmeckpeper, Brandon~B. May, et~al.
\newblock Theia: {{Distilling Diverse Vision Foundation Models}} for {{Robot Learning}}.
\newblock In \emph{Proceedings of {{The}} 8th {{Conference}} on {{Robot Learning}}}, pages 724--748. PMLR, 2025.

\bibitem[Siméoni et~al.(2025)Siméoni, Vo, Seitzer, et~al.]{simeoni2025dinov3}
Oriane Siméoni, Huy~V. Vo, Maximilian Seitzer, et~al.
\newblock {DINOv3}.
\newblock \emph{CoRR}, abs/2508.10104, 2025.

\bibitem[Song et~al.(2025)Song, Xu, Zhang, et~al.]{dinoreg}
Xinrui Song, Xuanang Xu, Jiajin Zhang, et~al.
\newblock {{DINO-Reg}}: {{Efficient Multimodal Image Registration With Distilled Features}}.
\newblock \emph{IEEE Trans Med Imaging}, 44\penalty0 (9):\penalty0 3809--3819, 2025.

\bibitem[Tang et~al.(2022)Tang, Yang, Li, et~al.]{tang2022swinunetr}
Yucheng Tang, Dong Yang, Wenqi Li, et~al.
\newblock Self-{{Supervised Pre-Training}} of {{Swin Transformers}} for {{3D Medical Image Analysis}}.
\newblock In \emph{IEEE Conf Comput Vis Pattern Recog}, pages 20698--20708, 2022.

\bibitem[Touvron et~al.(2020)Touvron, Cord, Douze, et~al.]{touvron2021deit}
Hugo Touvron, Matthieu Cord, Matthijs Douze, et~al.
\newblock {Training Data-Efficient Image Transformers \& Distillation through Attention}.
\newblock 2020.
\newblock arxiv:2012.12877.

\bibitem[Tung and Mori(2019)]{tung2019similarity}
Fred Tung and Greg Mori.
\newblock Similarity-{{Preserving Knowledge Distillation}}.
\newblock In \emph{Int Conf Comput Vis}, pages 1365--1374, 2019.

\bibitem[Vaish et~al.(2025)Vaish, Meister, Heimann, et~al.]{cvaVaish}
Puru Vaish, Felix Meister, Tobias Heimann, et~al.
\newblock Consistent {{View Alignment Improves Foundation Models}} for {{3D Medical Image Segmentation}}, 2025.
\newblock arxiv:2509.13846.

\bibitem[Vaish et~al.(2026)Vaish, Dashti, Strisciuglio, et~al.]{compositionalityPrimitives}
Puru Vaish, Meelad Dashti, Nicola Strisciuglio, et~al.
\newblock Compositionality with {{Primitives}}: {{A Unified View}} from {{Euclidean}} to {{Hyperbolic Vision-Language Models}}, 2026.
\newblock arxiv:2610.29811.

\bibitem[Wald et~al.(2025{\natexlab{a}})Wald, Ulrich, Lukyanenko, et~al.]{wald2025revisitingmae}
Tassilo Wald, Constantin Ulrich, Stanislav Lukyanenko, et~al.
\newblock Revisiting {{MAE}} pre-training for {{3D}} medical image segmentation.
\newblock In \emph{IEEE Conf Comput Vis Pattern Recog}, pages 5186--5196, 2025{\natexlab{a}}.

\bibitem[Wald et~al.(2025{\natexlab{b}})Wald, Ulrich, Suprijadi, et~al.]{wald2025openmind}
Tassilo Wald, Constantin Ulrich, Jonathan Suprijadi, et~al.
\newblock An {{OpenMind}} for {{3D Medical Vision Self}}-supervised {{Learning}}.
\newblock In \emph{Int Conf Comput Vis}, pages 23839--23879, 2025{\natexlab{b}}.

\bibitem[Wald et~al.(2026)Wald, Roy, Isensee, et~al.]{wald2025primus}
Tassilo Wald, Saikat Roy, Fabian Isensee, et~al.
\newblock {Primus}: {Enforcing Attention Usage for 3D Medical Image Segmentation}.
\newblock \emph{Trans Mach Learn Res}, 2026.

\bibitem[Wang et~al.(2026)Wang, Zhang, Gao, et~al.]{beyondscaling2025}
Chong Wang, Yabin Zhang, Yunhe Gao, et~al.
\newblock A data- and compute-efficient chest {{X}}-ray foundation model beyond aggressive scaling.
\newblock 2026.
\newblock arxiv:2602.22843.

\bibitem[Wang et~al.(2023)Wang, Wu, Luo, et~al.]{wang2023misfm}
Guotai Wang, Jianghao Wu, Xiangde Luo, et~al.
\newblock {{MIS-FM}}: {{3D Medical Image Segmentation}} using {{Foundation Models Pretrained}} on a {{Large-Scale Unannotated Dataset}}, 2023.
\newblock arxiv:2306.16925.

\bibitem[Wei et~al.(2022)Wei, Hu, Xie, et~al.]{wei2022fd}
Yixuan Wei, Han Hu, Zhenda Xie, et~al.
\newblock Contrastive {{Learning Rivals Masked Image Modeling}} in {{Fine}}-tuning via {{Feature Distillation}}.
\newblock 2022.
\newblock arxiv:2205.14141.

\bibitem[Wen et~al.(2021)Wen, Chen, Deng, et~al.]{wen2021rethinking}
Yang Wen, Leiting Chen, Yu Deng, et~al.
\newblock Rethinking pre-training on medical imaging.
\newblock \emph{J. Vis. Comun. Image Represent.}, 78\penalty0 (C), 2021.

\bibitem[Wiesenfarth et~al.(2021)Wiesenfarth, Reinke, Landman, et~al.]{Wiesenfarth2021}
Manuel Wiesenfarth, Annika Reinke, Bennett~A Landman, et~al.
\newblock Methods and open-source toolkit for analyzing and visualizing challenge results.
\newblock \emph{{Scientific Reports}}, 11\penalty0 (1):\penalty0 2369, 2021.

\bibitem[Wu et~al.(2024)Wu, Zhuang, and Chen]{wu2024voco}
Linshan Wu, Jiaxin Zhuang, and Hao Chen.
\newblock {{VoCo}}: {{A Simple-Yet-Effective Volume Contrastive Learning Framework}} for {{3D Medical Image Analysis}}.
\newblock In \emph{IEEE Conf Comput Vis Pattern Recog}, pages 22873--22882, 2024.

\bibitem[Xie et~al.(2022)Xie, Zhang, Cao, et~al.]{xie2022simmim}
Zhenda Xie, Zheng Zhang, Yue Cao, et~al.
\newblock Simmim: A simple framework for masked image modeling.
\newblock In \emph{IEEE Conf Comput Vis Pattern Recog}, pages 9653--9663, 2022.

\bibitem[Xu et~al.(2025)Xu, Hosseini, Anderson, et~al.]{3dino}
Tony Xu, Sepehr Hosseini, Chris Anderson, et~al.
\newblock A generalizable {{3D}} framework and model for self-supervised learning in medical imaging.
\newblock \emph{npj Digital Medicine}, 8\penalty0 (1):\penalty0 639, 2025.

\bibitem[Zagoruyko and Komodakis(2017)]{zagoruyko2017attention}
Sergey Zagoruyko and Nikos Komodakis.
\newblock Paying more attention to attention: Improving the performance of convolutional neural networks via attention transfer.
\newblock In \emph{Int Conf on Learn Represent}, 2017.

\bibitem[Zhang et~al.(2023)Zhang, Wang, Qiao, et~al.]{zhangLearning3DRepresentations2023}
Renrui Zhang, Liuhui Wang, Yu Qiao, et~al.
\newblock Learning {{3D Representations}} from {{2D Pre-Trained Models}} via {{Image}}-to-{{Point Masked Autoencoders}}.
\newblock In \emph{IEEE Conf Comput Vis Pattern Recog}, pages 21769--21780, 2023.

\bibitem[Zhao et~al.(2026)Zhao, Yang, Li, et~al.]{segmic}
Jianwei Zhao, Fan Yang, Xin Li, et~al.
\newblock {{SegMIC}}: {{A}} universal model for medical image segmentation through in-context learning.
\newblock \emph{Pattern Recognition}, 171\penalty0 (PA), 2026.

\bibitem[Zhou et~al.(2022)Zhou, Wei, Wang, et~al.]{zhou2022ibot}
Jinghao Zhou, Chen Wei, Huiyu Wang, et~al.
\newblock Image {BERT} pre-training with online tokenizer.
\newblock In \emph{Int Conf on Learn Represent}, 2022.

\bibitem[Zhou et~al.(2019)Zhou, Sodha, Siddiquee, et~al.]{zhou2019modelsgenesis}
Zongwei Zhou, Vatsal Sodha, Md~Mahfuzur~Rahman Siddiquee, et~al.
\newblock Models {{Genesis}}: {{Generic Autodidactic Models}} for {{3D Medical Image Analysis}}.
\newblock \emph{MICCAI}, 11767:\penalty0 384--393, 2019.

\bibitem[Ziegler and Asano(2022)]{ziegler2022leopart}
Adrian Ziegler and Yuki~M. Asano.
\newblock Self-supervised learning of object parts for semantic segmentation.
\newblock In \emph{IEEE Conf Comput Vis Pattern Recog}, pages 14502--14511, 2022.

\end{thebibliography}
}

\clearpage
\appendix
\maketitlesupplementary
\section{Implementation details}\label{supp:impl}

\paragraph{Pretraining hyperparameters.}
\Cref{tab:supp_hparams} lists the pretraining configuration used for all \ours runs.
These settings are fixed and were not tuned.

\begin{table}[h]
    \centering
    \caption{Pretraining hyperparameters of \ours.}
    \label{tab:supp_hparams}
    \resizebox{\linewidth}{!}{%
    \small
\begin{tabular}{l l}
    \toprule
    Setting & Value \\
    \midrule
    Optimizer & AdamW \\
    Peak learning rate & $2\times10^{-3}$ \\
    LR schedule & linear warm-up from $10^{-5}$ (5\%), cosine to $10^{-6}$ \\
    Weight decay & cosine, $0.04 \rightarrow 0.40$ \\
    Layer-wise LR decay & 0.9 \\
    Patch-embedding LR multiplier & 0.2 \\
    Gradient clipping & 0.5 \\
    Training steps & 30k \\
    Batch size & 64 (4 per GPU $\times$ 8 GPUs $\times$ 2 Grad. Acc.) \\
    Precision & bf16 mixed \\
    Input crop & $192^3$ \\
    Student patch size & $8^3$ \\
    Reference resolution & $336\times336$ ($q_T=14$) \\
    Running-norm momentum & cosine $0.1\rightarrow0$ over first 10\%, then 0 \\
    \midrule
    Spatial augmentation & $90^\circ$ rotations, flips, axis permutations \\
    & (all three axes, uniform probability) \\
    Intensity augmentation & none \\
    \bottomrule
\end{tabular}

}
\end{table}

\paragraph{Pseudocode.}
\Cref{alg:mimic} summarizes one training step.

\begin{algorithm}[h]
    \caption{One training step of \ours}
    \label{alg:mimic}
    \small
    \begin{algorithmic}[1]
        \Require volume batch $\mathbf{x}$, frozen 2D model $f_T$, 3D student $f_S$, running statistics $(\bm{\mu},\bm{\sigma}^2)$
        \State $\widetilde{\mathbf{x}} \gets$ min--max normalize $\mathbf{x}$, apply ImageNet normalization
        \For{$p \in \{\mathrm{ax},\mathrm{cor},\mathrm{sag}\}$}
            \State slice $\widetilde{\mathbf{x}}$ along $p$, resize slices to $(q_T G_1, q_T G_2)$
            \State $(\mathbf{c}^p_i,\mathbf{Z}^p_i) \gets f_T(\mathbf{x}^p_i)$ for all slices $i$
            \State $\mathbf{T}^p \gets$ stack $\mathbf{Z}^p_i$, average-pool along slicing axis
            \State $\bar{\mathbf{c}}^p \gets$ Euclidean medoid of $\{\mathbf{c}^p_i\}_i$
        \EndFor
        \State $\mathbf{T} \gets \tfrac13\sum_p \mathbf{T}^p$, \quad $\mathbf{A} \gets \tfrac13\sum_p \mathcal{A}(\bar{\mathbf{c}}^p,\mathbf{T}^p)$
        \State $\mathbf{M} \gets \mathbf{A} \geq \tfrac12(\min\mathbf{A}+\max\mathbf{A})$ \Comment{per volume}
        \State update $(\bm{\mu},\bm{\sigma}^2)$ from $\mathbf{T}$ over $\mathbf{M}$ with current momentum
        \State $\widehat{\mathbf{T}} \gets (\mathbf{T}-\bm{\mu})/\sqrt{\bm{\sigma}^2+\epsilon}$
        \State $(\mathbf{s}_{\mathrm{cls}},\mathbf{S}) \gets f_S(\mathbf{x})$
        \State $\mathcal{L} \gets \operatorname{MSE}(\mathbf{S},\widehat{\mathbf{T}}) + \operatorname{MSE}(\mathcal{A}(\mathbf{s}_{\mathrm{cls}},\mathbf{S}),\mathbf{A})$
        \State update $f_S$ with $\nabla\mathcal{L}$
    \end{algorithmic}
\end{algorithm}

\paragraph{Compute.}
Pretraining a \ours takes in total 480 GPU-hours on 8$\times$ NVIDIA L40s.
The reference model runs in inference mode on $3\times336$ slices per volume.
\section{Datasets}\label{supp:data}

\subsection{Pretraining data}
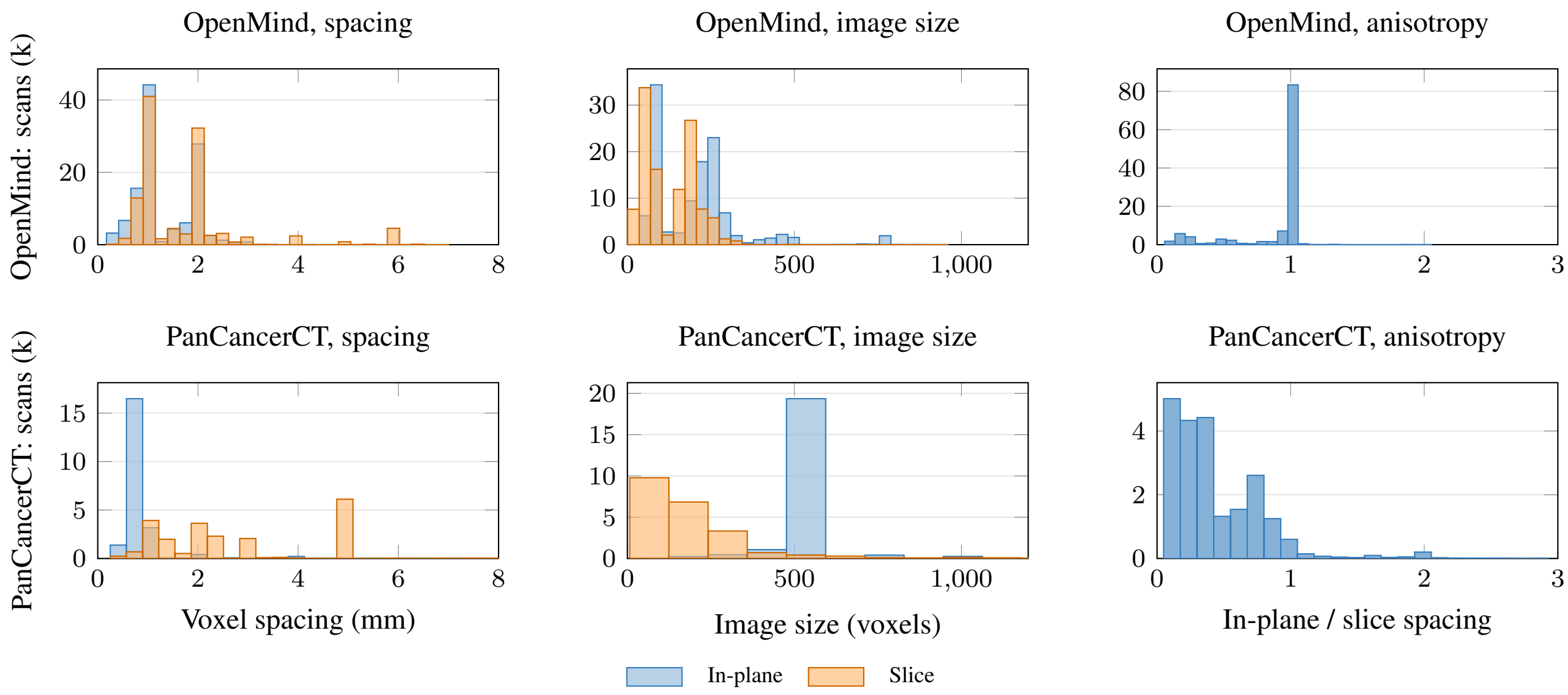
\begin{figure*}[t]
    \centering
    \begin{tikzpicture}
    \begin{groupplot}[
        group style={group name=pre, group size=3 by 2, horizontal sep=1.4cm, vertical sep=1.5cm},
        width=0.34\textwidth, height=0.2\textwidth,
        ymin=0,
        tick label style={font=\footnotesize}, label style={font=\small}, title style={font=\small},
        ymajorgrids, grid style={gray!20},
        legend style={font=\scriptsize, at={(0.98,0.97)}, anchor=north east, draw=none, fill=white, fill opacity=0.8, text opacity=1},
        scaled y ticks=false,
    ]
        \nextgroupplot[ylabel={OpenMind: scans (k)},  xmin=0, xmax=8, title={OpenMind, spacing}]
        \addplot[ybar interval, area legend, fill=cvprblue!50, fill opacity=0.7, draw=cvprblue] coordinates {(0.1739,3.214) (0.4177,6.701) (0.6615,15.6) (0.9053,44.2) (1.149,0.846) (1.393,4.251) (1.637,6.045) (1.88,27.86) (2.124,2.52) (2.368,1.235) (2.612,0.625) (2.856,0.749) (3.099,0.027) (3.343,0) (3.587,0) (3.831,0) (4.075,0.012) (4.318,0) (4.562,0) (4.806,0) (5.05,0) (5.294,0) (5.537,0) (5.781,0) (6.025,0) (6.269,0) (6.513,0) (6.756,0) (7,0)};
        \addplot[ybar interval, area legend, fill=orange!60, fill opacity=0.6, draw=orange!80!black] coordinates {(0.1739,0.126) (0.4177,1.748) (0.6615,12.91) (0.9053,41) (1.149,1.638) (1.393,4.495) (1.637,2.962) (1.88,32.24) (2.124,2.571) (2.368,3.115) (2.612,0.745) (2.856,2.085) (3.099,0.075) (3.343,0.066) (3.587,0.003) (3.831,2.428) (4.075,0.017) (4.318,0.005) (4.562,0.001) (4.806,0.816) (5.05,0.015) (5.294,0.132) (5.537,0.002) (5.781,4.54) (6.025,0.011) (6.269,0.135) (6.513,0.001) (6.756,0.003) (7,0)};
        \nextgroupplot[  xmin=0, xmax=1200, title={OpenMind, image size}]
        \addplot[ybar interval, area legend, fill=cvprblue!50, fill opacity=0.7, draw=cvprblue] coordinates {(1,0.003) (35.25,6.211) (69.5,34.36) (103.8,2.728) (138,2.518) (172.2,9.411) (206.5,17.84) (240.8,23.01) (275,6.844) (309.2,1.946) (343.5,0.412) (377.8,1.051) (412,1.405) (446.2,2.207) (480.5,1.552) (514.8,0.021) (549,0.014) (583.2,0.017) (617.5,0.041) (651.8,0.038) (686,0.154) (720.2,0.088) (754.5,1.917) (788.8,0.037) (823,0.023) (857.2,0.024) (891.5,0.007) (925.8,0.004) (960,0)};
        \addplot[ybar interval, area legend, fill=orange!60, fill opacity=0.6, draw=orange!80!black] coordinates {(1,7.623) (35.25,33.75) (69.5,16.2) (103.8,2.054) (138,11.88) (172.2,26.73) (206.5,7.651) (240.8,5.773) (275,1.248) (309.2,0.796) (343.5,0.076) (377.8,0.014) (412,0.004) (446.2,0.025) (480.5,0.054) (514.8,0.002) (549,0) (583.2,0) (617.5,0) (651.8,0) (686,0) (720.2,0) (754.5,0) (788.8,0) (823,0) (857.2,0.001) (891.5,0) (925.8,0) (960,0)};
        \nextgroupplot[  xmin=0, xmax=3, title={OpenMind, anisotropy}]
        \addplot[ybar interval, area legend, fill=cvprblue!60, draw=cvprblue] coordinates {(0.05718,1.792) (0.134,5.777) (0.2108,4.087) (0.2876,0.558) (0.3644,0.813) (0.4412,2.924) (0.518,2.296) (0.5948,0.677) (0.6716,0.496) (0.7484,1.607) (0.8253,1.568) (0.9021,7.148) (0.9789,83.39) (1.056,0.48) (1.132,0.038) (1.209,0.023) (1.286,0.138) (1.363,0.003) (1.44,0) (1.517,0) (1.593,0) (1.67,0) (1.747,0) (1.824,0.047) (1.901,0.004) (1.977,0.019) (2.054,0)};
        \nextgroupplot[ylabel={PanCancerCT: scans (k)}, xlabel={Voxel spacing (mm)}, xmin=0, xmax=8, title={PanCancerCT, spacing}]
        \addplot[ybar interval, area legend, fill=cvprblue!50, fill opacity=0.7, draw=cvprblue] coordinates {(0.25,1.386) (0.5729,16.49) (0.8958,3.174) (1.219,0.056) (1.542,0.001) (1.865,0.409) (2.188,0) (2.51,0.07) (2.833,0) (3.156,0) (3.479,0) (3.802,0.213) (4.125,0) (4.448,0) (4.771,0) (5.094,0) (5.417,0) (5.74,0) (6.062,0) (6.385,0) (6.708,0) (7.031,0) (7.354,0) (7.677,0) (8,0)};
        \addplot[ybar interval, area legend, fill=orange!60, fill opacity=0.6, draw=orange!80!black] coordinates {(0.25,0.241) (0.5729,0.689) (0.8958,3.923) (1.219,1.986) (1.542,0.519) (1.865,3.635) (2.188,2.303) (2.51,0.004) (2.833,2.061) (3.156,0.088) (3.479,0.118) (3.802,0.051) (4.125,0.001) (4.448,0) (4.771,6.121) (5.094,0) (5.417,0) (5.74,0.003) (6.062,0) (6.385,0) (6.708,0.006) (7.031,0) (7.354,0.043) (7.677,0.003) (8,0)};
        \nextgroupplot[ xlabel={Image size (voxels)}, xmin=0, xmax=1200, title={PanCancerCT, image size}]
        \addplot[ybar interval, area legend, fill=cvprblue!50, fill opacity=0.7, draw=cvprblue] coordinates {(7,0) (124.3,0.235) (241.7,0.45) (359,1.071) (476.3,19.36) (593.7,0.017) (711,0.404) (828.3,0.002) (945.7,0.252) (1063,0) (1180,0) (1298,0) (1415,0) (1532,0) (1650,0) (1767,0) (1884,0) (2002,0) (2119,0) (2236,0) (2354,0) (2471,0)};
        \addplot[ybar interval, area legend, fill=orange!60, fill opacity=0.6, draw=orange!80!black] coordinates {(7,9.779) (124.3,6.841) (241.7,3.327) (359,0.706) (476.3,0.417) (593.7,0.291) (711,0.096) (828.3,0.074) (945.7,0.122) (1063,0.081) (1180,0.045) (1298,0.008) (1415,0) (1532,0.001) (1650,0.001) (1767,0.004) (1884,0) (2002,0) (2119,0) (2236,0.001) (2354,0.001) (2471,0)};
        \nextgroupplot[ xlabel={In-plane / slice spacing}, xmin=0, xmax=3, title={PanCancerCT, anisotropy}]
        \addplot[ybar interval, area legend, fill=cvprblue!60, draw=cvprblue] coordinates {(0.04888,5.014) (0.1741,4.33) (0.2994,4.421) (0.4246,1.322) (0.5499,1.541) (0.6751,2.607) (0.8004,1.25) (0.9256,0.6) (1.051,0.146) (1.176,0.07) (1.301,0.042) (1.427,0.029) (1.552,0.095) (1.677,0.034) (1.802,0.05) (1.928,0.2) (2.053,0.023) (2.178,0.013) (2.303,0.004) (2.429,0.002) (2.554,0.001) (2.679,0) (2.804,0.001) (2.93,0)};
    \end{groupplot}
\end{tikzpicture}

\begin{tikzpicture}
    \begin{customlegend}[legend columns=2, legend style={font=\scriptsize, draw=none, column sep=2mm, at={($(pre c2r1.north)+(0,5mm)$)}, anchor=south}, legend entries={In-plane, Slice}]
        \addlegendimage{area legend, fill=cvprblue!50, fill opacity=0.7, draw=cvprblue}
        \addlegendimage{area legend, fill=orange!60, fill opacity=0.6, draw=orange!80!black}
    \end{customlegend}
\end{tikzpicture}
    \caption{Voxel spacing, image size, and anisotropy of the pretraining sources (top: OpenMind, bottom: PanCancerCT). Spacing and image size are shown separately for the two in-plane axes (blue) and the slice axis (orange). Anisotropy is the ratio of in-plane to slice spacing, where 1 is isotropic.}
    \label{fig:supp_pretrain_stats}
\end{figure*}

\begin{figure*}[t]
    \centering
    \begin{tikzpicture}
    \begin{groupplot}[
        group style={group size=2 by 1, horizontal sep=2.6cm},
        width=0.42\textwidth, height=0.36\textwidth,
        xbar, /pgf/bar width=4pt,
        xmode=log, log basis x=10, xmin=10,
        y dir=reverse, ytick=data,
        tick label style={font=\scriptsize}, label style={font=\small}, title style={font=\small},
        xmajorgrids, grid style={gray!20},
        enlarge y limits=0.03,
    ]
        \nextgroupplot[height=0.5\textwidth, title={(a) OpenMind: scans per MRI sequence}, xlabel={Scans}, symbolic y coords={{T1w},{T2w},{FA},{MD},{FLAIR},{MP2RAGE},{sbref},{ADC},{TRACE},{UNIT1},{inplaneT2},{SWI},{UNIT1\_denoised},{minIP},{PET},{T1map},{inplaneT1},{angio},{PDw},{T2starw},{FLASH},{T2map},{T2starmap},{DWI}}]
        \addplot[fill=cvprblue!60, draw=cvprblue] coordinates {(42701,T1w) (22998,T2w) (13407,FA) (13407,MD) (5583,FLAIR) (2859,MP2RAGE) (1795,sbref) (1757,ADC) (1715,TRACE) (927,UNIT1) (892,inplaneT2) (784,SWI) (724,UNIT1\_denoised) (687,minIP) (653,PET) (557,T1map) (529,inplaneT1) (488,angio) (466,PDw) (409,T2starw) (305,FLASH) (106,T2map) (76,T2starmap) (55,DWI)};
        \nextgroupplot[height=0.5\textwidth, title={(b) PanCancerCT: scans per anatomy}, xlabel={Scans}, symbolic y coords={{lung},{whole body},{abdomen},{pancreas},{liver},{lymph nodes},{kidney},{head neck},{colon},{esophagus},{pelvis},{spine},{multi region},{spleen},{adrenal}}]
        \addplot[fill=orange!60, draw=orange!80!black] coordinates {(6979,lung) (6759,whole body) (2350,abdomen) (1936,pancreas) (1841,liver) (530,lymph nodes) (488,kidney) (240,head neck) (190,colon) (154,esophagus) (79,pelvis) (72,spine) (64,multi region) (61,spleen) (52,adrenal)};
    \end{groupplot}
\end{tikzpicture}
    \caption{Composition of the pretraining sources: (a) OpenMind scans per MRI sequence or derived map, (b) PanCancerCT scans per anatomy.}
    \label{fig:supp_pretrain_counts}
\end{figure*}
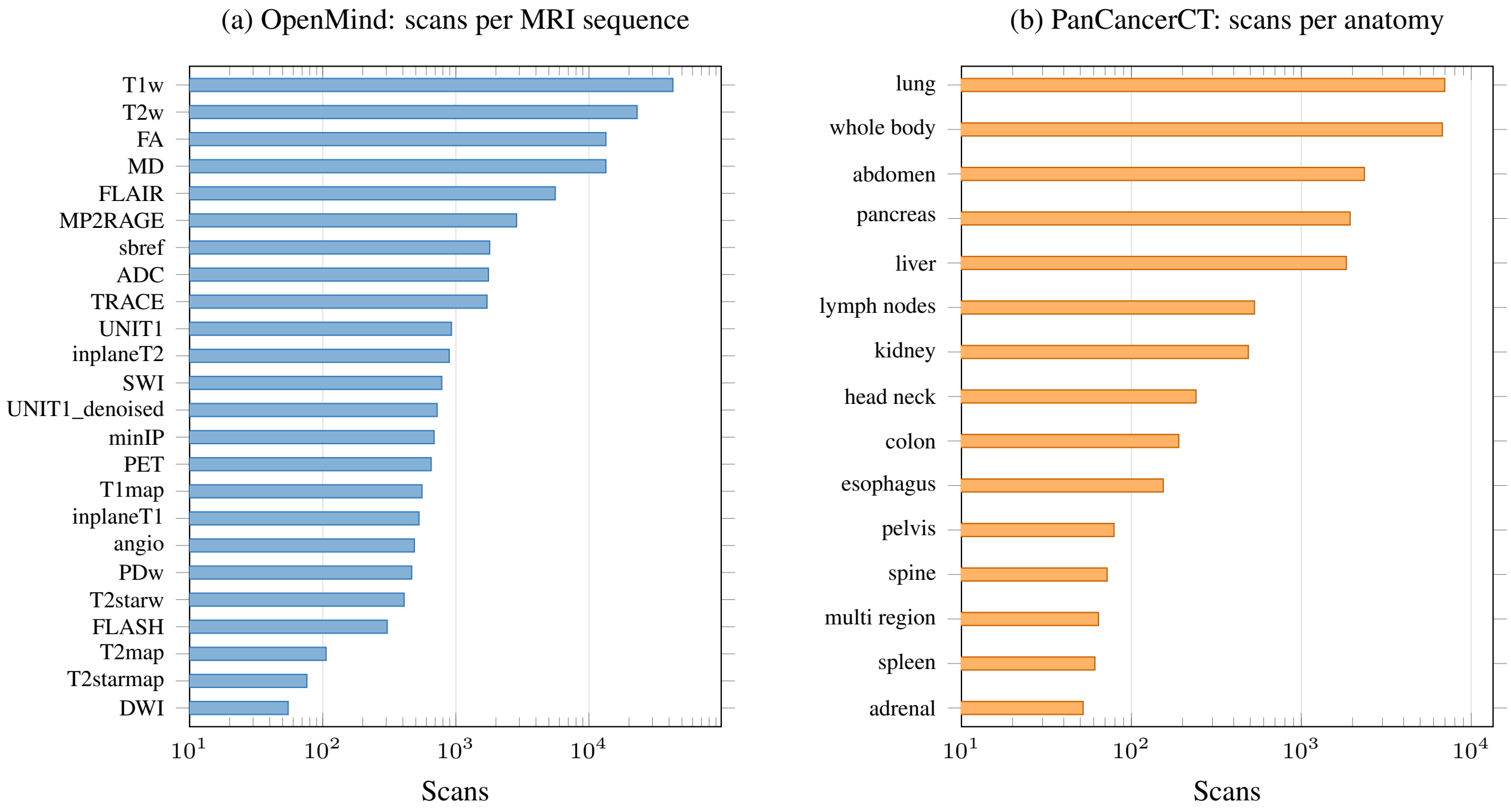

\Cref{tab:supp_pretrain} summarizes both pretraining sources, \cref{fig:supp_pretrain_stats} compares their voxel spacing, image size, and anisotropy, and \cref{fig:supp_pretrain_counts} shows their composition by MRI sequence and by anatomy.
OpenMind is dominated by near-isotropic brain MRI at about 1\,mm, whereas PanCancerCT consists mostly of anisotropic CT with fine in-plane (median 0.76\,mm) and coarser slice spacing (median 2\,mm) and larger in-plane image sizes.

\begin{table}[t]
    \centering
    \caption{Pretraining sources. Statistics are computed per scan from the image headers.}
    \label{tab:supp_pretrain}
    \resizebox{\linewidth}{!}{
    \small
\begin{tabular}{l r r}
    \toprule
    & OpenMind & PanCancerCT \\
    \midrule
    Scans & 113{,}880 & 21{,}802 \\
    Source datasets & 799 & 17 \\
    Subjects & 34{,}132 & -- \\
    Modalities / anatomies & 24 sequences & 15 anatomies \\
    \midrule
    \multicolumn{3}{l}{\textit{Median (min--max)}} \\
    In-plane spacing (mm) & 1.00 (0.17--4.17) & 0.76 (0.25--4.07) \\
    Slice spacing (mm) & 1.20 (0.20--7.00) & 2.00 (0.25--8.00) \\
    In-plane size (voxels) & 210 (34--960) & 512 (168--1024) \\
    Slice count & 100 (1--867) & 136 (7--2471) \\
    Anisotropy (max/min spacing) & 1.00 (1.00--17.49) & 3.07 (1.00--20.46) \\
    \bottomrule
\end{tabular}

    }
\end{table}

\subsection{Downstream data}
\Cref{tab:supp_datasets} lists the downstream datasets with their input channels, classes, number of cases, and the nnU-Net configuration used for all models.

\begin{table*}[t]
    \centering
    \caption{Spacing and size are the medians over all cases
    The batch size is 2 for all datasets. \dev: developmental dataset. 
    }
    \label{tab:supp_datasets}
    \resizebox{\textwidth}{!}{%
    \scriptsize
\setlength{\tabcolsep}{3pt}
\begin{tabular}{l l l >{\raggedright\arraybackslash}p{3.4cm} c r c c c}
    \toprule
    Group & Dataset & Input channels & Classes (foreground) & \#Cl. & \#Cases & Median spacing (mm) & Median size (voxels) & Input patch (voxels) \\
    \midrule
    Brain MRI & MSD Brain Tumour & FLAIR, T1w, t1gd, T2w & edema, non-enhancing tumor, enhancing tumour & 3 & 484 & 1 $\times$ 1 $\times$ 1 & 138 $\times$ 169 $\times$ 138 & 144 $\times$ 176 $\times$ 144 \\
     & MSD Hippocampus & MRI & Anterior, Posterior & 2 & 260 & 1 $\times$ 1 $\times$ 1 & 36 $\times$ 50 $\times$ 35 & 48 $\times$ 64 $\times$ 48 \\
     & YBM & T1C, T1N, T2F, T2W & TumourNecrosis, PeritumoralEdema, ContrastEnhancingTumour & 3 & 200 & 1 $\times$ 1 $\times$ 1 & 143 $\times$ 177 $\times$ 138 & 144 $\times$ 192 $\times$ 144 \\
     & ISLES 2022\dev & DWI, ADC & stroke & 1 & 250 & 2 $\times$ 2 $\times$ 2 & 65 $\times$ 84 $\times$ 70 & 80 $\times$ 96 $\times$ 80 \\
     & BraTS 2024 GLI & T1, T1C, T2W, T2F & NETC, SNFH, ET, RC & 4 & 1350 & 1 $\times$ 1 $\times$ 1 & 142 $\times$ 175 $\times$ 136 & 160 $\times$ 192 $\times$ 160 \\
    \midrule
    Body MRI & MSD Heart & MRI & left atrium & 1 & 20 & 1.37 $\times$ 1.25 $\times$ 1.25 & 115 $\times$ 320 $\times$ 232 & 128 $\times$ 224 $\times$ 192 \\
     & MSD Prostate & T2, ADC & PZ, TZ & 2 & 32 & 3.6 $\times$ 0.62 $\times$ 0.62 & 20 $\times$ 320 $\times$ 320 & 32 $\times$ 192 $\times$ 192 \\
     & ACDC\dev & MRI & RV, MLV, LVC & 3 & 200 & 10 $\times$ 1.56 $\times$ 1.56 & 9 $\times$ 256 $\times$ 216 & 32 $\times$ 256 $\times$ 224 \\
     & P158\dev & T2, ADC & PZ, TZ & 2 & 139 & 3 $\times$ 0.4 $\times$ 0.4 & 25 $\times$ 270 $\times$ 270 & 32 $\times$ 256 $\times$ 256 \\
    \midrule
    CT & MSD Liver & CT & liver, cancer & 2 & 131 & 1 $\times$ 0.77 $\times$ 0.77 & 432 $\times$ 512 $\times$ 512 & 160 $\times$ 160 $\times$ 160 \\
     & MSD Lung & CT & cancer & 1 & 63 & 1.24 $\times$ 0.79 $\times$ 0.79 & 252 $\times$ 512 $\times$ 512 & 128 $\times$ 160 $\times$ 160 \\
     & MSD Pancreas & CT & pancreas, cancer & 2 & 281 & 2.5 $\times$ 0.8 $\times$ 0.8 & 93 $\times$ 512 $\times$ 512 & 96 $\times$ 160 $\times$ 160 \\
     & MSD Hepatic Vessel & CT & Vessel, Tumour & 2 & 303 & 5 $\times$ 0.8 $\times$ 0.8 & 49 $\times$ 512 $\times$ 512 & 96 $\times$ 160 $\times$ 160 \\
     & MSD Spleen & CT & spleen & 1 & 41 & 5 $\times$ 0.79 $\times$ 0.79 & 90 $\times$ 512 $\times$ 512 & 96 $\times$ 160 $\times$ 160 \\
     & MSD Colon & CT & colon cancer primaries & 1 & 126 & 5 $\times$ 0.78 $\times$ 0.78 & 95 $\times$ 512 $\times$ 512 & 96 $\times$ 160 $\times$ 160 \\
    \bottomrule
\end{tabular}

}
\end{table*}

\section{Evaluation protocol and baselines}\label{supp:eval}
\noindent\textbf{Encoders.}
\Cref{tab:supp_arch} lists the architecture and size of every encoder and the split of parameters between encoder and decoder.

\noindent\textbf{Segmentation framework.}
All downstream experiments use nnU-Net~\cite{isensee2021nnunet} for preprocessing, augmentation, training, and inference, with nnU-Net's default intensity normalization for CT and MRI and a UNETR decoder~\cite{hatamizadeh2022unetr} attached to every encoder.
The UNETR decoder taps four evenly spaced encoder layers.
Both regimes train for 150 epochs, the schedule of previous works~\cite{nnfoundation,wald2025revisitingmae}, with AdamW and a linear warm-up over the first 5 epochs.
The decoder (the segmentation head) is trained with a learning rate of $3\cdot10^{-4}$.
After the warm-up, the learning rate follows a polynomial decay with exponent 0.9, the weight decay is a constant $5\cdot10^{-2}$, and no deep supervision is used. Except the UNETR decoder, all of the settings are inherited from previous works~\cite{nnfoundation,wald2025revisitingmae}.

\noindent\textbf{Evaluation protocol and splits.}
We follow the evaluation protocols of OpenMind~\cite{wald2025openmind}, Revisiting MAE~\cite{wald2025revisitingmae}, and nnFoundation~\cite{nnfoundation}.
Each downstream dataset is split once into two halves: 50\% of the cases are used for training and the other 50\% for validation, and all reported DSC scores are computed on the validation half.
The split is identical for all methods and both regimes.

\noindent\textbf{Input spacing and resolution.}
For each dataset, we use the nnU-Net 3D full-resolution configuration (target spacing and input patch size) and share it across all models (\cref{tab:supp_datasets}).
To accommodate the $16^3$ patch size of 3DINO, every axis of the input patch is rounded up to a multiple of 16 and the batch size is 2 for all datasets.

\noindent\textbf{Frozen-encoder regime.}
The entire encoder, including the patch embedding and the final norm, is fixed and only the decoder is trained from scratch.
For datasets with $C_\mathrm{in}>1$ input modalities, the pretrained patch-embedding weights are replicated across the input channels, rescaled by $1/C_\mathrm{in}$ so that the response to identical inputs is unchanged, and kept frozen.

\noindent\textbf{Full fine-tuning regime.}
All parameters are trained for 150 epochs with the same optimizer, warm-up, and decoder learning rate as in the frozen regime.
The encoder is trained with a learning rate of $10^{-4}$, and all weights of the patch embedding with $10^{-5}$, \ie, always $0.1\times$ the encoder learning rate. nnFoundation and TriPlane-Ref full finetuning experiments were performed on 122GiB NVIDIA DGX Spark.
All other experiments were done on NVIDIA A40/L40/L40S including pretraining.

\noindent\textbf{Random encoder.}
The random encoder is a Primus-M with random initialization, matching the SSL baselines.
In the frozen regime, it is a sanity check of whether a task requires learned features. It is trained three times, and we report the mean of the three runs.
In the fine-tuning regime, all of its parameters are trained with the same schedule and learning rates as the pretrained encoders, which amounts to training from scratch.

\noindent\textbf{SSL baselines.}
All SSL baselines use a Primus-M encoder pretrained on OpenMind, and we refer to the OpenMind benchmark~\cite{wald2025openmind} for their pretraining details.
We use the officially released weights from HuggingFace and do not pretrain them ourselves.

\noindent\textbf{Larger encoders.}
nnFoundation~\cite{nnfoundation} is a Primus-style transformer pretrained with MAE on about 2.1 million 3D CT, MRI, and PET volumes from 125 institutional and public datasets collected within the Human Radiome Project.
3DINO~\cite{3dino} is pretrained with a 3D adaptation of DINOv2 on about 100{,}000 CT and MRI scans covering more than 10 organs.
We use the released checkpoints.

\section{Extended Results}\label{supp:full}

Frozen and fully fine-tuned per-dataset results are reported in \cref{tab:main,tab:large} of the main paper.

\subsection{Fine-tuned and Frozen performance Gap}\label{supp:gap}
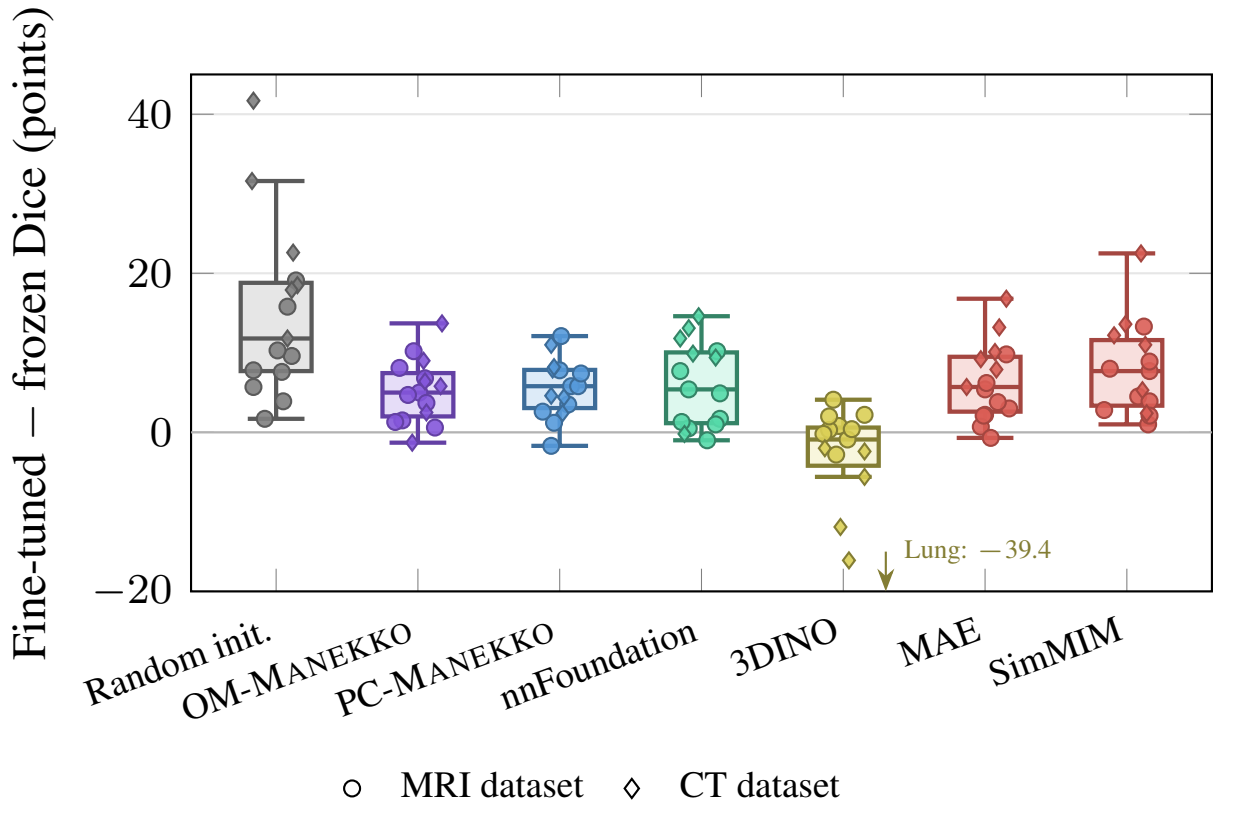
\begin{figure}[t]
    \centering
    \begin{tikzpicture}
    \begin{axis}[
        name=gap,
        width=\linewidth, height=0.6\linewidth,
        boxplot/draw direction=y,
        xtick={1,2,3,4,5,6,7},
        xticklabels={Random init., \omours, \pcours, nnFoundation, 3DINO, MAE, SimMIM},
        x tick label style={font=\scriptsize, rotate=20, anchor=north east},
        ylabel={Fine-tuned $-$ frozen Dice (points)},
        xmin=0.4, xmax=7.6, ymin=-20, ymax=45,
        ymajorgrids, grid style={gray!20},
        tick label style={font=\footnotesize},
        label style={font=\small},
    ]
        \addplot[gray!60, domain=0.4:7.6, samples=2] {0};
        \addplot[boxplot prepared={median=11.80, upper quartile=18.80, lower quartile=7.70, upper whisker=31.60, lower whisker=1.70, draw position=1, box extend=0.5}, draw=grijsD, fill=grijs!20, thick] coordinates {};
        \addplot[only marks, mark=*, mark size=1.5pt, mark options={draw=grijsD, fill=grijs}, fill opacity=0.9] coordinates {(1.05,3.9) (0.92,1.7) (0.84,7.8) (0.84,5.7) (1.11,9.6) (1.14,19.1) (1.04,7.6) (1.08,15.8) (1.01,10.3)};
        \addplot[only marks, mark=diamond*, mark size=1.5pt, mark options={draw=grijsD, fill=grijs}, fill opacity=0.9] coordinates {(1.15,18.5) (1.11,17.9) (0.83,31.6) (1.12,22.6) (0.84,41.7) (1.08,11.8)};
        \addplot[boxplot prepared={median=5.00, upper quartile=7.45, lower quartile=2.00, upper whisker=13.70, lower whisker=-1.30, draw position=2, box extend=0.5}, draw=paarsD, fill=paars!20, thick] coordinates {};
        \addplot[only marks, mark=*, mark size=1.5pt, mark options={draw=paarsD, fill=paars}, fill opacity=0.9] coordinates {(1.89,1.5) (2.12,0.6) (2.01,5.0) (1.93,4.7) (1.97,10.2) (1.84,1.3) (1.87,8.1) (2.06,3.7) (2.05,6.8)};
        \addplot[only marks, mark=diamond*, mark size=1.5pt, mark options={draw=paarsD, fill=paars}, fill opacity=0.9] coordinates {(2.04,9.0) (1.96,-1.3) (2.17,13.7) (2.16,5.8) (2.06,2.5) (2.05,6.4)};
        \addplot[boxplot prepared={median=5.80, upper quartile=7.85, lower quartile=3.05, upper whisker=12.10, lower whisker=-1.70, draw position=3, box extend=0.5}, draw=blauwD, fill=blauw!20, thick] coordinates {};
        \addplot[only marks, mark=*, mark size=1.5pt, mark options={draw=blauwD, fill=blauw}, fill opacity=0.9] coordinates {(3.06,3.5) (2.96,1.2) (2.88,2.6) (3.08,5.8) (3.01,12.1) (2.94,-1.7) (3.00,7.8) (3.13,5.8) (3.15,7.4)};
        \addplot[only marks, mark=diamond*, mark size=1.5pt, mark options={draw=blauwD, fill=blauw}, fill opacity=0.9] coordinates {(2.95,7.9) (3.02,2.4) (2.94,11.0) (3.03,4.4) (2.94,4.6) (2.96,8.2)};
        \addplot[boxplot prepared={median=5.40, upper quartile=10.05, lower quartile=1.15, upper whisker=14.60, lower whisker=-1.00, draw position=4, box extend=0.5}, draw=groenD, fill=groen!20, thick] coordinates {};
        \addplot[only marks, mark=*, mark size=1.5pt, mark options={draw=groenD, fill=groen}, fill opacity=0.9] coordinates {(4.13,1.7) (3.91,0.5) (4.04,-1.0) (3.86,1.3) (4.11,10.2) (4.10,1.0) (3.91,5.4) (4.13,4.9) (3.85,7.7)};
        \addplot[only marks, mark=diamond*, mark size=1.5pt, mark options={draw=groenD, fill=groen}, fill opacity=0.9] coordinates {(3.94,9.9) (3.88,-0.2) (3.98,14.6) (4.10,9.4) (3.91,13.1) (3.85,11.8)};
        \addplot[boxplot prepared={median=-0.90, upper quartile=0.60, lower quartile=-4.20, upper whisker=4.10, lower whisker=-5.60, draw position=5, box extend=0.5}, draw=geleD, fill=gele!20, thick] coordinates {};
        \addplot[only marks, mark=*, mark size=1.5pt, mark options={draw=geleD, fill=gele}, fill opacity=0.9] coordinates {(4.97,0.8) (4.90,0.3) (4.86,-0.2) (5.03,-0.9) (4.93,4.1) (5.06,0.4) (4.90,2.0) (5.15,2.2) (4.95,-2.8)};
        \addplot[only marks, mark=diamond*, mark size=1.5pt, mark options={draw=geleD, fill=gele}, fill opacity=0.9] coordinates {(4.87,-2.0) (5.04,-16.1) (5.15,-5.6) (4.98,-11.9) (5.15,-2.4)};
        \addplot[boxplot prepared={median=5.70, upper quartile=9.50, lower quartile=2.60, upper whisker=16.80, lower whisker=-0.70, draw position=6, box extend=0.5}, draw=rodeD, fill=rode!20, thick] coordinates {};
        \addplot[only marks, mark=*, mark size=1.5pt, mark options={draw=rodeD, fill=rode}, fill opacity=0.9] coordinates {(6.00,2.2) (5.97,0.7) (6.04,-0.7) (6.17,3.0) (6.15,9.8) (5.99,2.1) (6.09,3.8) (6.00,5.4) (6.01,6.2)};
        \addplot[only marks, mark=diamond*, mark size=1.5pt, mark options={draw=rodeD, fill=rode}, fill opacity=0.9] coordinates {(6.10,13.2) (5.97,9.2) (6.08,7.9) (6.07,10.1) (6.15,16.8) (5.87,5.7)};
        \addplot[boxplot prepared={median=7.70, upper quartile=11.60, lower quartile=3.35, upper whisker=22.50, lower whisker=1.00, draw position=7, box extend=0.5}, draw=rodeD, fill=rode!20, thick] coordinates {};
        \addplot[only marks, mark=*, mark size=1.5pt, mark options={draw=rodeD, fill=rode}, fill opacity=0.9] coordinates {(7.08,4.5) (7.15,1.0) (7.16,3.9) (6.84,2.8) (7.12,13.3) (7.16,2.1) (7.16,7.7) (6.88,8.0) (7.16,8.9)};
        \addplot[only marks, mark=diamond*, mark size=1.5pt, mark options={draw=rodeD, fill=rode}, fill opacity=0.9] coordinates {(7.13,11.0) (7.11,5.3) (6.99,13.6) (6.91,12.2) (7.10,22.5) (7.14,2.4)};
        \draw[-{Stealth[length=1.5mm]}, geleD] (axis cs:5.3,-15) -- (axis cs:5.3,-20) node[right, font=\tiny, pos=0] {Lung: $-$39.4};
    \end{axis}
\end{tikzpicture}

\begin{tikzpicture}
    \begin{customlegend}[legend columns=2, legend style={font=\scriptsize, draw=none, column sep=2mm, at={(gap.north)}, anchor=south}, legend entries={MRI dataset, CT dataset}]
        \addlegendimage{only marks, mark=*, mark size=1.5pt, mark options={draw=black, fill=white}}
        \addlegendimage{only marks, mark=diamond*, mark size=1.8pt, mark options={draw=black, fill=white}}
    \end{customlegend}
\end{tikzpicture}
    \caption{Gap of frozen performance from full fine-tuning per dataset on all 15 datasets and for the random encoder, training from scratch minus frozen. Boxes show the median and quartiles over datasets, whiskers extend to 1.5 times the interquartile range, and each point is one dataset (circles: MRI, diamonds: CT).}
    \label{fig:supp_gap}
\end{figure}

\Cref{fig:supp_gap} shows, for each encoder evaluated on all 15 datasets, the distribution over datasets of the difference between its fine-tuned and frozen DSC. For the random encoder, this is the difference between training from scratch and the frozen random encoder.
\omours has the lowest median gap (5.0 points), followed by nnFoundation (5.4), MAE (5.7), \pcours (5.8), and SimMIM (7.7).
All of these are far below the gap of the random encoder (median 11.8, up to 41.7 on Spleen), so a lot of the fine-tuned performance of a pre-trained encoder is already available frozen.
On MRI, the gaps of all pre-trained encoders except 3DINO are similar (3.5--5.8 points on average), and GLI has the largest gap for all of them.
On CT, the frozen features of \ours are much closer to their fine-tuned performance (6.0 and 6.5 points on average for \omours and \pcours) than those of nnFoundation, MAE, and SimMIM (9.7--11.1).
3DINO is the only encoder with a negative median gap. The fine-tuning of larger encoders is unstable, and for 3DINO particularly on CT it ends below its frozen performance on all six CT datasets, most strongly on Lung ($-39.4$).

\subsection{Classification, Regression and Retrieval.}\label{supp:cls_reg_ret}
We evaluate frozen global representations across CT and MRI downstream tasks using linear probes for classification and regression, together with embedding-based retrieval. The CT benchmarks comprise COVID-19 and severity classification on STOIC, reported using balanced accuracy and AUROC, and lesion-group and cancer-status retrieval on the liver, lung, pancreas, and colon subsets of 3D-MIR. Retrieval is summarized using mean average precision (mAP), which measures how highly relevant cases are ranked across the full retrieval list, with higher values indicating better overall retrieval quality. The MRI benchmarks comprise autism classification on ABIDE, age regression on FOMO25 and IXI, MGMT classification on BraTS21, and lesion classification on ODELIA. Classification is reported using balanced accuracy and AUROC, while age regression is reported using RMSE, where lower values indicate smaller prediction error.

Across the CT tasks (Tab.~\ref{tab:global_ct}), \ours models are consistently among the strongest methods, with PanCancerCT Manekko achieving the best or near-best performance on most STOIC and 3D-MIR evaluations. The MRI results (Tab.~\ref{tab:global_mri}) are more varied, with different methods performing best across ABIDE, brain-age regression, BraTS21 MGMT, and ODELIA. However, the Manekko variants remain competitive across several of these tasks. Overall, the results show a clearer and more consistent advantage for \pcours on CT, while performance on MRI is more task-dependent.

\begin{table*}[t]
\centering
\small
\setlength{\tabcolsep}{3.5pt}
\caption{
Frozen-feature performance on CT benchmarks.
STOIC reports balanced accuracy and AUROC for COVID-19 and severity classification.
3D-MIR reports mean average precision (mAP) for lesion-group and cancer-flag retrieval across four organ-specific benchmarks.
Bold and underlined values follow the highlighting in the source results table.
}
\label{tab:global_ct}
\resizebox{\linewidth}{!}{
    \begin{tabular}{lcccccccccccc}
\toprule

& \multicolumn{4}{c}{STOIC}
& \multicolumn{2}{c}{Liver}
& \multicolumn{2}{c}{Lung}
& \multicolumn{2}{c}{Pancreas}
& \multicolumn{2}{c}{Colon} \\
\cmidrule(lr){2-5}
\cmidrule(lr){6-7}
\cmidrule(lr){8-9}
\cmidrule(lr){10-11}
\cmidrule(lr){12-13}

& \multicolumn{2}{c}{COVID Classification}
& \multicolumn{2}{c}{Severity}
& Lesion Group & Cancer Flag
& Lesion Group & Cancer Flag
& Lesion Group & Cancer Flag
& Lesion Group & Cancer Flag \\
\cmidrule(lr){2-3}
\cmidrule(lr){4-5}
\cmidrule(lr){6-6}
\cmidrule(lr){7-7}
\cmidrule(lr){8-8}
\cmidrule(lr){9-9}
\cmidrule(lr){10-10}
\cmidrule(lr){11-11}
\cmidrule(lr){12-12}
\cmidrule(lr){13-13}

Method
& Bal. Acc. $\uparrow$
& AUROC $\uparrow$
& Bal. Acc. $\uparrow$
& AUROC $\uparrow$
& mAP $\uparrow$
& mAP $\uparrow$
& mAP $\uparrow$
& mAP $\uparrow$
& mAP $\uparrow$
& mAP $\uparrow$
& mAP $\uparrow$
& mAP $\uparrow$ \\
\midrule

\rowcolor{oursrow}
\pcours
& $\mathbf{0.697 \pm 0.021}$
& \underline{$0.758 \pm 0.022$}
& \underline{$0.677 \pm 0.018$}
& $\mathbf{0.761 \pm 0.018}$
& \underline{$0.431$}
& \underline{$0.732$}
& $\mathbf{0.613}$
& $\mathbf{0.886}$
& $\mathbf{0.559}$
& $\mathbf{0.950}$
& $\mathbf{0.406}$
& $\mathbf{0.729}$ \\

\rowcolor{oursrow}
\omours
& \underline{$0.691 \pm 0.024$}
& $\mathbf{0.762 \pm 0.022}$
& $\mathbf{0.678 \pm 0.023}$
& \underline{$0.740 \pm 0.021$}
& $\mathbf{0.436}$
& $\mathbf{0.742}$
& $0.567$
& $0.834$
& \underline{$0.550$}
& \underline{$0.941$}
& \underline{$0.405$}
& \underline{$0.721$} \\

MAE
& $0.656 \pm 0.027$
& $0.717 \pm 0.031$
& $0.652 \pm 0.025$
& $0.727 \pm 0.020$
& $0.412$
& $0.711$
& $0.507$
& $0.761$
& $0.539$
& $0.932$
& $0.378$
& $0.674$ \\

SimMIM
& $0.649 \pm 0.023$
& $0.701 \pm 0.028$
& $0.634 \pm 0.042$
& $0.715 \pm 0.049$
& $0.401$
& $0.676$
& $0.469$
& $0.720$
& $0.532$
& $0.922$
& $0.361$
& $0.657$ \\

SimCLR
& $0.678 \pm 0.017$
& $0.735 \pm 0.028$
& $0.647 \pm 0.027$
& $0.715 \pm 0.030$
& $0.386$
& $0.660$
& $0.452$
& $0.707$
& $0.517$
& $0.902$
& $0.366$
& $0.655$ \\

VoCo
& $0.663 \pm 0.029$
& $0.714 \pm 0.035$
& $0.621 \pm 0.024$
& $0.703 \pm 0.011$
& $0.396$
& $0.675$
& $0.432$
& $0.696$
& $0.507$
& $0.894$
& $0.332$
& $0.612$ \\

3DINO
& $0.606 \pm 0.012$
& $0.652 \pm 0.018$
& $0.573 \pm 0.032$
& $0.649 \pm 0.025$
& $0.414$
& $0.703$
& \underline{$0.584$}
& \underline{$0.854$}
& $0.538$
& $0.930$
& $0.375$
& $0.687$ \\

nnFoundation
& $0.650 \pm 0.021$
& $0.712 \pm 0.026$
& $0.617 \pm 0.015$
& $0.700 \pm 0.039$
& $0.402$
& $0.690$
& $0.552$
& $0.819$
& \underline{$0.550$}
& \underline{$0.941$}
& $0.381$
& $0.685$ \\

\bottomrule
\end{tabular}

}
\end{table*}

\begin{table*}[t]
\centering
\small
\setlength{\tabcolsep}{5pt}
\caption{
Frozen-feature performance on MRI benchmarks.
ABIDE, BraTS21 MGMT, and ODELIA report classification performance, while Brain Age reports RMSE on the FOMO25 and IXI cohorts.
Lower RMSE is better and all other metrics are higher-is-better.
Bold and underlined values follow the highlighting in the source results table.
}
\label{tab:global_mri}
\resizebox{\linewidth}{!}{
\begin{tabular}{lcccccccc}
\toprule
& \multicolumn{2}{c}{ABIDE}
& Brain Age - FOMO25
& Brain Age - IXI
& \multicolumn{2}{c}{BraTS21 MGMT}
& \multicolumn{2}{c}{ODELIA} \\
\cmidrule(lr){2-3}
\cmidrule(lr){4-4}
\cmidrule(lr){5-5}
\cmidrule(lr){6-7}
\cmidrule(lr){8-9}

Method
& Bal. Acc. $\uparrow$
& AUROC $\uparrow$
& RMSE $\downarrow$
& RMSE $\downarrow$
& Bal. Acc. $\uparrow$
& AUROC $\uparrow$
& Bal. Acc. $\uparrow$
& Macro AUROC $\uparrow$ \\
\midrule

\rowcolor{oursrow}
\pcours
& $\mathbf{0.626 \pm 0.023}$
& $\mathbf{0.674 \pm 0.024}$
& $\mathbf{7.604 \pm 0.688}$
& $7.348 \pm 0.439$
& \underline{$0.563 \pm 0.047$}
& $0.571 \pm 0.051$
& $0.544 \pm 0.039$
& $0.745 \pm 0.033$ \\

\rowcolor{oursrow}
\omours
& $0.594 \pm 0.038$
& \underline{$0.655 \pm 0.028$}
& \underline{$7.637 \pm 0.528$}
& $\mathbf{6.994 \pm 0.628}$
& $0.558 \pm 0.010$
& $\mathbf{0.588 \pm 0.024}$
& $\mathbf{0.572 \pm 0.048}$
& $0.744 \pm 0.049$ \\

MAE
& \underline{$0.614 \pm 0.019$}
& $0.653 \pm 0.031$
& $8.494 \pm 0.229$
& $8.988 \pm 0.595$
& $0.509 \pm 0.040$
& $0.532 \pm 0.047$
& $\mathbf{0.572 \pm 0.043}$
& $\mathbf{0.761 \pm 0.040}$ \\

SimMIM
& $0.607 \pm 0.016$
& $0.641 \pm 0.026$
& $11.131 \pm 0.425$
& $10.709 \pm 0.514$
& $\mathbf{0.568 \pm 0.044}$
& \underline{$0.586 \pm 0.030$}
& \underline{$0.566 \pm 0.050$}
& $0.744 \pm 0.049$ \\

SimCLR
& $0.592 \pm 0.037$
& $0.635 \pm 0.024$
& $8.876 \pm 0.447$
& $9.222 \pm 0.577$
& $0.518 \pm 0.029$
& $0.559 \pm 0.042$
& $0.554 \pm 0.055$
& \underline{$0.750 \pm 0.054$} \\

VoCo
& $0.574 \pm 0.018$
& $0.599 \pm 0.030$
& $10.971 \pm 1.058$
& $11.716 \pm 0.641$
& $0.490 \pm 0.062$
& $0.482 \pm 0.059$
& $0.508 \pm 0.047$
& $0.693 \pm 0.052$ \\

3DINO
& $0.582 \pm 0.033$
& $0.606 \pm 0.019$
& $13.856 \pm 1.200$
& $12.839 \pm 0.658$
& $0.512 \pm 0.029$
& $0.516 \pm 0.024$
& $0.529 \pm 0.037$
& $0.698 \pm 0.029$ \\

nnFoundation
& $0.592 \pm 0.030$
& $0.617 \pm 0.053$
& $9.031 \pm 0.619$
& \underline{$7.344 \pm 0.257$}
& $0.525 \pm 0.048$
& $0.550 \pm 0.047$
& $0.526 \pm 0.063$
& $0.738 \pm 0.045$ \\

\bottomrule
\end{tabular}

}
\end{table*}

\subsection{Compute Burden.}\label{supp:burden}
\begin{figure}[t]
    \centering
    \begin{tikzpicture}
    \begin{axis}[
        ybar,
        bar width=8pt,
        width=\linewidth,
        height=0.5\linewidth,
        ylabel={Peak GPU memory (GiB)},
        symbolic x coords={{Manekko},{3DINO},{Primus-M},{ViT-S/14 - Slice},{ViT-S/14 - TriPlane},{nnFoundation}},
        xtick={{Manekko},{3DINO},{Primus-M},{ViT-S/14 - Slice},{ViT-S/14 - TriPlane},{nnFoundation}},
        x tick label style={font=\scriptsize, rotate=20, anchor=north east},
        tick label style={font=\footnotesize},
        label style={font=\small},
        enlarge x limits=0.15,
        ymin=0,
        ymax=82,
        ymajorgrids, grid style={gray!20},
        name=peakmem,
    ]
        \addplot[bar shift=-4.5pt, fill=grijs!70, draw=black] coordinates {
            ({Manekko},29.299208)};
        \addplot[bar shift=-4.5pt, fill=grijs!70, draw=black] coordinates {
            ({3DINO},31.916216)};
        \addplot[bar shift=-4.5pt, fill=grijs!70, draw=black] coordinates {
            ({Primus-M},37.830696)};
        \addplot[bar shift=-4.5pt, fill=grijs!70, draw=black] coordinates {
            ({ViT-S/14 - Slice},39.027564)};
        \addplot[bar shift=-4.5pt, fill=grijs!70, draw=black] coordinates {
            ({ViT-S/14 - TriPlane},57.826682)};
        \addplot[bar shift=-4.5pt, fill=grijs!70, draw=black] coordinates {
            ({nnFoundation},72.506709)};
        \addplot[bar shift=4.5pt, fill=gray!35, draw=gray!60] coordinates {
            ({Manekko},21.871716)};
        \addplot[bar shift=4.5pt, fill=gray!35, draw=gray!60] coordinates {
            ({3DINO},22.907857)};
        \addplot[bar shift=4.5pt, fill=gray!35, draw=gray!60] coordinates {
            ({Primus-M},22.528697)};
        \addplot[bar shift=4.5pt, fill=gray!35, draw=gray!60] coordinates {
            ({ViT-S/14 - Slice},29.419533)};
        \addplot[bar shift=4.5pt, fill=gray!35, draw=gray!60] coordinates {
            ({ViT-S/14 - TriPlane},29.419657)};
        \addplot[bar shift=4.5pt, fill=gray!35, draw=gray!60] coordinates {
            ({nnFoundation},24.188064)};
    \end{axis}
\end{tikzpicture}

\begin{tikzpicture}
    \begin{customlegend}[legend columns=2, legend style={font=\scriptsize, draw=none, column sep=2mm, at={(peakmem.north)}, anchor=south}, legend entries={Full fine-tuning, Frozen encoder}]
        \addlegendimage{area legend, fill=grijs!70, draw=black}
        \addlegendimage{area legend, fill=gray!35, draw=gray!60}
    \end{customlegend}
\end{tikzpicture}
    \caption{Memory required for one step using a input volume of size $160^3$ and a batch size of $2$, under full fine-tuning or frozen evaluation. nnFoundation and ViT-S/14 TriPlane requires $\geq$ 80GB memory device to full finetune at larger resolution settings.}
    \label{fig:supp_params_training}
\end{figure}
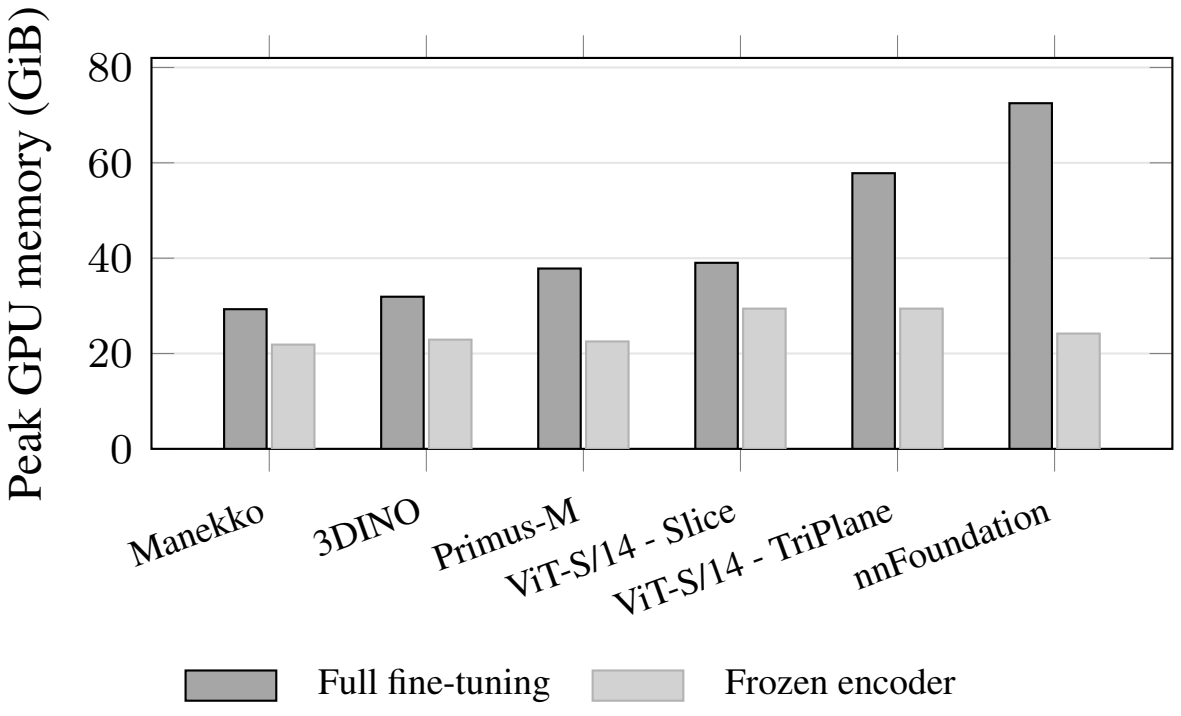

\Cref{tab:supp_arch} reports the number of trainable parameters and encoder settings with its UNETR decoder under frozen encoder and full-finetuning regime.
\Cref{fig:supp_params_training} reports the memory burden during training time under frozen encoder and full-finetuning training regimes.

\begin{table*}[t]
    \centering
    \caption{Encoder architectures and parameter counts with the UNETR decoder. Under frozen evaluation, only the decoder is trained; under full fine-tuning, encoder and decoder are both optimized. Since 3DINO uses a patch size of 16, the decoder is larger.}
    \label{tab:supp_arch}
    \small
\setlength{\tabcolsep}{5pt}
\begin{tabular}{l l c c c c rr rr}
    \toprule
    & & & & & & \multicolumn{2}{c}{\textbf{Params (M)}} & \multicolumn{2}{c}{\textbf{Trainable (M)}} \\
    \cmidrule(lr){7-8}\cmidrule(lr){9-10}
    Encoder & Used by & Layers & Width & Heads & Patch & Enc. & Dec. & Frozen & Fine-tuned \\
    \midrule
    ViT-S/14 & NeCo on DINOv2-R & 12 & 384 & 6 & $14^2$ & 22.06 & 25.25 & 25.25 & 47.31 \\
    Primus & \ours & 12 & 384 & 6 & $8^3$ & 68.43 & 6.50 & 6.50 & 74.94 \\
    Primus & SSL baselines, random init. & 16 & 864 & 12 & $8^3$ & 156.03 & 6.96 & 6.96 & 162.99 \\
    Primus & nnFoundation & 40 & 1056 & 16 & $8^3$ & 551.36 & 7.15 & 7.15 & 558.51 \\
    ViT-L/16 & 3DINO & 24 & 1024 & 16 & $16^3$ & 308.33 & 26.52 & 26.52 & 334.85 \\
    \bottomrule
\end{tabular}

\end{table*}

\section{Additional hyperparameter results}\label{supp:sens}

Unless stated otherwise, all runs pretrain \ours on OpenMind and are evaluated frozen on the developmental datasets.

\begin{table}[t]
    \centering
    \caption{\textbf{Data scaling}, frozen DSC on the developmental datasets when pretraining on a random fraction of OpenMind (\omours) or PanCancerCT (\pcours).
    }
    \label{tab:supp_sens}
    \resizebox{\linewidth}{!}{
        \small
\setlength{\tabcolsep}{4pt}
\begin{tabular}{l cccc cccc}
    \toprule
    & \multicolumn{4}{c}{\textbf{\omours}} & \multicolumn{4}{c}{\textbf{\pcours}} \\
    \cmidrule(lr){2-5}\cmidrule(lr){6-9}
    Fraction & ISLES & ACDC & P158 & Avg. & ISLES & ACDC & P158 & Avg. \\
    \midrule
    1/16 & {68.8} & {83.9} & {69.6} & 74.1 & {63.9} & {80.1} & {67.1} & 70.4 \\
    1/8  & {71.1} & {85.2} & {71.4} & 75.9 & {65.8} & {81.1} & {68.5} & 71.8 \\
    1/4  & {72.4} & {85.9} & {72.4} & 76.9 & {68.1} & {82.4} & {70.3} & 73.6 \\
    1/2  & {72.7} & {86.0} & {72.6} & 77.1 & {68.9} & {82.8} & {70.9} & 74.2 \\
    1    & 72.8 & 86.1 & 72.7 & 77.2 & 70.0 & 83.4 & 71.8 & 75.1 \\
    \bottomrule
\end{tabular}

    }
\end{table}

\begin{figure}[t]
    \centering
    \begin{tikzpicture}
    \begin{axis}[
        name=dur,
        width=\linewidth, height=0.6\linewidth,
        ylabel={Frozen Dice (\%)},
        ymin=60, ymax=90,
        grid=major, grid style={gray!20},
        tick label style={font=\footnotesize},
        label style={font=\small},
        xlabel={Pretraining steps (k)},
        xmin=0, xmax=65, xtick={5,10,20,30,40,60},
    ]
        \draw[gray!70!black, densely dotted, thick] (axis cs:30,60) -- (axis cs:30,90);
        \addplot[dev isles] coordinates {(5,64.1) (10,68.3) (20,72.6) (30,72.8) (40,73.1) (60,73.3)};
        \addplot[dev acdc]  coordinates {(5,77.8) (10,81.9) (20,84.0) (30,86.1) (40,86.4) (60,86.5)};
        \addplot[dev pros]  coordinates {(5,64.2) (10,68.6) (20,70.5) (30,72.7) (40,72.9) (60,72.6)};
    \end{axis}
\end{tikzpicture}

\begin{tikzpicture}
    \begin{customlegend}[legend columns=3, legend style={font=\scriptsize, draw=none, column sep=2mm, at={(dur.north)}, anchor=south}, legend entries={ISLES\dev, ACDC\dev, P158\dev}]
        \addlegendimage{dev isles}
        \addlegendimage{dev acdc}
        \addlegendimage{dev pros}
    \end{customlegend}
\end{tikzpicture}
    \caption{\textbf{Pre-training duration.} Frozen DSC of \ours on the developmental datasets after different numbers of pretraining steps. The dotted line marks the default of 30k steps.
    }
    \label{fig:supp_duration}
\end{figure}
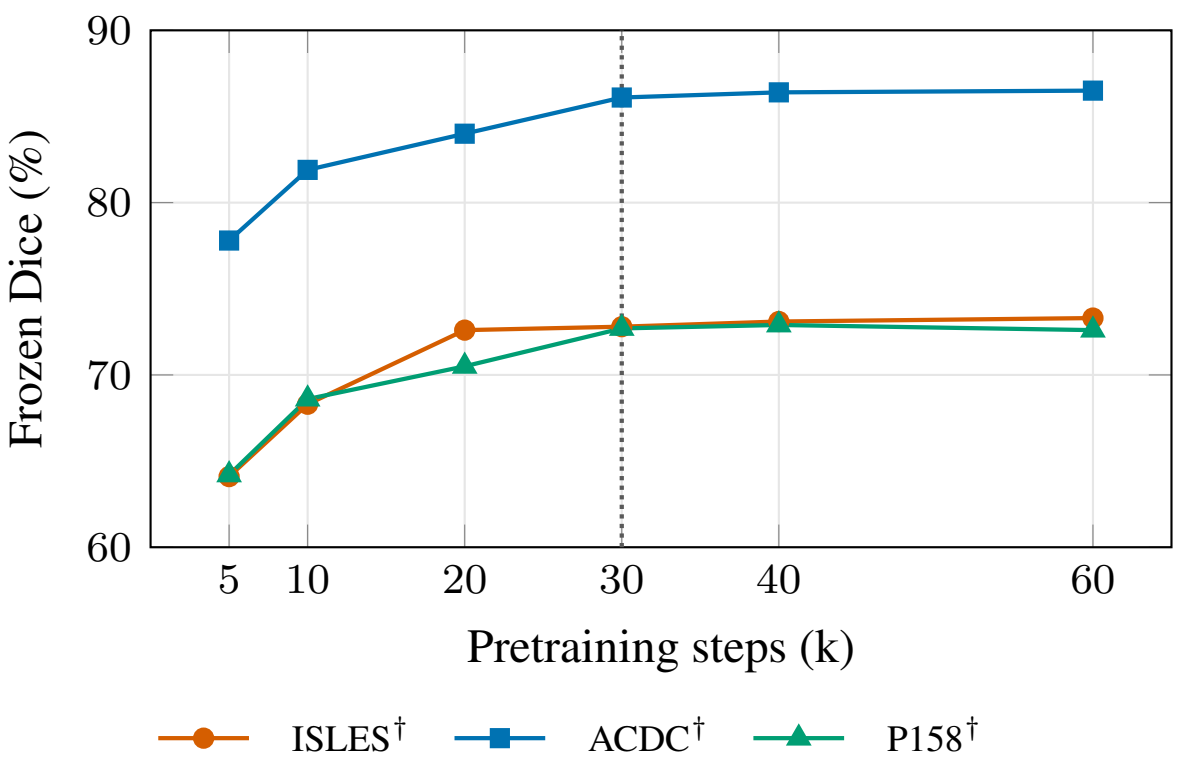

\noindent\textbf{Training duration.}
\Cref{fig:supp_duration} shows the frozen DSC of \omours on each developmental dataset as a function of the number of pretraining steps.
Most of the final performance is reached after 20k steps, and training beyond our default of 30k steps brings only marginal gains on all three datasets, so the target is matched quickly.

\begin{figure}[t]
    \centering
    \begin{tikzpicture}
    \begin{axis}[
        name=lam,
        width=\linewidth, height=0.6\linewidth,
        ylabel={Frozen Dice (\%)},
        ymin=60, ymax=90,
        grid=major, grid style={gray!20},
        tick label style={font=\footnotesize},
        label style={font=\small},
        xlabel={Affinity loss weight $\lambda_{\mathrm{aff}}$},
        xmode=log, log basis x=2,
        xmin=0.08, xmax=5,
        xtick={0.1,0.25,0.5,1,2,4}, xticklabels={0.1,0.25,0.5,1,2,4},
    ]
        \draw[gray!70!black, densely dotted, thick] (axis cs:1,60) -- (axis cs:1,90);
        \addplot[dev isles] coordinates {(0.1,70.9) (0.5,72.4) (1,72.8) (2,72.5) (4,71.6)};
        \addplot[dev acdc]  coordinates {(0.1,84.5) (0.5,85.9) (1,86.1) (2,85.8) (4,84.7)};
        \addplot[dev pros]  coordinates {(0.1,71.0) (0.5,72.4) (1,72.7) (2,72.3) (4,71.2)};
    \end{axis}
\end{tikzpicture}

\begin{tikzpicture}
    \begin{customlegend}[legend columns=3, legend style={font=\scriptsize, draw=none, column sep=2mm, at={(lam.north)}, anchor=south}, legend entries={ISLES\dev, ACDC\dev, P158\dev}]
        \addlegendimage{dev isles}
        \addlegendimage{dev acdc}
        \addlegendimage{dev pros}
    \end{customlegend}
\end{tikzpicture}
    \caption{\textbf{Loss weighting.} Frozen DSC of \omours on the developmental datasets for different weights $\lambda_{\mathrm{aff}}$ of the affinity loss. The dotted line marks the default of 1. 
    }
    \label{fig:supp_lambda}
\end{figure}
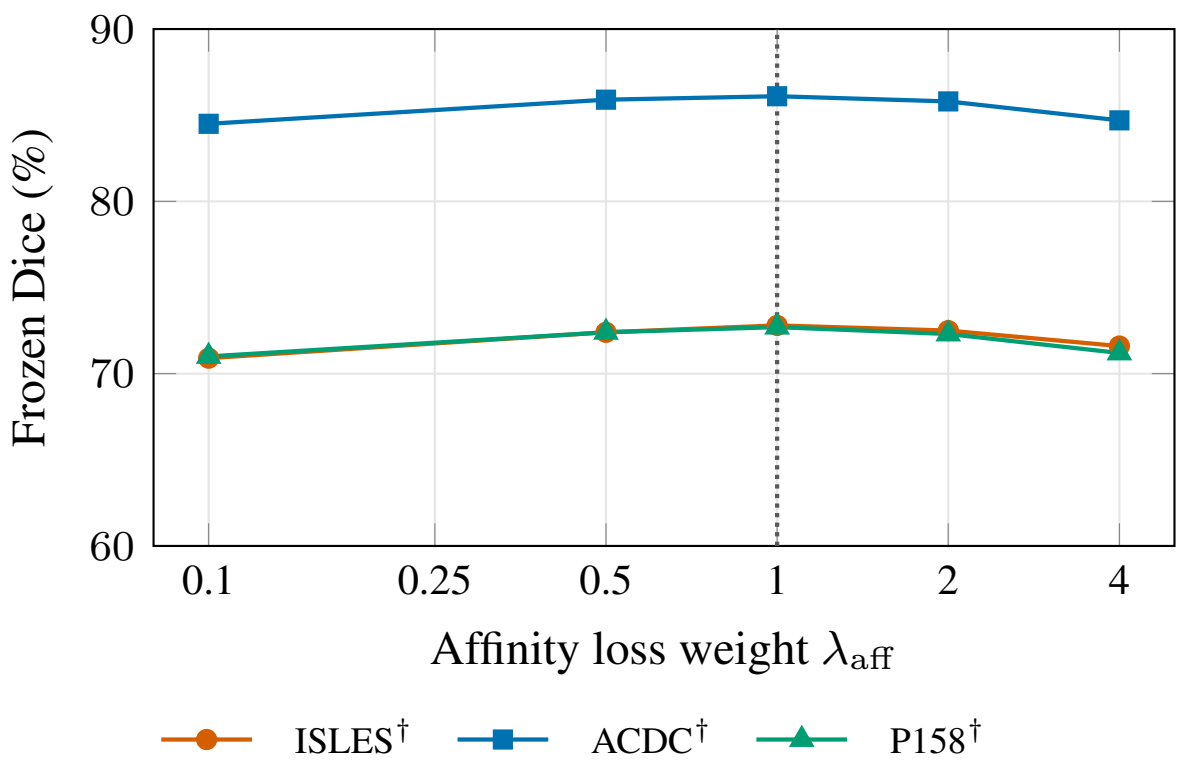

\noindent\textbf{Loss weighting.}
We weight the affinity term as $\mathcal{L} = \mathcal{L}_{\mathrm{patch}} + \lambda_{\mathrm{aff}}\,\mathcal{L}_{\mathrm{aff}}$ and vary $\lambda_{\mathrm{aff}}$ around our default of 1 (\cref{fig:supp_lambda}).
Performance is flat between $\lambda_{\mathrm{aff}}=0.5$ and $2$ on all three datasets and drops only at the extremes.

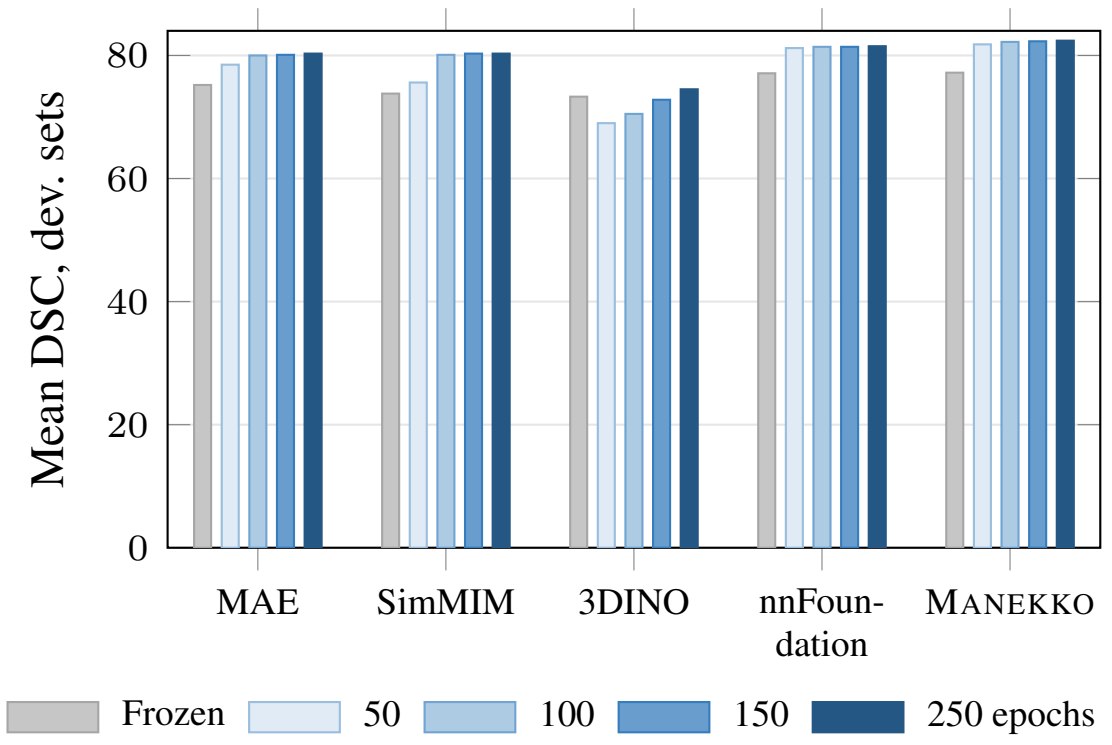
\begin{figure}[t]
    \centering
    \begin{tikzpicture}
    \begin{axis}[
        name=ftdur,
        width=0.93\linewidth, height=0.6\linewidth,
        ybar, bar width=3.2pt,
        ymin=0, ymax=84,
        ylabel={Mean DSC, dev.\ sets},
        symbolic x coords={MAE,SimMIM,DINO,nnF,OM},
        xtick=data,
        xticklabels={MAE, SimMIM, 3DINO, {nnFoun-\\dation}, {\ours}},
        x tick label style={font=\scriptsize, align=center},
        tick label style={font=\scriptsize},
        label style={font=\small},
        ymajorgrids, grid style={gray!20},
        enlarge x limits=0.12,
    ]
        \addplot[fill=gray!45, draw=gray!70] coordinates {
            (MAE,75.2) (SimMIM,73.8) (DINO,73.3) (nnF,77.1) (OM,77.2)
        };
        \addplot[fill=cvprblue!15, draw=cvprblue!50] coordinates {
            (MAE,78.5) (SimMIM,75.6) (DINO,69.0) (nnF,81.2) (OM,81.8)
        };
        \addplot[fill=cvprblue!40, draw=cvprblue!70] coordinates {
            (MAE,80.0) (SimMIM,80.1) (DINO,70.5) (nnF,81.4) (OM,82.2)
        };
        \addplot[fill=cvprblue!75, draw=cvprblue] coordinates {
            (MAE,80.1) (SimMIM,80.3) (DINO,72.8) (nnF,81.4) (OM,82.3)
        };
        \addplot[fill=cvprblue!70!black, draw=cvprblue!70!black] coordinates {
            (MAE,80.3) (SimMIM,80.3) (DINO,74.5) (nnF,81.5) (OM,82.4)
        };
    \end{axis}
\end{tikzpicture}

\begin{tikzpicture}
    \begin{customlegend}[legend columns=5, legend style={font=\scriptsize, draw=none, column sep=0.8mm, at={(ftdur.north)}, anchor=south}, legend entries={Frozen, 50, 100, 150, {250 epochs}}]
        \addlegendimage{area legend, fill=gray!45, draw=gray!70}
        \addlegendimage{area legend, fill=cvprblue!15, draw=cvprblue!50}
        \addlegendimage{area legend, fill=cvprblue!40, draw=cvprblue!70}
        \addlegendimage{area legend, fill=cvprblue!75, draw=cvprblue}
        \addlegendimage{area legend, fill=cvprblue!70!black, draw=cvprblue!70!black}
    \end{customlegend}
\end{tikzpicture}
    \caption{
        \textbf{Fine-tuning duration.} Mean DSC on the developmental datasets for the frozen encoder and after full fine-tuning for 50, 100, 150, and 250 epochs, each with a 5-epoch warm-up.
    }
    \label{fig:supp_ftduration}
\end{figure}

\noindent\textbf{Full Finetuning Duration}
Full fine-tuning has to be long enough for the pretrained encoder to adapt, and a pretraining objective can look better or worse depending on the schedule.
We therefore sweep the number of fine-tuning epochs over 50, 100, 150, and 250 for MAE, SimMIM, 3DINO, nnFoundation, and \ours on the developmental datasets (\cref{fig:supp_ftduration}).
Every schedule uses the same 5-epoch warm-up and the learning rates of \cref{supp:eval}.
MAE, SimMIM, nnFoundation, and \omours already saturate at around 50 epochs and exceed their frozen performance.
At 250 epochs, MAE, SimMIM, and nnFoundation perform similarly, whereas \ours maintains the lead at almost every duration.
3DINO struggles with fine-tuning at every duration and improves on its frozen performance only at 250 epochs.
All fine-tuning results in this paper use 150 epochs, which is also the schedule of nnFoundation~\cite{nnfoundation} and Revisiting MAE~\cite{wald2025revisitingmae}.
This result shows the challenge of full fine-tuning pretrained models and therefore further motivates the requirement for better frozen encoder performance.

\section{Reference models}\label{supp:reference}

\Cref{tab:supp_reference} lists the 2D reference models used in the main paper, their in-context scene understanding on natural images, and the frozen results of \omours trained with each on the developmental datasets.
In-context scene understanding is measured with the open Hummingbird implementation~\cite{balazevic2023hummingbird,pariza2024hbeval} on Pascal VOC and ADE20K, always with the full memory bank.
ImageNet-1k linear-probe accuracies are taken from the original publications where available. For Leopart, CrIBo, and NeCo, which do not report them, we run the DINOv2 linear evaluation protocol~\cite{oquab2024dinov2} with its learning-rate sweep.
DINO, Leopart, and CrIBo are pretrained on ImageNet-1k, whereas DINOv2, DINOv2-R, and both NeCo variants are pretrained on LVD-142M (NeCo post-training additionally uses COCO).
Our default reference model is NeCo applied to DINOv2-R.

\begin{table*}[t]
    \centering
    \caption{\textbf{Reference models}, their in-context scene understanding (Hummingbird mIoU, \%, full memory bank) and ImageNet linear-probe accuracy (\%), and the frozen DSC of the corresponding students on the developmental datasets.
    The highlighted row is the default reference model.
    }
    \label{tab:supp_reference}
    \small
\setlength{\tabcolsep}{4pt}
\begin{tabular}{l l cc c ccc >{\bfseries}c}
    \toprule
    & & \multicolumn{2}{c}{\textbf{Hummingbird}} & \textbf{ImageNet} & \multicolumn{4}{c}{\textbf{\ours, frozen DSC}} \\
    \cmidrule(lr){3-4}\cmidrule(lr){5-5}\cmidrule(lr){6-9}
    Reference model & Backbone & VOC & ADE20K & Linear & ISLES & ACDC & P158 & Avg. \\
    \midrule
    DINO~\cite{caron2021dino}              & ViT-S/16 & 48.7 & 17.9 & 77.0 & {62.7} & {80.6} & {64.9} & 69.4 \\
    Leopart~\cite{ziegler2022leopart}      & ViT-S/16 & 64.5 & 23.9 & 76.8 & {65.8} & {82.3} & {67.3} & 71.8 \\
    CrIBo~\cite{lebailly2024cribo}         & ViT-S/16 & 72.4 & 26.6 & 77.2 & {67.3} & {83.2} & {68.5} & 73.0 \\
    DINOv2~\cite{oquab2024dinov2}          & ViT-S/14 & 77.0 & 38.9 & 84.3 & {68.5} & {83.8} & {69.4} & 73.9 \\
    DINOv2-R~\cite{darcet2024registers}    & ViT-S/14 & 78.8 & 39.5 & 84.8 & {70.5} & {84.8} & {70.9} & 75.4 \\
    NeCo (DINOv2-XR)~\cite{pariza2025neco} & ViT-S/14 & 80.2 & 39.8 & 84.5 & {72.3} & {85.8} & {72.3} & 76.8 \\
    \rowcolor{oursrow}
    NeCo (DINOv2-R)~\cite{pariza2025neco}  & ViT-S/14 & 80.7 & 40.9 & 84.9 & 72.8 & 86.1 & 72.7 & 77.2 \\
    \bottomrule
\end{tabular}

\end{table*}

\begin{figure}[t]
    \centering
    \begin{tikzpicture}
    \begin{groupplot}[
        group style={group size=2 by 2, horizontal sep=8mm, vertical sep=13mm},
        width=0.56\linewidth, height=0.5\linewidth,
        ymin=68, ymax=79,
        grid=major, grid style={gray!20},
        tick label style={font=\scriptsize},
        label style={font=\scriptsize},
        title style={font=\scriptsize, yshift=-1mm},
        every axis plot/.append style={mark size=1.6pt},
    ]
        \nextgroupplot[xlabel={Hummingbird VOC (\%)}, ylabel={Frozen Dice (\%)}, title={$r=0.94$, $\rho=1.00$}]
        \addplot[only marks, mark=*, mark options={m om}, color=paarsD] coordinates {(48.7,69.4) (64.5,71.8) (72.4,73.0) (77.0,73.9) (78.8,75.4) (80.2,76.8) (80.7,77.2)};
        \addplot[gray, dashed, domain=46:83] {0.225*x + 57.78};
        \nextgroupplot[xlabel={Hummingbird ADE20K (\%)}, title={$r=0.94$, $\rho=1.00$}]
        \addplot[only marks, mark=*, mark options={m om}, color=paarsD] coordinates {(17.9,69.4) (23.9,71.8) (26.6,73.0) (38.9,73.9) (39.5,75.4) (39.8,76.8) (40.9,77.2)};
        \addplot[gray, dashed, domain=16:42] {0.277*x + 64.922};
        \nextgroupplot[xlabel={ImageNet linear probe (\%)}, ylabel={Frozen Dice (\%)}, title={$r=0.86$, $\rho=0.93$}]
        \addplot[only marks, mark=*, mark options={m om}, color=paarsD] coordinates {(77.0,69.4) (76.8,71.8) (77.2,73.0) (84.3,73.9) (84.8,75.4) (84.5,76.8) (84.9,77.2)};
        \addplot[gray, dashed, domain=76:86] {0.58902*x + 26.00756};
        \nextgroupplot[group/empty plot]
    \end{groupplot}
\end{tikzpicture}
    \caption{\textbf{Which property of the reference model predicts the student?}
    Frozen DSC of \omours on the developmental datasets against in-context scene understanding (top) and ImageNet linear-probe accuracy (bottom) of its reference model, with Pearson $r$, Spearman $\rho$, and a least-squares fit.}
    \label{fig:supp_reference_corr}
\end{figure}
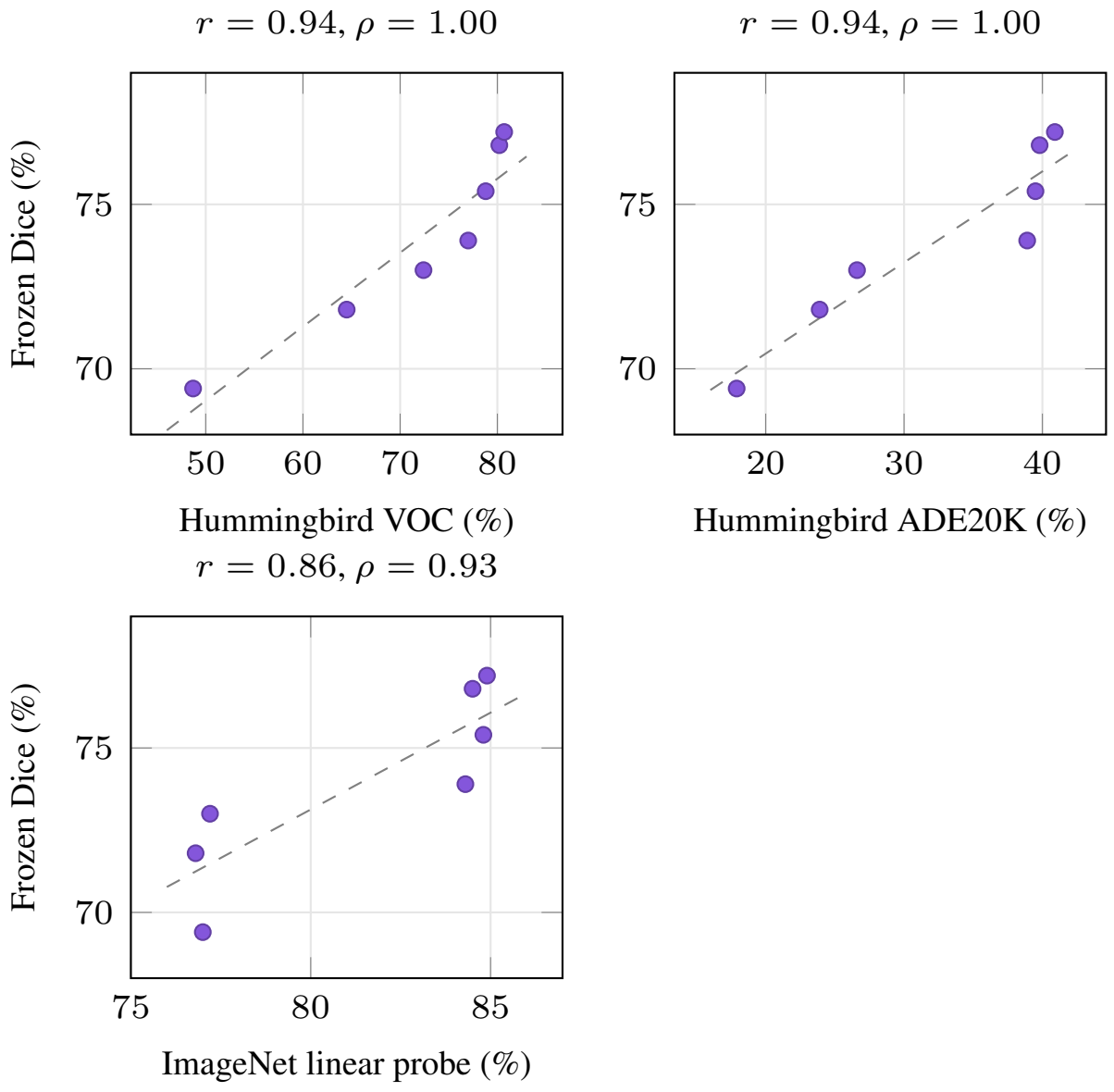

\noindent\textbf{Correlation with other metrics.}
\Cref{fig:supp_reference_corr} compares in-context scene understanding to global classification accuracy as a predictor of the student.
Hummingbird mIoU on both VOC and ADE20K correlates strongly with downstream DSC ($r=0.94$) and preserves the ranking of the reference models ($\rho=1.00$), whereas ImageNet linear-probe accuracy correlates less ($r=0.86$, $\rho=0.93$).
Linear-probe accuracy mainly separates the models by their pretraining data and barely changes within each group: Leopart and CrIBo are within 0.2 points of DINO on ImageNet (76.8 and 77.2 vs.\ 77.0) but improve the student by 2.4 and 3.6 DSC, and the NeCo variants are within 0.3 points of DINOv2-R but improve it by up to 1.8 DSC.
Objectives that target dense understanding thus improve in-context performance without improving global classification.
This indicates dense scene understanding, rather than global semantics, as the property of the 2D model that matters for \ours.

\section{Feature analysis}\label{supp:features}

\noindent\textbf{Global representation.}\label{supp:rank}
For each encoder and downstream dataset, we encode every volume of the held-out subset of the dataset with the frozen encoder.
Each volume is represented by one random resized crop (scale $0.35$--$1$, fixed seed) at the pretraining input size of the encoder, with one randomly chosen input modality.
The global representation is the CLS token for encoders that have one (\ours, 3DINO) and the mean patch token otherwise (Primus-M, nnFoundation).

\noindent\textbf{Normalized effective rank.}
Let $\mathbf{X}\in\mathbb{R}^{N\times C}$ stack the global representations of $N$ volumes, with singular values $\sigma_1\geq\dots\geq\sigma_K$, where $K=\min(N,C)$ is the largest possible rank.
With $p_i=\sigma_i/\sum_{j}\sigma_j$, the effective rank~\cite{roy2007effectiverank} and its normalized version are
\begin{align}
    \operatorname{erank}(\mathbf{X})&=\exp\Big(-\sum_{i=1}^{K}p_i\log p_i\Big),\\
    \overline{\operatorname{erank}}(\mathbf{X})&=\frac{\operatorname{erank}(\mathbf{X})}{K}\in\Big[\frac{1}{K},1\Big].
    \label{eq:erank}
\end{align}
A value near $1/K$ means that all volumes are mapped to nearly the same direction, and a value near one means that the volumes span all $K$ available directions with equal energy.
We use the global token rather than patch tokens because neighbouring patches are strongly correlated, which makes the effective rank of patch features low for every encoder other than a random encoder.

\noindent\textbf{Bootstrapping.}
Computed on all volumes of a dataset, $\overline{\operatorname{erank}}$ depends on $N$: the evaluation splits range from 10 (Heart) to 675 (GLI) volumes, and since the effective rank grows more slowly than $K=\min(N,C)$, larger splits receive lower values.
Values computed on all volumes are therefore not comparable across datasets.
We instead estimate it with a bootstrap: in each of $B=100$ replicates, we draw $n=10$ distinct volumes per dataset, compute \cref{eq:erank} on these $n$ vectors, so that $K=10$ for every dataset and encoder, and report the mean, standard deviation, and the 2.5th and 97.5th percentiles over the replicates as a 95\% interval.
\Cref{tab:supp_rank} reports the bootstrap means for all encoders.
The intervals are narrow compared to the differences between encoders, \eg, $0.48$ $[0.47, 0.49]$ for \omours against $0.30$ $[0.29, 0.30]$ for MAE.

\begin{table}[t]
    \centering
    \caption{\textbf{Normalized effective rank} of the frozen global representation, bootstrapped with 10 volumes per dataset and 100 replicates. Group columns average the per-dataset means. All 15 is the mean over all datasets with a 95\% percentile interval over replicates.}
    \label{tab:supp_rank}
    \small
\setlength{\tabcolsep}{4pt}
\begin{tabular}{l ccc c}
    \toprule
    Encoder & Brain & Body & CT & All 15 [95\% CI] \\
    \midrule
    Random init.     & 0.52 & 0.35 & 0.15 & 0.33 {\scriptsize[0.31, 0.34]} \\
    Models Genesis   & 0.43 & 0.41 & 0.22 & 0.34 {\scriptsize[0.33, 0.35]} \\
    Swin UNETR       & 0.47 & 0.50 & 0.22 & 0.38 {\scriptsize[0.37, 0.39]} \\
    SimCLR           & 0.52 & 0.64 & 0.27 & 0.45 {\scriptsize[0.44, 0.46]} \\
    VoCo             & 0.64 & 0.53 & 0.34 & 0.49 {\scriptsize[0.48, 0.50]} \\
    VF               & 0.27 & 0.31 & 0.34 & 0.31 {\scriptsize[0.30, 0.31]} \\
    SimMIM           & 0.33 & 0.40 & 0.21 & 0.30 {\scriptsize[0.29, 0.31]} \\
    MAE              & 0.32 & 0.33 & 0.26 & 0.30 {\scriptsize[0.29, 0.30]} \\
    \midrule
    3DINO            & 0.88 & 0.88 & 0.75 & 0.83 {\scriptsize[0.82, 0.84]} \\
    nnFoundation     & 0.42 & 0.42 & 0.30 & 0.37 {\scriptsize[0.36, 0.38]} \\
    \midrule
    \rowcolor{oursrow}
    \omours          & 0.57 & 0.42 & 0.45 & 0.48 {\scriptsize[0.47, 0.49]} \\
    \rowcolor{oursrow}
    \pcours          & 0.44 & 0.42 & 0.52 & 0.47 {\scriptsize[0.45, 0.48]} \\
    \bottomrule
\end{tabular}

\end{table}

\noindent\textbf{Effective rank does not predict performance.}
The normalized effective rank measures how evenly the global representations of different volumes spread over the available directions, not whether this spread is meaningful.
It does not order the encoders by their frozen performance.
3DINO has by far the highest rank (0.83) but its frozen performance is below \ours, nnFoundation, and MAE (60.6 average DSC over 15 datasets, \cref{tab:large}).
VoCo has a rank similar to \omours (0.49 vs.\ 0.48) but the lowest frozen DSC of all SSL baselines (65.7 on the nine MRI datasets, \cref{tab:main}), whereas MAE, the strongest SSL baseline, has one of the lowest ranks (0.30).
The random encoder has a rank on brain MRI (0.52) as high as SimCLR and higher than every other SSL baseline except VoCo, which shows that features without any learned structure can also be diverse in this sense.
A diverse global representation is therefore not sufficient for useful dense features, and therefore feature diversity can not be used to explain the frozen performance.

\section{CT intensity normalization for pretraining}\label{supp:pancancer}

For PanCancerCT pretraining, we normalize intensities with clinically motivated windows when the anatomy of a volume is known, and fall back to a globally estimated normalization otherwise (\cref{tab:supp_ct_norm}).
We never use the masks that are made available with the PanCancerCT dataset.
Each window clips Hounsfield units to $[p_{0.5}, p_{99.5}]$ and then standardizes with the given mean and standard deviation.

\begin{table}[h]
    \centering
    \caption{CT intensity normalization (HU) for PanCancerCT pretraining. ``Lung'' is a broad lung channel, ``Soft'' a soft-tissue channel, and ``Unknown'' uses a large window estimated over the entire PanCancerCT dataset.}
    \label{tab:supp_ct_norm}
    \small
\setlength{\tabcolsep}{4pt}
\begin{tabular}{l rrrr}
    \toprule
    Anatomy & Mean & Std & $p_{0.5}$ & $p_{99.5}$ \\
    \midrule
    Lung    & $-525.0$ & $475.0$ & $-950$ & $0$ \\
    Soft    & $80.0$ & $65.0$ & $-15$ & $175$ \\
    Unknown & $-62.0$ & $263.5$ & $-929$ & $703$ \\
    \bottomrule
\end{tabular}

\end{table}
The window is chosen from the anatomy label of each scan (\cref{fig:supp_pretrain_counts}b).
Scans labeled lung use the lung window (6{,}979 scans).
Scans labeled abdomen, pancreas, liver, kidney, spleen, colon, esophagus, adrenal, or lymph nodes use the soft window (7{,}602 scans).
All remaining scans, \ie, whole body, head and neck, pelvis, spine, and multi-region, use the global nnU-Net CT normalization (7{,}221 scans).

\section{Limitations and broader impact}\label{supp:limitations}

\noindent\textbf{Dependence on a 2D reference model.}
\ours requires a pretrained 2D natural-image model and inherits its biases and failure modes.
Its quality is bounded by how well the 2D model represents medical slices, and structures that are not distinguishable in any of the three planes cannot be learned from the target alone.
However, our results comparing \ours to directly using the reference model (TriPlane-Ref and Slice-Ref) show that the model can better make use of the 3D context than simply fusing planar-features into a 3D volume.

\noindent\textbf{Scope of evaluation.}
We pretrain on publicly accessible large datasets of brain MRI and multi-anatomy CT and evaluate segmentation on 15 datasets.
Larger encoders like nnFoundation and 3DINO were trained on private datasets that were inaccessible to us and at times evaluated on more tasks, anatomies, modalities, and in context of VLMs, and our comparison with them should be read within the presented scope.

\noindent\textbf{Broader impact.}
Better pretrained encoders can reduce the annotation effort required for medical image analysis, uncover potentially missed pathologies and provide better understanding of the organisation of information for medical vision foundation models.
Models trained with \ours are research artifacts and are not validated for clinical use.

\section{Pooled ranking analysis and robustness}
\label{supp:ranking_reloaded}

We complement the dataset-specific evaluation with an analysis of pooled segmentation performance using \emph{Rankings Reloaded} toolkit of Wiesenfarth et al.~\cite{Wiesenfarth2021}.
This framework distinguishes the observed ordering of algorithms from its sensitivity to the ranking rule and to the sampled evaluation cases.
We examine case-level rank distributions, alternative ranking rules, pairwise statistical comparisons, and bootstrap rank distributions.
Together, these analyses assess whether the leading methods remain competitive under different summaries of the same evaluation data and whether their positions are stable under case resampling.

\subsection{Pooling and evaluation settings}
For each algorithm, we concatenate the case-level foreground scores from all 15 evaluation datasets into a single list, preserving the correspondence between algorithms evaluated on the same case.
We analyze the Dice similarity coefficient (DSC) and normalized surface Dice (NSD) separately under frozen-encoder evaluation and full fine-tuning.
All NSD values use a surface-distance tolerance of $1\,\mathrm{mm}$.
In the frozen regime, only the decoder is trained and full fine-tuning updates both encoder and decoder, following the protocol in Sec.~\ref{supp:eval}.
The term \emph{task} in the exported figures refers to one of these four metric-regime combinations, rather than to an individual dataset or case.

The primary ranking averages scores over the pooled cases and then orders algorithms by decreasing mean score, with rank~1 denoting the best result.
For algorithm $a$, metric $m$, and regime $r$, the pooled score is
\begin{equation}
    \overline{s}_{a}^{(m,r)}
    = \frac{1}{N}\sum_{d=1}^{15}\sum_{i=1}^{n_d}s_{a,d,i}^{(m,r)},
    \qquad N=\sum_{d=1}^{15}n_d,
\end{equation}
where $n_d$ is the number of evaluation cases contributed by dataset $d$.
Thus, each case has equal weight, and datasets contribute in proportion to their number of cases.
This differs from averaging dataset-level scores or ranks with equal weight per dataset.
DSC and NSD are never averaged together, and the two adaptation regimes are ranked separately.

\subsection{Case-level ranks and sensitivity}
\begin{figure*}[p]
    \centering
    \begin{subfigure}[t]{0.48\textwidth}
        \centering
        \includegraphics[width=\linewidth]{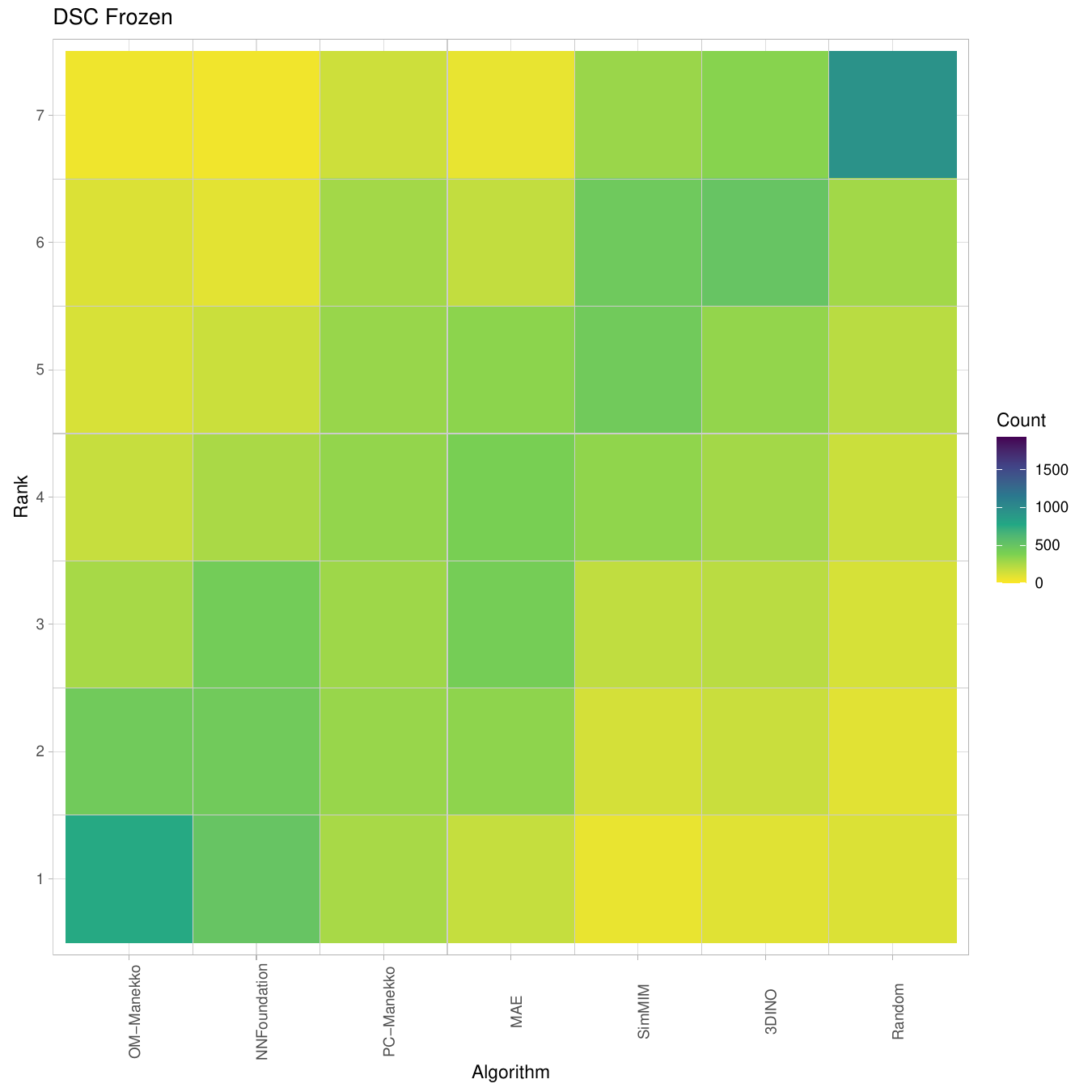}
        \caption{Frozen encoder: DSC.}
    \end{subfigure}\hfill
    \begin{subfigure}[t]{0.48\textwidth}
        \centering
        \includegraphics[width=\linewidth]{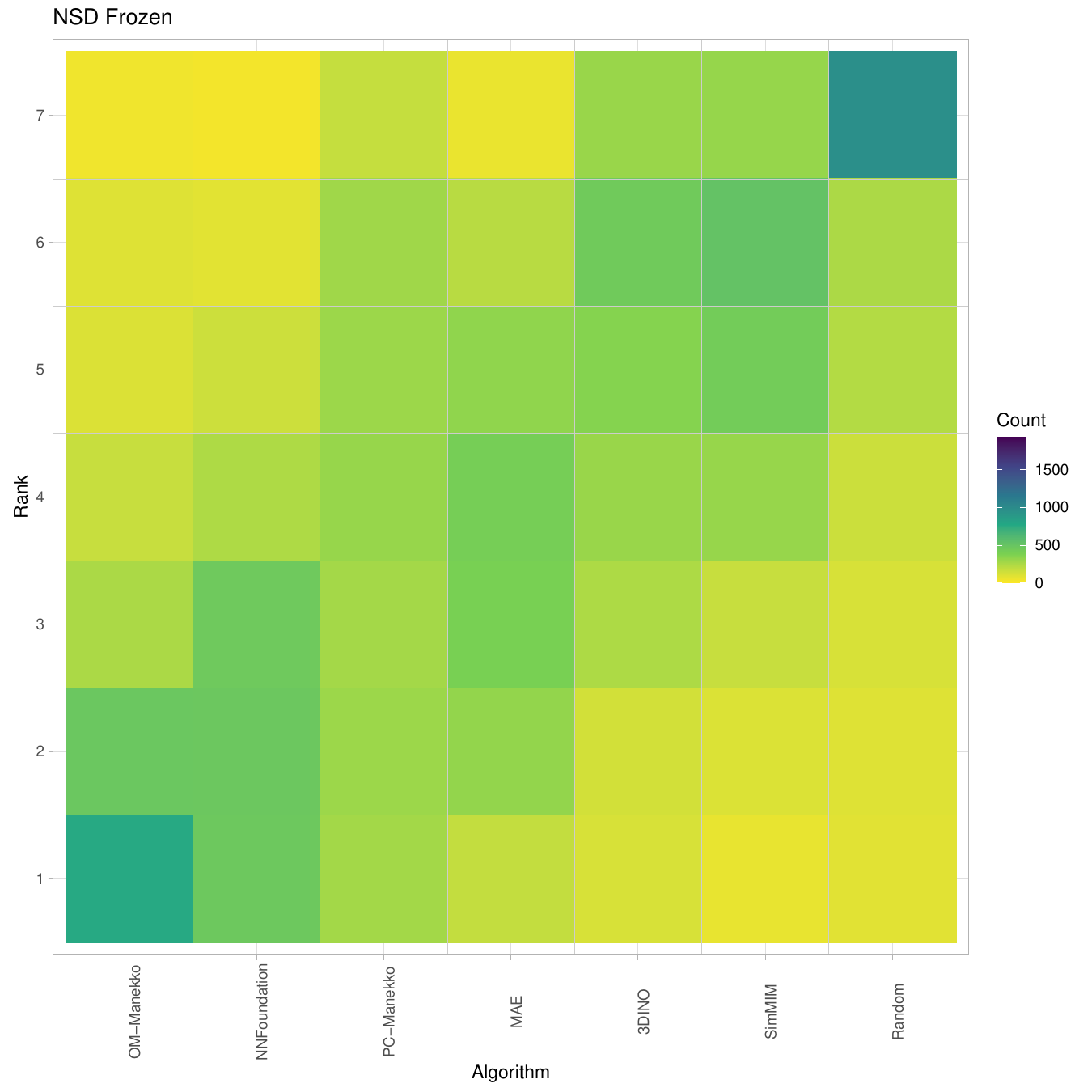}
        \caption{Frozen encoder: NSD.}
    \end{subfigure}
    \par\medskip
    \begin{subfigure}[t]{0.48\textwidth}
        \centering
        \includegraphics[width=\linewidth]{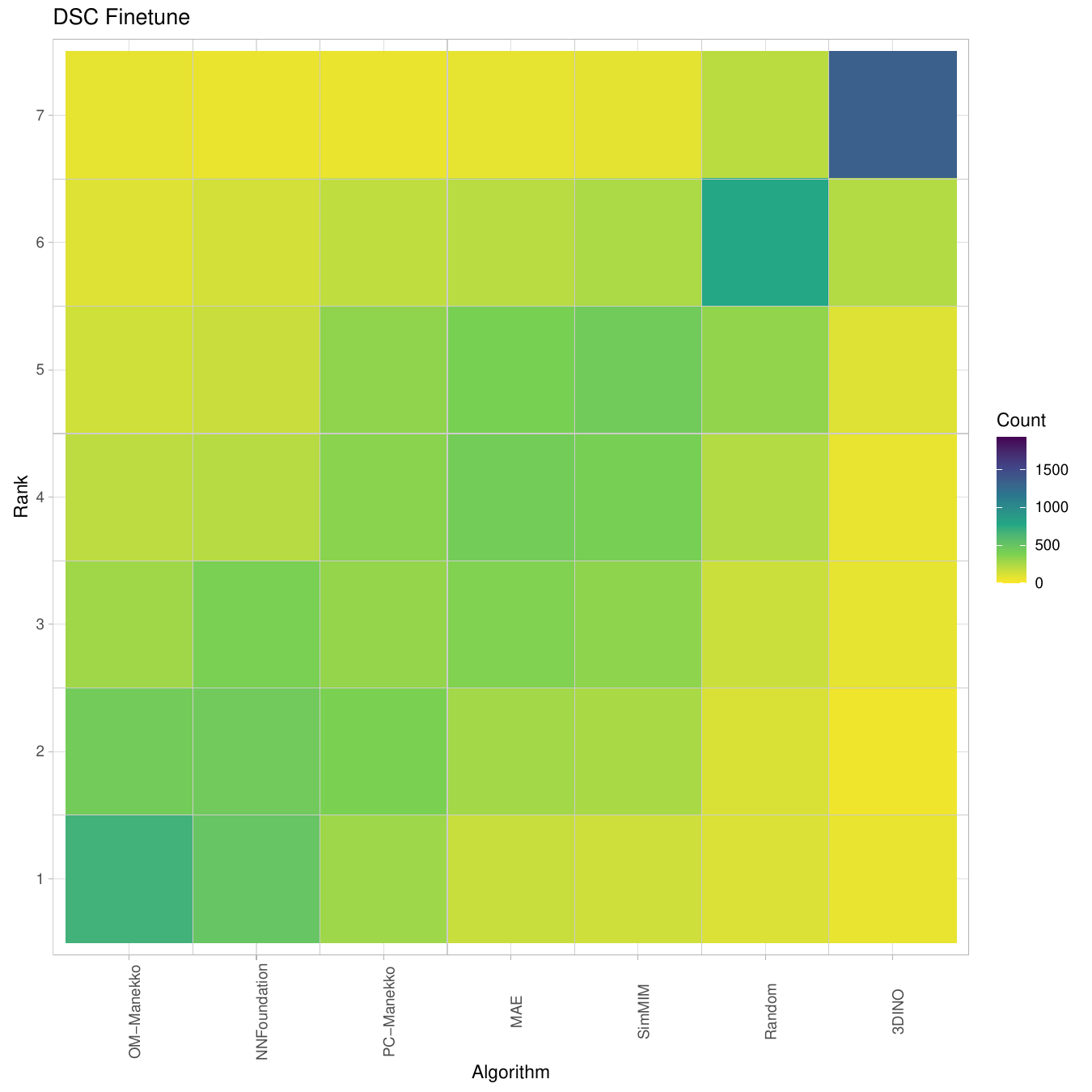}
        \caption{Full fine-tuning: DSC.}
    \end{subfigure}\hfill
    \begin{subfigure}[t]{0.48\textwidth}
        \centering
        \includegraphics[width=\linewidth]{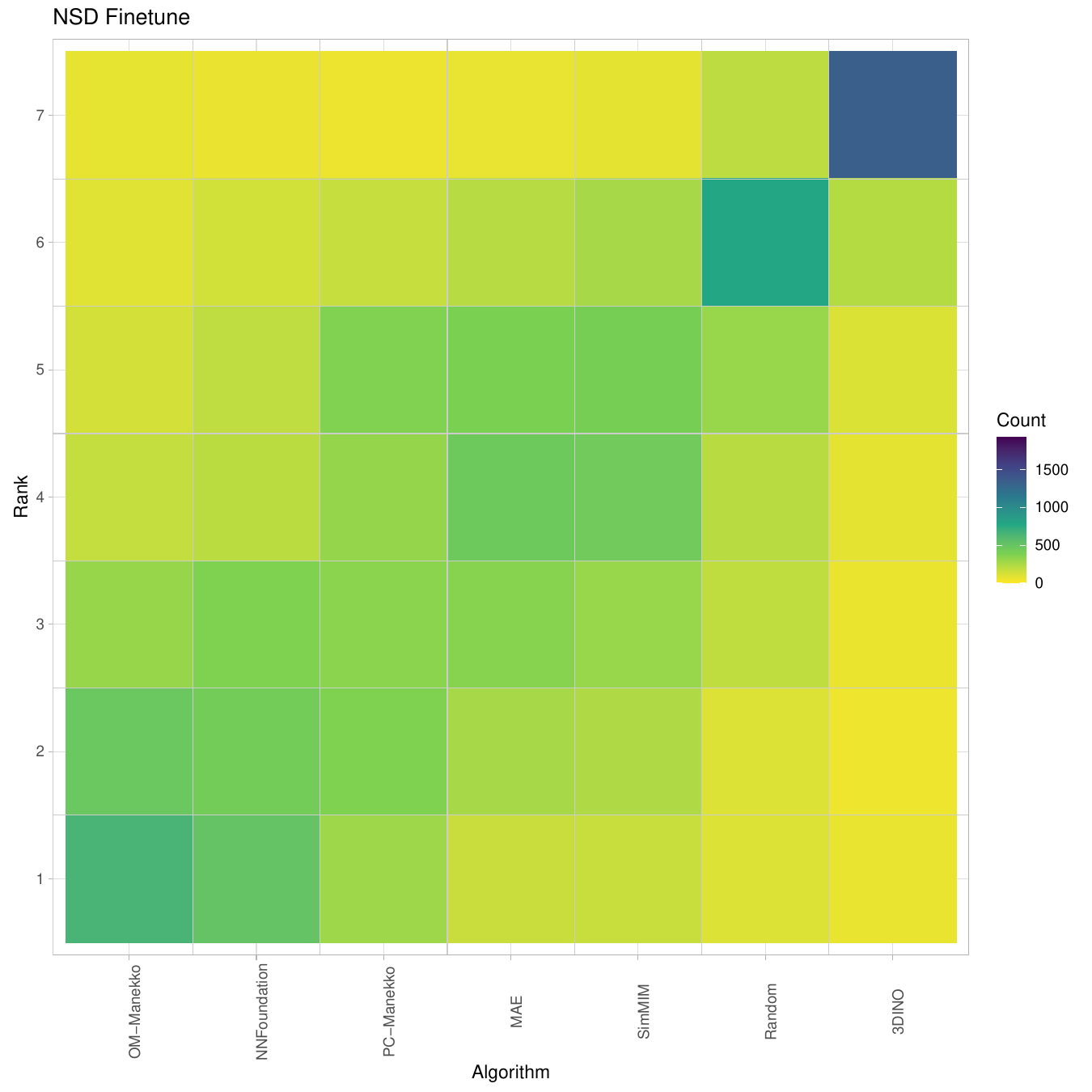}
        \caption{Full fine-tuning: NSD.}
    \end{subfigure}
    \caption{
    Rows show frozen-encoder evaluation and full fine-tuning and columns show DSC and NSD ($1\,\mathrm{mm}$ tolerance).
    Each cell gives the number of pooled evaluation cases on which the algorithm on the horizontal axis attains the rank on the vertical axis where rank~1 is best.
    Algorithms are ordered by their pooled mean score within each panel.
    Concentration at favourable ranks indicates consistently strong relative performance across cases, while a broad distribution reveals case-to-case variation.
    }
    \label{fig:rr_heatmaps}
\end{figure*}

Figure~\ref{fig:rr_heatmaps} shows how frequently each algorithm attains each rank when algorithms are compared on individual cases.

\begin{figure*}[p]
    \centering
    \includegraphics[width=0.95\textwidth]{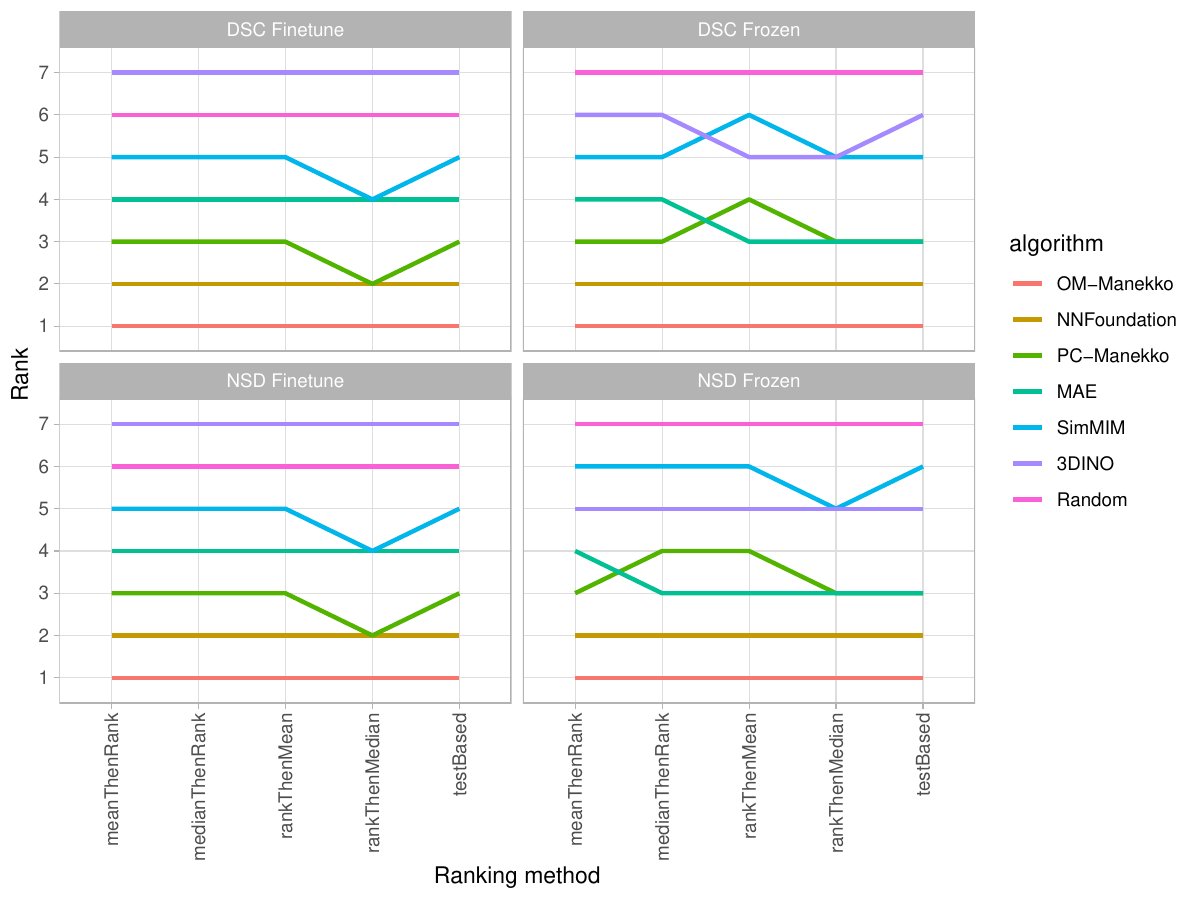}
    \caption{
    Each colored line tracks one algorithm across mean-then-rank, median-then-rank, rank-then-mean, rank-then-median, and test-based ranking.
    Lower ranks are better and horizontal lines indicate unchanged positions, and overlapping lines can indicate tied ranks.
    The panels show DSC on the top row and NSD ($1\,\mathrm{mm}$ tolerance) on the bottom row, with full fine-tuning on the left and frozen encoders on the right.
    \omours remains first under all displayed rules in all four settings.
    This figure assesses sensitivity to the ranking procedure, rather than sampling variability.
    }
    \label{fig:rr_ranking_rules}
\end{figure*}

Figure~\ref{fig:rr_ranking_rules} evaluates sensitivity to the choice of ranking rule.
The \texttt{meanThenRank} and \texttt{medianThenRank} rules first summarize each algorithm's case-level scores and then rank these summaries.
Conversely, \texttt{rankThenMean} and \texttt{rankThenMedian} first rank algorithms within each case and then aggregate those ranks, with smaller aggregate ranks preferred.
The \texttt{testBased} rule orders algorithms using the number of competitors shown to be significantly better in pairwise testing.
These rules emphasize different properties: score aggregation retains the magnitude of performance differences, whereas rank aggregation emphasizes relative ordering within cases.
Ties can occur, so a displayed position need not be unique.
\omours retains rank~1 under every displayed rule in all four settings, nnFoundation retains rank~2, although this position is shared under some rules.
The leading position of \omours is therefore not specific to the primary mean-then-rank summary.

\subsection{Paired statistical comparisons}

\begin{figure*}[p]
    \centering
    \begin{subfigure}[t]{0.48\textwidth}
        \centering
        \includegraphics[width=\linewidth]{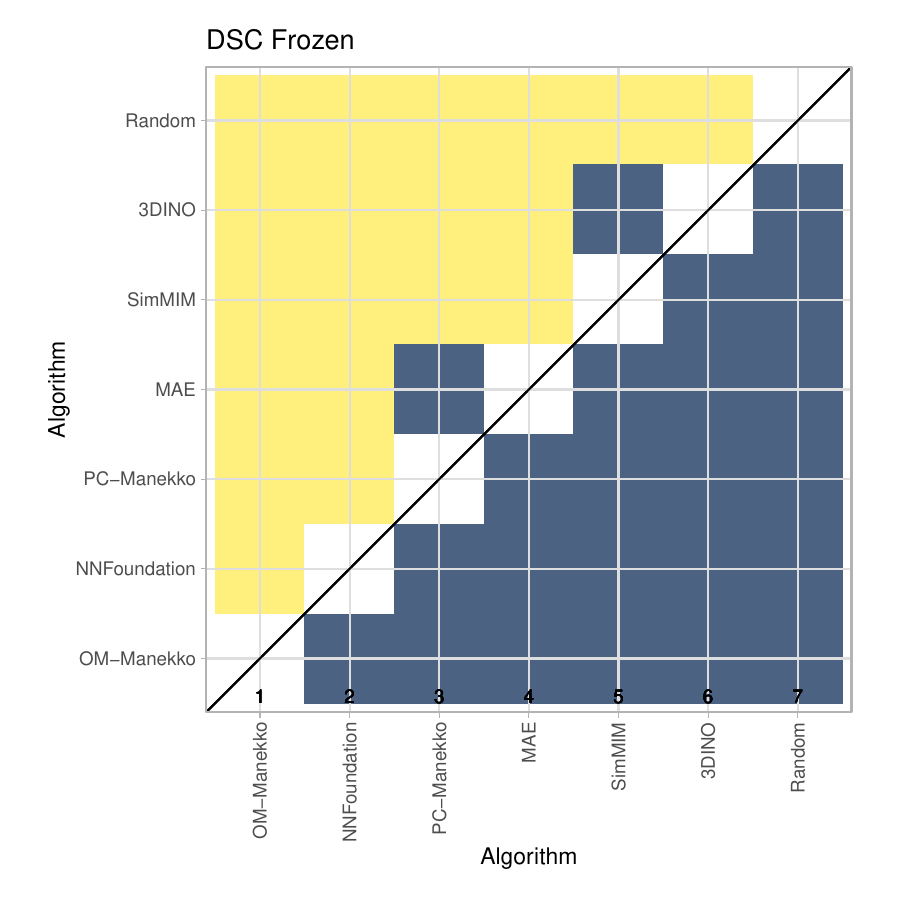}
        \caption{Frozen encoder: DSC.}
    \end{subfigure}\hfill
    \begin{subfigure}[t]{0.48\textwidth}
        \centering
        \includegraphics[width=\linewidth]{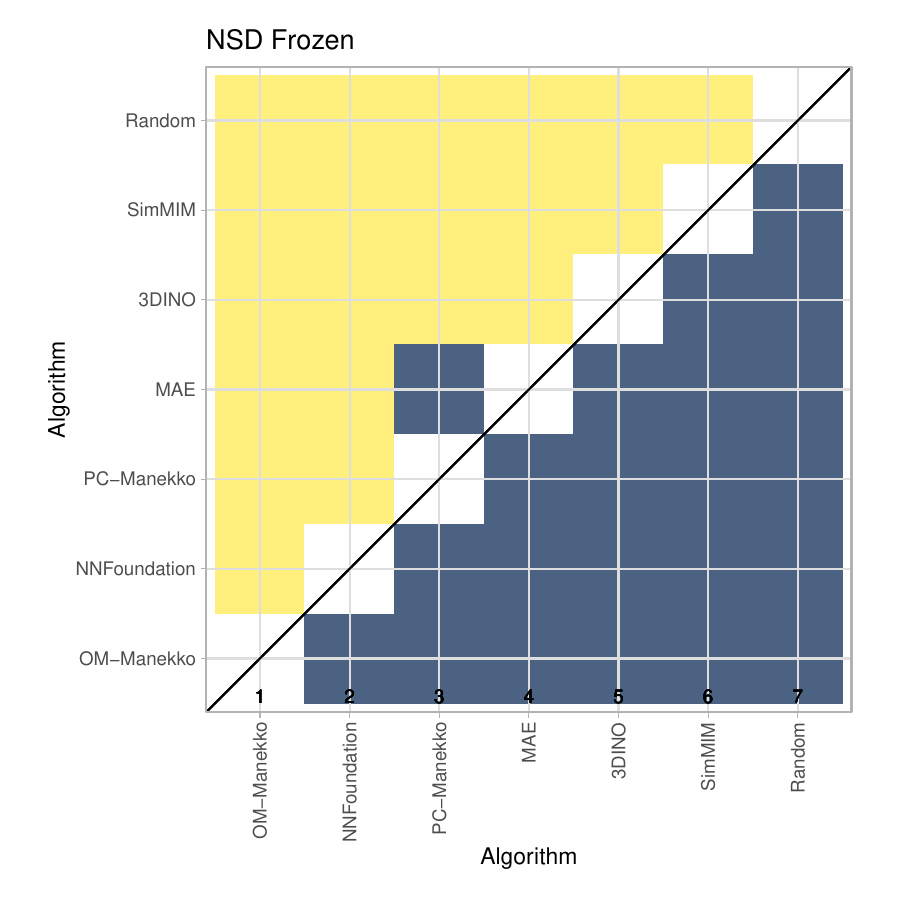}
        \caption{Frozen encoder: NSD.}
    \end{subfigure}
    \par\medskip
    \begin{subfigure}[t]{0.48\textwidth}
        \centering
        \includegraphics[width=\linewidth]{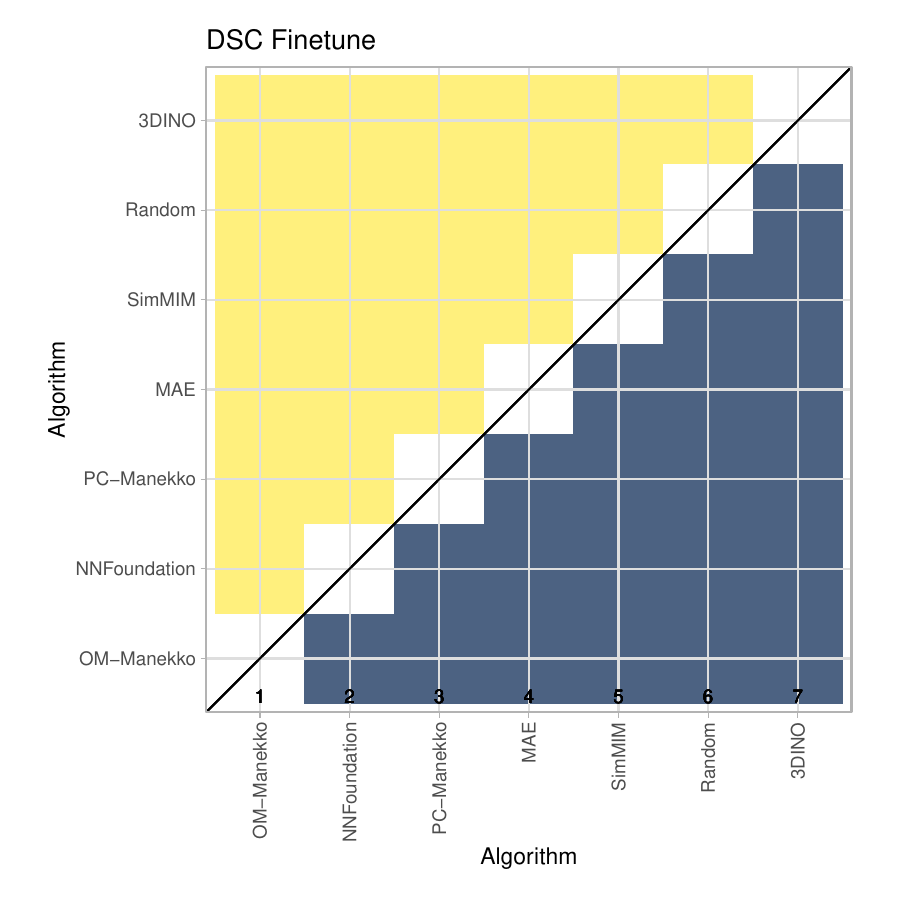}
        \caption{Full fine-tuning: DSC.}
    \end{subfigure}\hfill
    \begin{subfigure}[t]{0.48\textwidth}
        \centering
        \includegraphics[width=\linewidth]{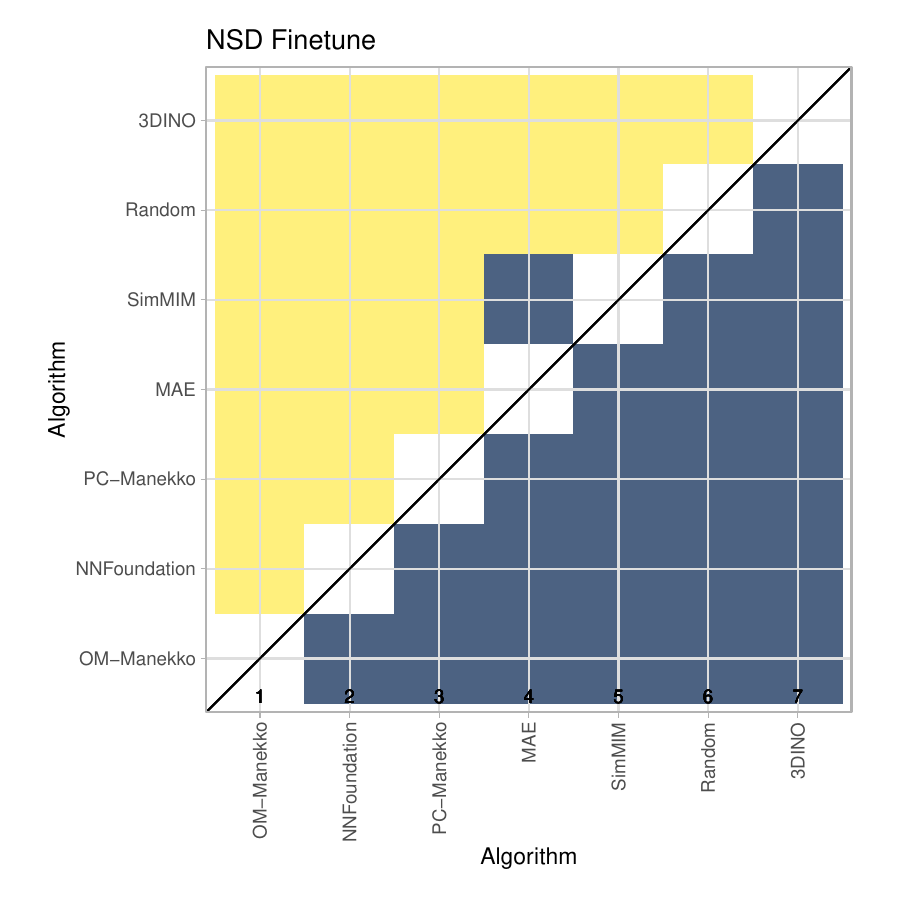}
        \caption{Full fine-tuning: NSD.}
    \end{subfigure}
    \caption{
    Yellow indicates that the algorithm on the horizontal axis is significantly superior to the algorithm on the vertical axis under a one-sided paired Wilcoxon signed-rank test with Holm correction at $\alpha=0.05$.
    Blue indicates that superiority in that direction was not established.
    Algorithms are ordered by pooled mean performance within each panel.}
    \label{fig:rr_significance}
\end{figure*}

Figure~\ref{fig:rr_significance} reports paired, one-sided Wilcoxon signed-rank tests on the case-level scores, with Holm correction for multiple comparisons at a significance level of $0.05$ within each setting.
The maps indicate significant superiority of \omours over each of the six comparators for both metrics and both regimes.
These comparisons concern paired case-level differences and are distinct from simply comparing pooled means.

\subsection{Bootstrap stability}

To assess sensitivity to the sampled cases, the report uses $B=1{,}000$ bootstrap replicates per setting.
Each replicate samples $N$ cases with replacement from the pooled list, retaining paired scores across algorithms, and recomputes the mean-then-rank ordering.
The resulting rank frequencies describe how often each aggregate position is recovered under case resampling.

\begin{figure*}[p]
    \centering
    \includegraphics[width=0.95\textwidth]{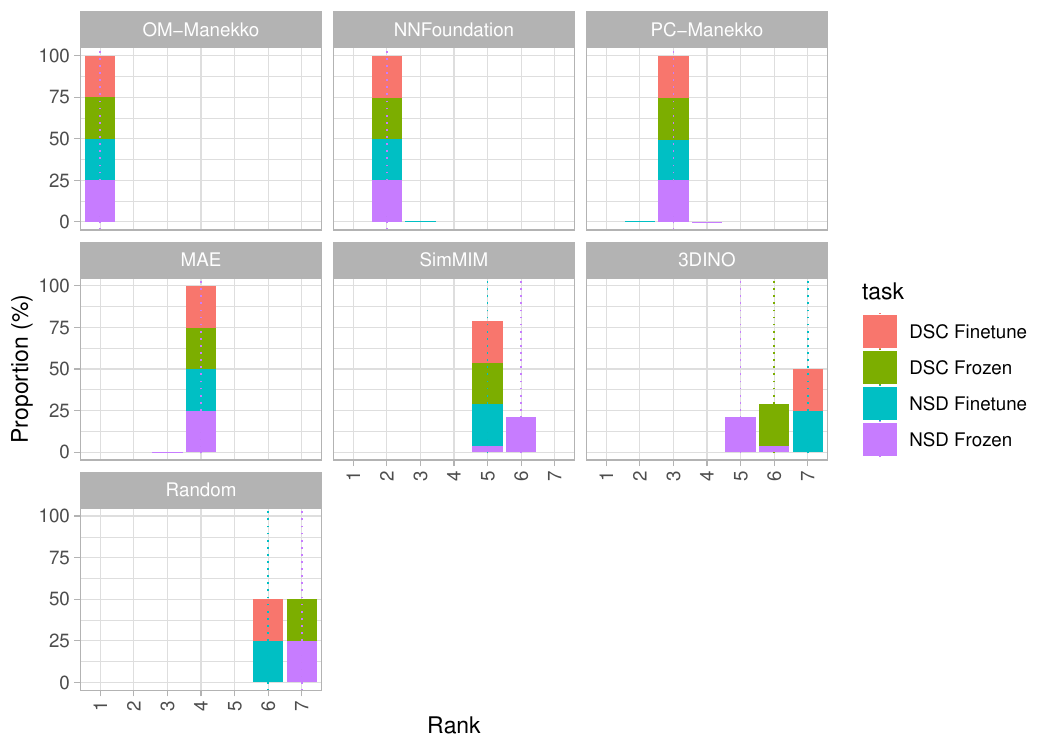}
    \caption{
    The horizontal axis gives the rank obtained after recomputing the pooled mean-based ranking in each bootstrap sample.
    Colours identify the four metric--regime settings. All NSD results use a $1\,\mathrm{mm}$ tolerance.
    The stacks are normalized over all four settings together: with $1{,}000$ replicates per setting, each color contributes $25\%$ of the total mass in an algorithm's panel.
    Thus, a $100\%$ bar at one rank indicates agreement across settings and bootstrap replicates, rather than a single setting's frequency.
    Colored vertical reference lines mark the corresponding full-data ranks and overlap when settings give the same rank.
    Splitting across ranks can reflect either differences between settings or bootstrap variability within a setting.}
    \label{fig:rr_stacked}
\end{figure*}

\begin{figure*}[p]
    \centering
    \includegraphics[width=0.95\textwidth]{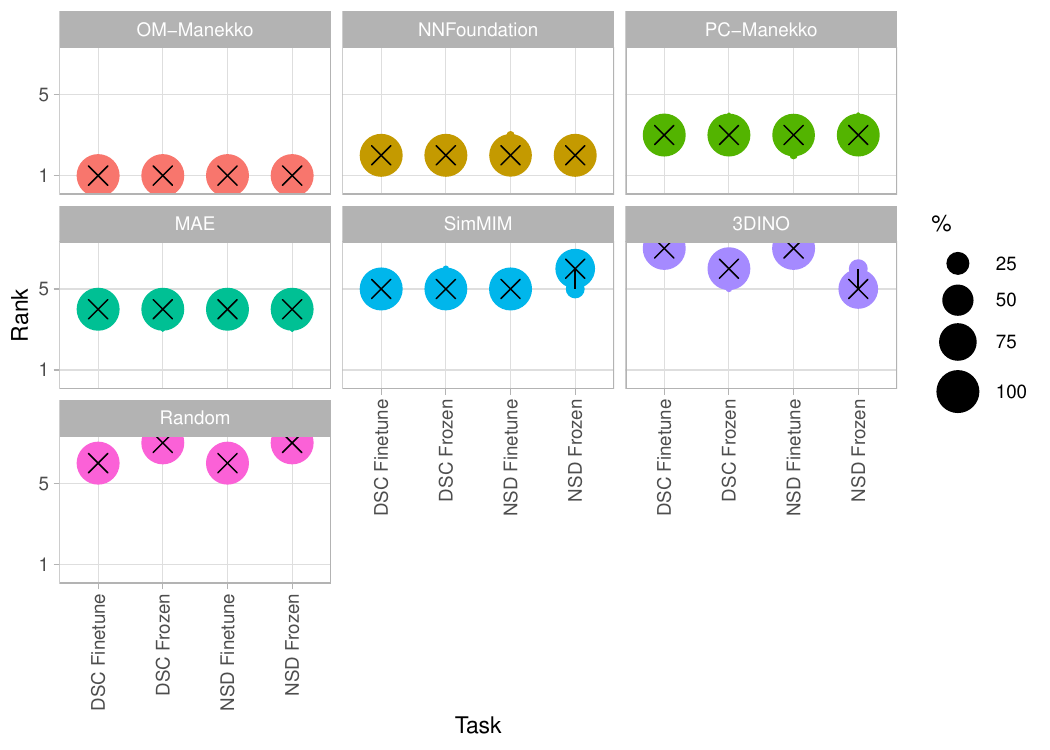}
    \caption{
    Each panel follows one algorithm across the four metric--regime settings on the horizontal axis.
    Circle area is proportional to the percentage of the $1{,}000$ bootstrap replicates in that setting attaining the indicated rank and colours identify algorithms.
    Black crosses mark median ranks and black vertical lines show the $2.5$th--$97.5$th percentile intervals, which can collapse to a single rank.
    This view separates within-setting rank uncertainty from shifts between settings.
    NSD uses a $1\,\mathrm{mm}$ tolerance.}
    \label{fig:rr_blobs_algorithm}
\end{figure*}

\begin{figure*}[p]
    \centering
    \includegraphics[width=0.88\textwidth]{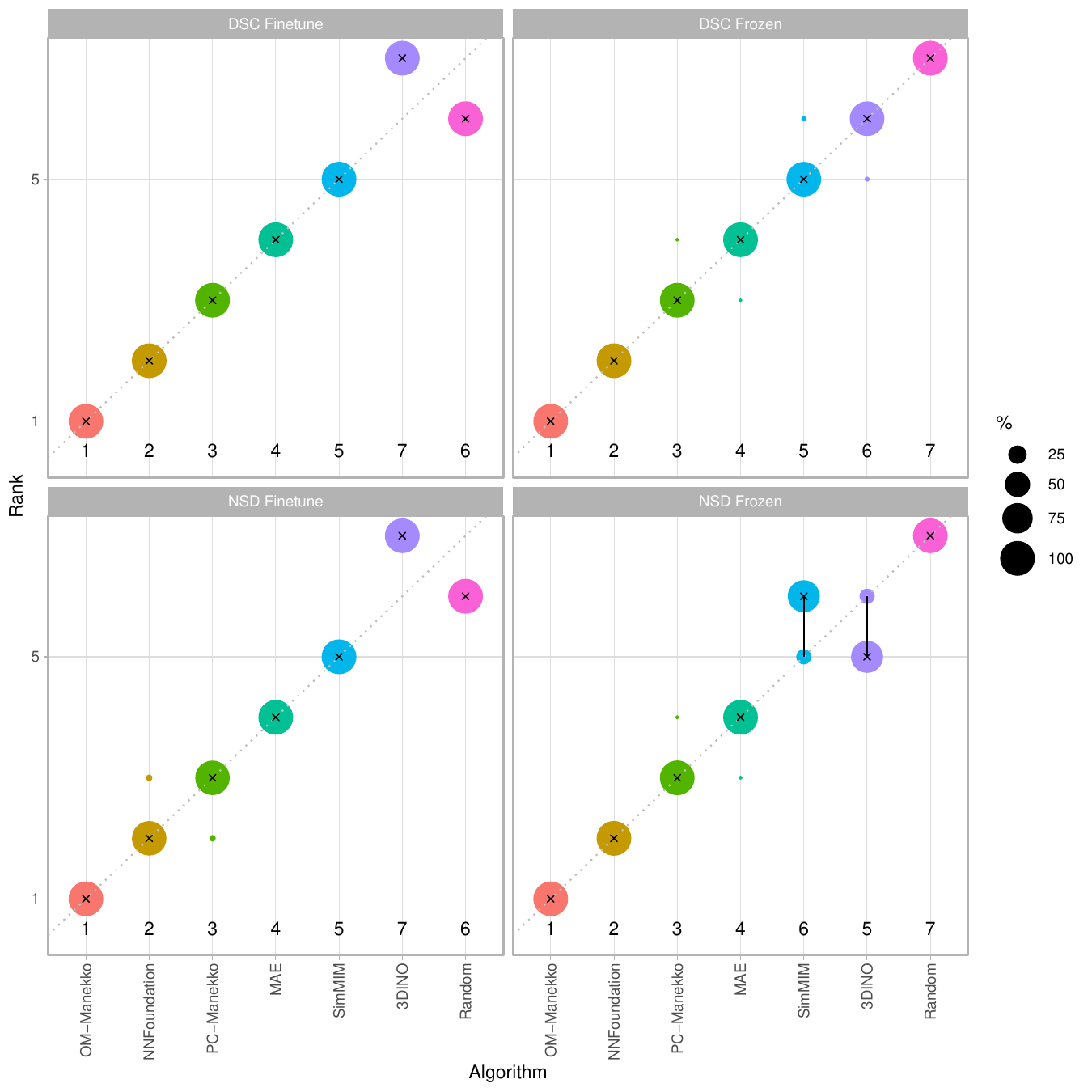}
    \caption{
    The panels show on the top row DSC and then NSD ($1\,\mathrm{mm}$ tolerance), with full fine-tuning on the left and frozen encoders on the right.
    Circle areas encode within-setting bootstrap rank frequencies. Black crosses and vertical lines mark median ranks and central $95\%$ bootstrap intervals.
    Algorithms retain a common horizontal order given by the cross-setting consensus ranking.
    The dotted diagonal marks agreement with this consensus order, and the numbers at the bottom of each panel give the full-data ranks for that setting.
    Vertical spread indicates sampling sensitivity, and displacement from the diagonal indicates a setting-specific departure from the consensus order.
    This view allows the separation of competing algorithms to be compared across settings.}
    \label{fig:rr_blobs_task}
\end{figure*}

We show three complementary views of these same bootstrap results.
The stacked frequency plots in Fig.~\ref{fig:rr_stacked} summarize the ranks attained by each algorithm across the 2 regimes and 2 metrics (total 4 settings).
The algorithm-wise \emph{blob plots} in Fig.~\ref{fig:rr_blobs_algorithm} separate these settings explicitly, making it easier to distinguish changes between settings from uncertainty within a setting.
Finally, the setting-wise blob plots in Fig.~\ref{fig:rr_blobs_task} compare all algorithms within each metric-regime combination.
In a blob plot, circle area represents bootstrap frequency: concentration at one rank indicates a stable position, and spread over several ranks indicates sensitivity to which cases are sampled.
Black crosses and vertical intervals summarize the median and central $95\%$ of the bootstrap rank distribution.
The distributions for \omours are concentrated at rank~1 across all four settings, supporting the stability of its leading pooled position.
These visualizations are alternative summaries of the same resampling analysis, rather than independent confirmations.

\end{document}